\pdfoutput=1

\documentclass[11pt]{article}

\usepackage{EMNLP2022}

\usepackage{times}
\usepackage{latexsym}

\usepackage[T1]{fontenc}

\usepackage[utf8]{inputenc}

\usepackage{microtype}

\usepackage{inconsolata}

\usepackage{xspace,mfirstuc,tabulary}
\usepackage{booktabs}
\usepackage{amssymb}
\usepackage{amsmath}
\usepackage{multirow,booktabs, hhline}
\usepackage[ruled,noend]{algorithm2e}
\usepackage{amsmath, bm}
\usepackage{graphicx}
\usepackage{color}
\usepackage{subfigure}
\usepackage{adjustbox}
\usepackage{cleveref}
\crefname{section}{§}{§§}
\Crefname{section}{§}{§§}
\crefname{figure}{Figure}{Figure}
\Crefname{figure}{Figure}{Figure}
\crefname{table}{Table}{Table}
\Crefname{table}{Table}{Table}
\newcommand\RBT{RoBERTa$_{\textsc{Base}}$\xspace}
\newcommand\Tasknum{$7$\xspace}

\title{Finding Skill Neurons in Pre-trained Transformer-based Language Models}

\author{ Xiaozhi~Wang$^{1}\thanks{\quad indicates equal contribution.}$\hspace{0.5em}, Kaiyue~Wen$^{2*}$, Zhengyan~Zhang$^{1}$,\\ \textbf{Lei~Hou}$^{1,3}$\thanks{\quad Corresponding author: Z.Liu and L.Hou.}, \textbf{Zhiyuan~Liu}$^{1,3\dag}$, \textbf{Juanzi~Li}$^{1,3}$\\
 $^1$Department of Computer Science and Technology, BNRist;\\
$^2$Institute for Interdisciplinary Information Sciences;\\
  $^3$KIRC, Institute for Artificial Intelligence,\\
  Tsinghua University, Beijing, 100084, China \\
  \texttt{\{wangxz20,wenky20\}@mails.tsinghua.edu.cn}
}

\begin{document}
\maketitle
\begin{abstract}
Transformer-based pre-trained language models have demonstrated superior performance on various natural language processing tasks. However, it remains unclear how the skills required to handle these tasks distribute among model parameters. In this paper, we find that after prompt tuning for specific tasks, the activations of some neurons within pre-trained Transformers\footnote{For brevity, \textit{Transformer-based language models} are often referred to as \textit{Transformers} in this paper.} are highly predictive of the task labels. We dub these neurons \textit{skill neurons} and confirm they encode task-specific skills by finding that: (1) Skill neurons are crucial for handling tasks. Performances of pre-trained Transformers on a task significantly drop when corresponding skill neurons are perturbed. (2) Skill neurons are task-specific. Similar tasks tend to have similar distributions of skill neurons. Furthermore, we demonstrate the skill neurons are most likely generated in pre-training rather than fine-tuning by showing that the skill neurons found with prompt tuning are also crucial for other fine-tuning methods freezing neuron weights, such as the adapter-based tuning and BitFit. We also explore the applications of skill neurons, including accelerating Transformers with network pruning and building better transferability indicators. These findings may promote further research on understanding Transformers. The source code can be obtained from \url{https://github.com/THU-KEG/Skill-Neuron}.

\end{abstract}

\section{Introduction}
\label{sec:intro}
Pre-trained language models (PLMs), mostly based on Transformer architecture~\citep{vaswani2017attention}, have achieved remarkable performance on broad and diverse natural language processing (NLP) tasks~\citep{han2021pre}. However, it remains unclear how the skills required to handle these tasks distribute among model parameters. Are there specific neurons within pre-trained Transformers encoding these skills? Progress on this problem may help to understand the working mechanisms of pre-trained Transformers~\citep{zeiler2014visualizing,karpathy2015visualizing,bau2020understanding,suau2020finding}, intervene model behaviors~\citep{bau2018identifying,mitchell2021fast}, and improve model efficiency~\citep{dalvi2020analyzing,zhang2021moefication}.

\begin{figure}[!t]
\centering
\includegraphics[width=0.45\textwidth]{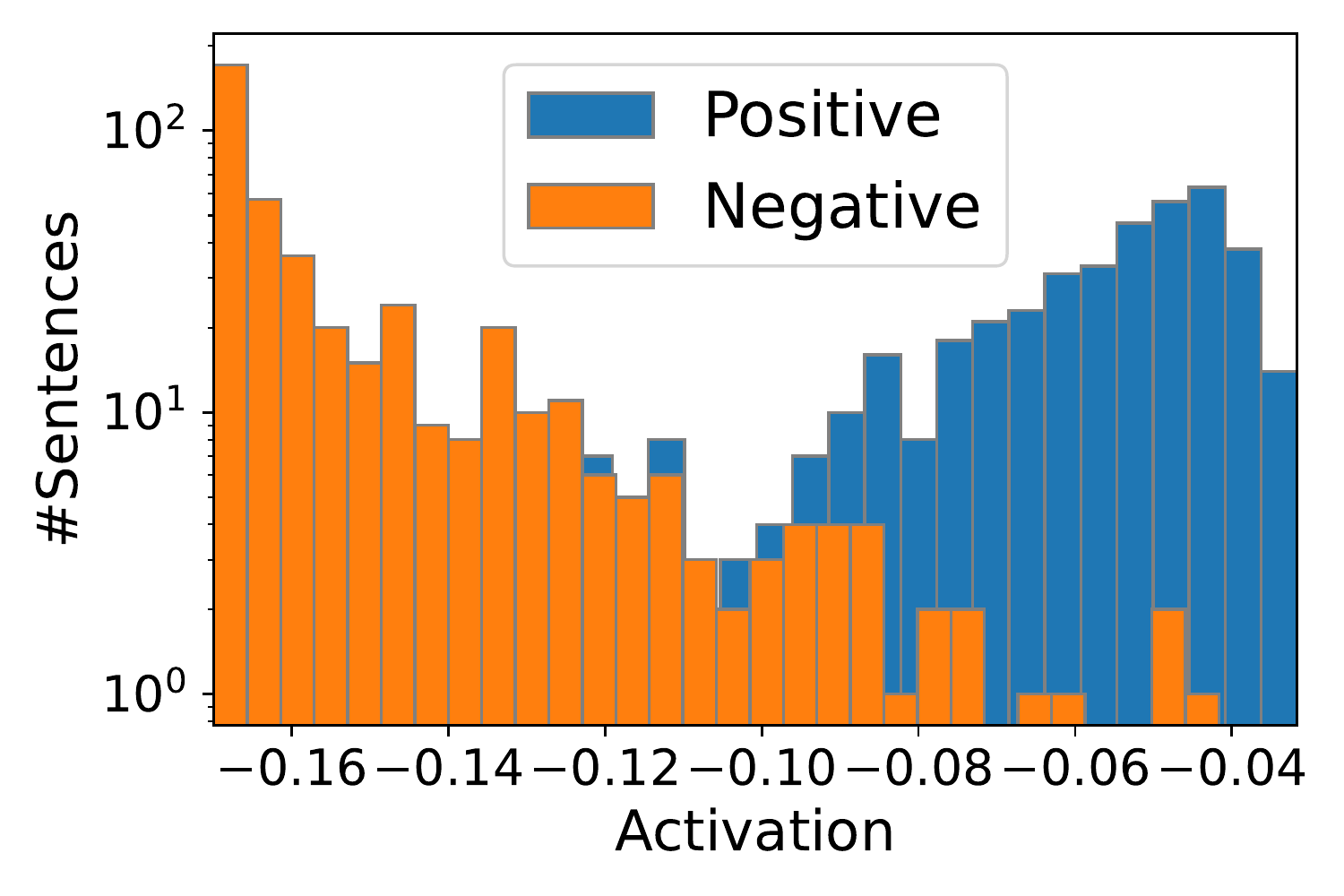}
\caption{Histogram of activation of a neuron within \RBT on positive-label (\textcolor{blue}{blue}) and negative-label (\textcolor{orange}{orange}) sentences in \texttt{SST-2} validation set.}
\label{fig:activation_demo}
\end{figure}

Prompt tuning~\citep{li-liang-2021-prefix,lester2021power} prepends some trainable embeddings, i.e., \textit{soft prompts}, into the inputs and adapts PLMs to handle tasks by only tuning the soft prompts while freezing all the PLM parameters. It has attracted wide attention recently as a promising parameter-efficient fine-tuning methods~\citep{su2021transferability,liu2021ptuningv2}. In this paper, we find that after prompt tuning for a task, the activations on soft prompts of some neurons within pre-trained Transformers are \textbf{highly predictive} for the task. 
For instance, \cref{fig:activation_demo} shows the activation distribution of a specific neuron within \RBT~\citep{liu2019roberta}. This neuron's activation is highly predictive of the labels of \texttt{SST-2}~\citep{socher-etal-2013-recursive}, an established sentiment analysis dataset. When the input sentences express positive sentiments, the activations on soft prompts of this neuron tend to be much higher than when they express negative sentiments. It suggests that this neuron may encode the skill of distinguishing sentiments.

We dub these special neurons \textit{skill neurons} and develop a simple and effective method to find them for classification tasks via prompt tuning. For a binary classification task, we first calculate the empirical mean activation on a soft prompt token over the training set for each neuron and use it as this neuron's baseline activation. If this neuron's activation for an input sample is higher than the baseline, we regard it as predicting one label and vice versa. We aggregate the prediction accuracies on the validation set of multiple soft prompts as the neuron's predictivity score. The neurons with the highest predictivity scores are identified as skill neurons. For multi-class classification tasks, we decompose them into multiple binary classification subtasks and aggregate the skill neurons of subtasks as the skill neurons of the multi-class task.

We confirm the skill neurons encode task-specific skills with a series of experimental findings: (1) Skill neurons generally and stably emerge. For all the \Tasknum investigated tasks and $5$ random trials, we can consistently find skill neurons with high predictivities close to prompt tuning. (2) Skill neurons are crucial for handling tasks. When we perturb skill neurons by adding random noises to their activations, the performances on corresponding tasks drop much more significantly than when random neurons are perturbed. (3) Skill neurons are task-specific. Similar tasks exhibit similar predictivity rankings of skill neurons, and skill neurons of same-type tasks are more important for handling a task than those of different-type tasks. (4) Skill neurons are not from shallow word selectivity. The skill neurons typically do not selectively activate on keywords relating to the task, and their predictivities are not significantly influenced by the label words used in prompt tuning.

After showing that skill neurons encode skills, we further demonstrate that skill neurons are most likely generated in pre-training rather than manufactured by the fine-tuning process of prompt tuning. This is concluded from: (1) Even for randomly generated prompts and untuned hard prompts, the skill neurons still exhibit much better predictivity performance than random guesses. (2) Skill neurons are also crucial for other fine-tuning methods freezing neuron weights. Performance of models trained with adapter-based tuning~\citep{houlsby2019parameter} and BitFit~\citep{BenZaken2021BitFitSP} significantly drops when the skill neurons found with prompt tuning are perturbed.

Moreover, we explore the practical applications of skill neurons. First, we apply skill neurons to network pruning~\citep{anwar2017structured,dalvi2020analyzing}, which aims at removing redundant parameters to reduce memory cost and accelerate inference. Experiments show that by only keeping top skill neurons active, we can reduce the pre-trained Transformer to $66.6\%$ of its original parameters and achieve about $1.4$ inference speedup. Then we explore building better prompt transferability indicators following~\citet{su2021transferability}. We improve their \textit{overlapping rate of activated neurons} metric by only taking skill neurons into account, and this achieves significantly better performance.

To summarize, our contributions are four-fold: (1) We observe the existence of skill neurons, the special neurons within pre-trained Transformers, which are highly predictive for specific tasks, and develop a method to find them via prompt tuning. (2) We empirically confirm that skill neurons do encode the skills required to handle tasks. (3) We show skill neurons are generated in pre-training rather than fine-tuning. (4) We preliminarily explore the applications of skill neurons. We hope these findings could facilitate future research on understanding the mechanism of PLMs.
\section{Preliminary}
We introduce the basic knowledge about prompt tuning~(\cref{sec:pre_PT}), the definition of investigated neurons~(\cref{sec:pre_neuron}), and the investigation setup~(\cref{sec:setup}).

\subsection{Prompt Tuning}
\label{sec:pre_PT}

Prompt tuning (PT), or soft prompting, is a recently-developed parameter-efficient fine-tuning method, which has attracted wide attention with its capability to effectively adapt PLMs to downstream tasks~\citep{li-liang-2021-prefix,lester2021power} and query inner knowledge of PLMs~\citep{qin-eisner-2021-learning,zhong-etal-2021-factual}. PT prepends some \textit{soft prompts} into the input sequences to prompt the PLM to decode the desired \textit{label words} of the training task in the same way as the pre-training objective. For each task, a \textit{verbalizer} function~\citep{schick-schutze-2021-exploiting} is used to map the specific label words to the labels of the task. Each soft prompt is a virtual token, which is essentially a trainable embedding. During prompt tuning, only the parameters in soft prompts are tuned, and all the PLM's original parameters are frozen.

Formally, given an input sequence with $n$ tokens $X = \{w_1, w_2,\ldots,w_{n}\}$, prompt tuning prepends $l$ randomly initialized soft prompts $P = \{p_1,p_2,\ldots,p_l\}$ before them, where $p_{i}\in \mathbb{R}^{d}$ and $d$ is the input dimension of the PLM. Taking the PLMs pre-trained with the masked language modeling objective~\citep{devlin-etal-2019-bert} as an example, a special \texttt{[MASK]} token is prepended, and the prompt tuning objective is to maximize the likelihood of filling desired label word $y$ into it:
\begin{equation}
\small
\label{eq:prompt_tuning}
    \mathcal{L}=p(y|\mathtt{[MASK]},P,x_1,\ldots,x_n).
\end{equation}

Some initial prompt tuning works~\citep{qin-eisner-2021-learning,zhong-etal-2021-factual} regard soft prompts as the relaxation of natural language \textit{hard} prompts, which are initially designed to query inner factual knowledge of PLMs~\citep{petroni-etal-2019-language,jiang-etal-2020-know}. \citet{su2021transferability} hypothesize that soft prompts work by stimulating PLMs' inner abilities. Inspired by these, we observe the inner activations of PLMs and find skill neurons.

\subsection{Neurons in Transformers}
\label{sec:pre_neuron}

Transformer~\citep{vaswani2017attention} is the state-of-the-art NLP model architecture, which is used by the majority of PLMs~\citep{devlin-etal-2019-bert,liu2019roberta,brown2020GPT3,raffel2020T5}. A pre-trained Transformer model is typically stacked with multiple identical Transformer layers. Each Transformer layer consists of a self-attention module and a feed-forward network (FFN), among which the FFN carries two-thirds of the parameters. Previous work has highlighted the importance of FFN~\citep{press-etal-2020-improving,dong2021attention} and found FFN encodes rich information~\citep{suau2020finding,geva2021transformer,dai2021knowledge}. Inspired by these, we study the neurons and activations within FFN.

Formally, the FFN in a Transformer layer is: 
\begin{equation}
\small
\begin{aligned}
\mathrm{FFN}(\mathbf{x}) = f(\mathbf{x} \mathbf{K}^{\top}+\mathbf{b}_1) \mathbf{V} + \mathbf{b}_2,
\end{aligned}
\label{eq:ffn}
\end{equation}
where $\mathbf{x}\in\mathbb{R}^{d}$ is the hidden embedding of a token, $f(\cdot)$ is the activation function, 
$\mathbf{K},\mathbf{V}\in\mathbb{R}^{d_{m} \times d}$ are trainable matrices, and $\mathbf{b}_1, \mathbf{b}_2$ are biases.

For simplicity, let $\mathbf{a}=f(\mathbf{x} \mathbf{K}^{\top}+\mathbf{b}_1) \in \mathbb{R}^{d_m}$. We regard $\mathbf{a}_{i}$, the $i$-th element of $\mathbf{a}$, as the activation of the $i$-th neuron on input $\mathbf{x}$. It represents the importance of $\mathbf{K}_{i}$ and $\mathbf{V}_{i}$, the $i$-th column vectors of $\mathbf{K}$ and $\mathbf{V}$, respectively. Hence we define $\mathbf{K}_{i}$ and $\mathbf{V}_{i}$ as the weights of the $i$-th neuron in this layer.

Although they study essentially the same parameters as us, \citet{dai2021knowledge} and \citet{zhang2021moefication} use the term neuron to denote activations in our definition. Some other works~\citep{dalvi2019one,durrani2020analyzing,hennigen2020intrinsic,antverg2021pitfalls} define a dimension in contextualized representations as a neuron. Since we study how the skills distribute among model parameters rather than input-dependent representations, we study the neurons defined in this section.

\subsection{Investigation Setup}
\label{sec:setup}

To comprehensively investigate the skill neuron phenomenon, we use \RBT~\citep{liu2019roberta}, a widely-used Transformer model pre-trained with the masked language modeling objective~\citep{devlin-etal-2019-bert}, and conduct experiments on \Tasknum tasks of $3$ types, including: (1) \textbf{Sentiment Analysis}, including \texttt{SST-2}~\citep{socher-etal-2013-recursive}, \texttt{IMDB}~\citep{maas-EtAl:2011:ACL-HLT2011}, and TweetEval (\texttt{Tweet})~\citep{barbieri-etal-2020-Tweeteval}; (2) \textbf{Natural Language Inference}, including \texttt{MNLI}~\citep{williams-etal-2018-broad} and \texttt{QNLI}~\citep{wang2018glue}; (3) \textbf{Topic Classification}, including \texttt{AG News} and \texttt{DBpedia}~\citep{zhang2015character}. Details about the tasks and prompt tuning implementations are shown in \cref{app:task,app:pt}, respectively.

\section{Finding Skill Neurons}
\label{sec:find_method}

We use a simple and effective method to find skill neurons for a given pre-trained Transformer $\mathcal{M}$.

\subsection{Binary Classification Task}
\label{sec:binary_find}

We first introduce how to find skill neurons for binary classification tasks. Let $\mathcal{T}$ be a binary classification task and its dataset be $D=\{\left(x_1,y_1\right),\left(x_2,y_2\right),\ldots,\left(x_{|D|},y_{|D|}\right)\}$, which is divided into training set $D_{\mathrm{train}}$, development set $D_{\mathrm{dev}}$, and test set $D_{\mathrm{test}}$. The $i$-th sample $\left(x_i,y_i\right)$ contains an input $x_i$ and its label $y_{i}\in\{0,1\}$.

For a specific neuron $\mathcal{N}$ within $\mathcal{M}$, let $a(\mathcal{N},t,x)$ be the activation of it on token $t$ given the input sentence $x$. We firstly do prompt tuning on $\mathcal{M}$ with $D_{\mathrm{train}}$ and get a group of $l$ soft prompts $P=\{p_1,p_2,\ldots,p_l\}$. Given a soft prompt $p_i$, we calculate the baseline activation of $\mathcal{N}$ on $p_i$ over the training set as follows:

\begin{equation}
\small
\begin{aligned}
a_{\mathrm{bsl}}(\mathcal{N},p_i)=\frac{1}{|D_{\mathrm{train}}|}\sum_{x_j,y_j \in D_{\mathrm{train}}} a(\mathcal{N},p_i,x_j).
\end{aligned}
\label{eq:baseline_activation}
\end{equation}

Intuitively, we can regard that the neuron $\mathcal{N}$ predicts positive label $1$ for the input sentence $x$ when $a(\mathcal{N},p_i,x)>a_{\mathrm{bsl}}(\mathcal{N},p_i)$. Hence the prediction accuracy over the development set is as follows:

\begin{equation}
\small
\begin{aligned}
\mathrm{Acc}(\mathcal{N},p_i)=\frac{\sum_{x_j,y_j \in D_{\mathrm{dev}}} \mathbf{1}_{[\mathbf{1}_{[a(\mathcal{N},p_i,x_j)>a_{\mathrm{bsl}}(\mathcal{N},p_i)]}=y_j]}}{|D_{\mathrm{dev}}|},
\end{aligned}
\label{eq:accuracy}
\end{equation}
where $\mathbf{1}_{[\mathrm{condition}]}\in \{0,1\}$ is the indicator function evaluating to $1$ iff the $\mathrm{condition}$ holds. 

The above way only considers the positive correlations between the labels and neuronal activations, which is also the case of previous work~\citep{geva2021transformer,dai2021knowledge}. However, strong negative correlations also suggest that the information about skills is encoded in this neuron. Conceptually, this is similar to the fact that inhibitory neurons in brains also contribute to certain functions~\citep{rudy2011three}. Hence we define the predictivity of $\mathcal{N}$ on soft prompt token $p_i$ as:

\begin{equation}
\small
\begin{aligned}
\mathrm{Pred}(\mathcal{N},p_i)=\mathrm{max}(\mathrm{Acc}(\mathcal{N},p_i), 1-\mathrm{Acc}(\mathcal{N},p_i)).
\end{aligned}
\label{eq:predictivity}
\end{equation}

For each group of soft prompts $P$, the predictivity of $\mathcal{N}$ on it is defined as the predictivity of the best soft prompt token. Considering the skill neurons shall be consistently predictive, we conduct $5$ random trials of prompt tuning and get $5$ groups of prompts: $\mathcal{P}=\{P_1,P_2,\ldots,P_5\}$. The overall predictivity of neuron $\mathcal{N}$ is defined as:

\begin{equation}
\small
\begin{aligned}
\mathrm{Pred}(\mathcal{N})=\frac{1}{\left|\mathcal{P}\right|} \sum_{P_i\in\mathcal{P}}\mathrm{max}_{p_j\in P_i}(\mathrm{Pred}(\mathcal{N},p_j)).
\end{aligned}
\label{eq:overall_predictivity}
\end{equation}

Then we sort all the neurons within model $\mathcal{M}$ by the descending order of their predictivities and use the top neurons as the skill neurons in experiments. \Cref{app:design_choice} discusses some potential design choices considered in finding skill neurons.

\begin{figure}[!t]
\centering
\includegraphics[width=0.45\textwidth]{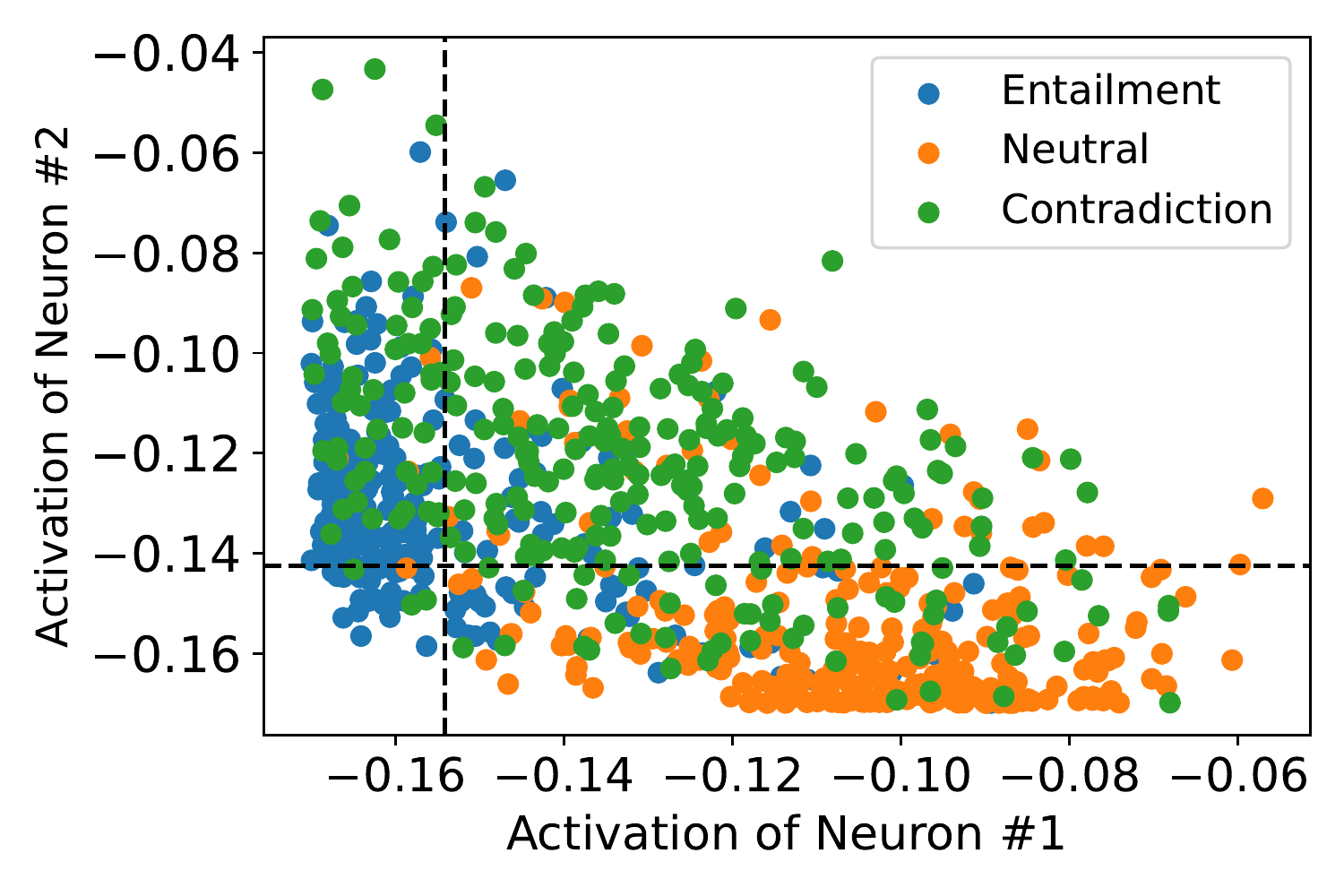}
\caption{Distribution of activations of two neurons on a soft prompt for samples in \texttt{MNLI} validation set. Dashed lines indicate baseline activations of the two neurons. }
\label{fig:mnli_demo}
\end{figure}

\subsection{Multi-class Classification Task}
To find skill neurons for a multi-class classification task, we first decompose it into multiple binary classification subtasks. Then we find skill neurons by ranking the neurons with their predictivities of the decomposed subtasks in a similar way as introduced in \cref{sec:binary_find} but use the soft prompts of the original task instead of subtasks. Skill neurons of the multi-class classification task consist of equal numbers of subtask skill neurons. For instance, \texttt{MNLI}~\citep{williams-etal-2018-broad} task requires to classify the relationships between sentence pairs into \textsc{Entailment}, \textsc{Neutral} and \textsc{Contradiction}. We decompose it into two subtasks: the first one is to classify \textsc{Entailment} and \textsc{Contradiction} samples, and the second one is to classify \textsc{Neutral} and \textsc{Non-neutral} samples. If we need top-$100$ skill neurons of \texttt{MNLI}, we will retrieve top-$50$ unique skill neurons for the two subtasks, respectively. \cref{fig:mnli_demo} shows the activation distribution of the two top skill neurons within \RBT of the two subtasks, respectively. The samples of three labels form three distinguishable clusters, which suggests the effectiveness of this skill-neuron-finding method. More details about how we decompose the investigated tasks are shown in \cref{app:task}.

\begin{table}[t!]
\small
\centering
\begin{tabular}{lrr}
\toprule
Task             & \multicolumn{1}{c}{\begin{tabular}[c]{@{}c@{}}Prompt\\ Tuning\end{tabular}} & \multicolumn{1}{c}{\begin{tabular}[c]{@{}c@{}}Skill\\ Neuron\end{tabular}} \\ \midrule
\texttt{SST-2}   & $91.8_{\pm0.5}$                                                             & $91.6_{\pm0.3}$                                                            \\
\texttt{IMDB}    & $91.6_{\pm0.5}$                                                             & $92.0_{\pm0.3}$                                                            \\
\texttt{Tweet}   & $70.0_{\pm0.2}$                                                             & $56.0_{\pm3.2}$                                                            \\
\texttt{MNLI}    & $76.8_{\pm1.8}$                                                             & $74.7_{\pm2.5}$                                                            \\
\texttt{QNLI}    & $85.7_{\pm0.7}$                                                             & $86.0_{\pm0.4}$                                                            \\
\texttt{AG News} & $98.8_{\pm0.1}$                                                             & $98.9_{\pm0.1}$                                                            \\
\texttt{DBpedia} & $99.7_{\pm0.1}$                                                             & $99.8_{\pm0.1}$                                                            \\ \bottomrule
\end{tabular}
\caption{Accuracies (\%) on various tasks of prompt tuning and skill neurons, along with standard deviations over $5$ random trials. For the binary classification tasks, the skill neuron performance is the predictivity of the top-1 skill neuron. For multi-class classification tasks, the skill neuron performance is obtained by training a logistic regression model taking only the activations of the top-1 neurons of decomposed subtasks as inputs.}
\label{tab:neuron_performance}
\end{table}

\section{Do Skill Neurons Encode Skills?}
\label{sec:do_encode}
We explore whether skill neurons really encode task-specific skills with a series of experiments.
\subsection{Skill Neurons Generally and Stably Emerge}
\label{sec:stable}
We first confirm that the skill neuron phenomenon is general and stable for various NLP tasks. 

\paragraph{Generality.} To explore whether we can generally find highly-predictive skill neurons for various tasks, we apply the skill-neuron-finding method in \cref{sec:find_method} to \Tasknum NLP tasks introduced in \cref{sec:setup}. The performances of the top-predictivity found skill neurons and prompt tuning are shown in \cref{tab:neuron_performance}. For all the tasks, we can find skill neurons achieving comparable performance to prompt tuning, which demonstrates specific skill neurons generally exist in pre-trained Transformers for various tasks.
\begin{figure}[!t]
\centering
\includegraphics[width=0.48\textwidth]{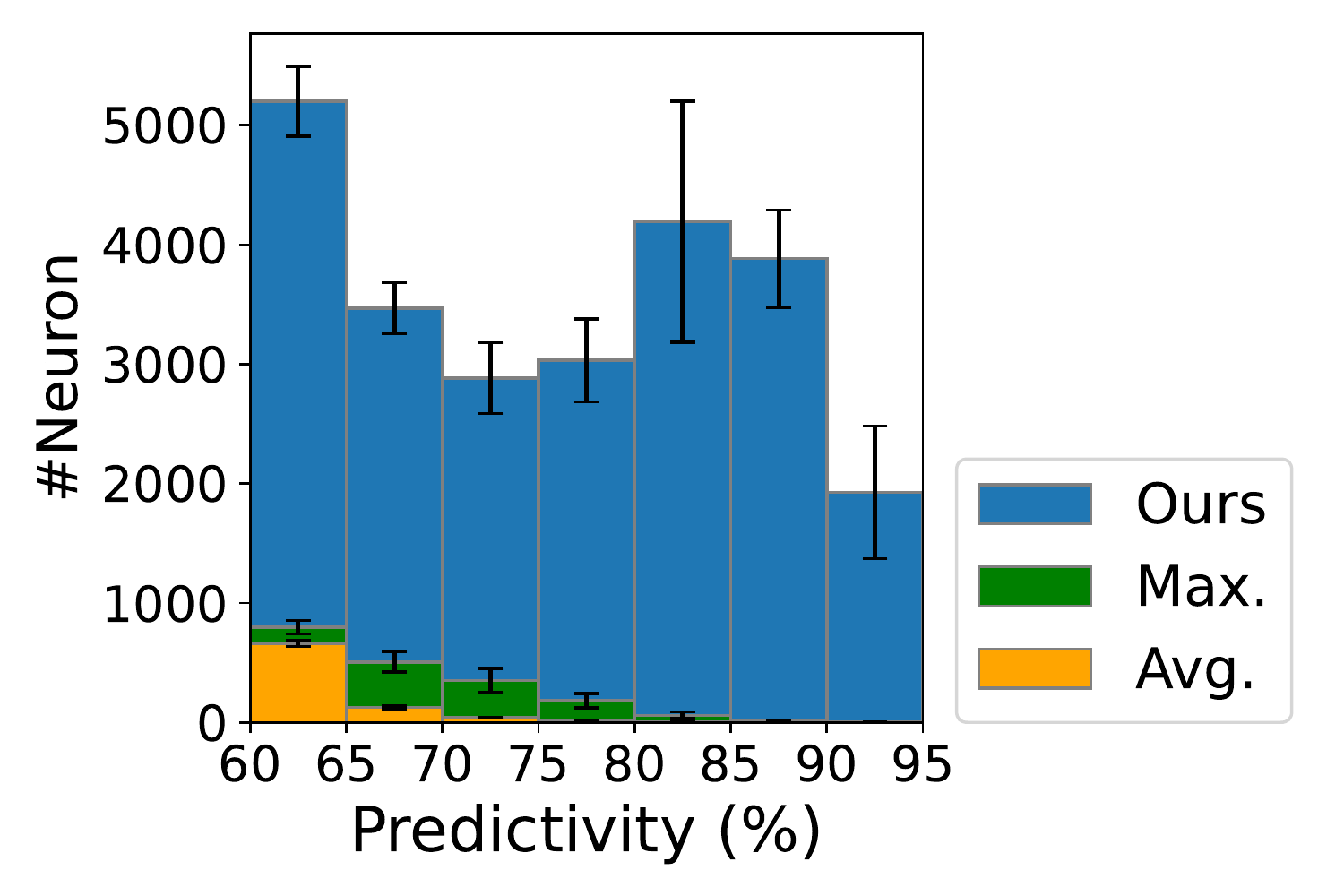}
\caption{Histogram of neuron's predictivity for \texttt{IMDB}. Error bars indicate $\pm 1$ s.e.m. over $5$ random trials.}
\label{fig:acc_dist}
\end{figure}

\paragraph{Stability.} To rule out the possibility that the skill neurons are just from randomness and confirm the stability of this phenomenon, we conduct $5$ random trails (with different data orders and prompt initializations) to find skill neurons for all the tasks. \cref{fig:acc_dist} shows the distributions of neuron predictivities within \RBT for \texttt{SST-2} task. Distributions for the other tasks are left in \cref{app:acc_dist}. We can see that our method can stably find substantial skill neurons with high predictivities via prompts. Previous methods use average~\citep{dai2021knowledge} and maximum~\citep{suau2020finding} activations on input tokens instead of activations on prompts to find selective neurons, which are shown as the ``Avg.'' and ``Max.'' results in \cref{fig:acc_dist}, respectively. The experimental results indicate that previous methods hardly find highly-predictive neurons, which suggests that prompt tuning is crucial for finding skill neurons. We encourage future work to explore the reason why prompt tuning can help in this.

\subsection{Skill Neurons are Crucial for Handling Tasks}
\begin{figure}[!t]
\centering
\includegraphics[width=0.48\textwidth]{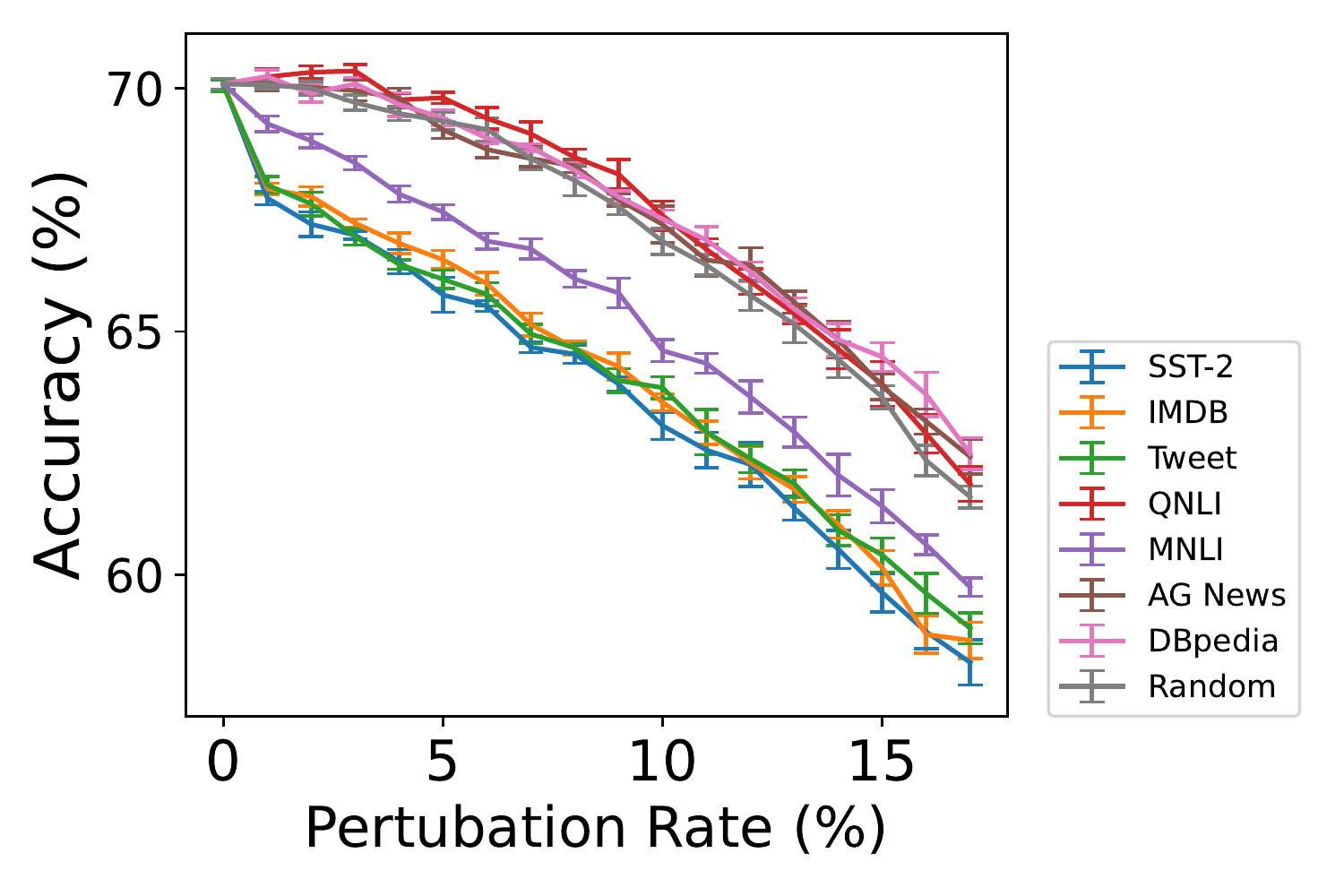}
\caption{Accuracy on \texttt{Tweet} drops along with the neuron perturbation rate. Error bars indicate $\pm 1$ s.e.m. over $5$ random trials. The perturbations are conducted in descending orders of neurons' predictivities for different tasks or in random order (the ``Random'' curve).}
\label{fig:mask_trend}
\end{figure}
A natural hypothesis is that if the skill neurons really encode skills, they shall be more important for PLMs to handle various tasks. To verify this, we perturb the skill neurons and see whether PLM's performance drops more than perturbing random neurons. Specifically, the perturbation is to add a Gaussian noise ($\mu=0$ and $\sigma=0.1$) into the neurons' activations~\citep{Arora2018StrongerGB}, so that the neurons cannot function properly, and then we observe the PLM's prompt tuning performances. 

The perturbation results on \texttt{Tweet} task are shown in \cref{fig:mask_trend}, from which we observe that when we perturb top skill neurons of this task, the PLM's performance drops much more significantly than when we perturb neurons in random order. It indicates that the highly-predictive skill neurons are indeed crucial for handling tasks and supports that skill neurons encode skills. Perturbation results on the other tasks are shown in \cref{app:mask_pt}, and they all exhibit similar phenomena.
\subsection{Skill Neurons are Task-specific}
\label{sec:exp_task_specific}

We further study whether skill neurons are task-specific, i.e., do skill neurons encode task-specific high-level skills like distinguishing sentiments for sentiment analysis, or do they just encode some task-general low-level skills like recognizing parts of speech, which are also helpful for handling tasks.

\begin{figure}[!t]
\centering
\includegraphics[width=0.48\textwidth]{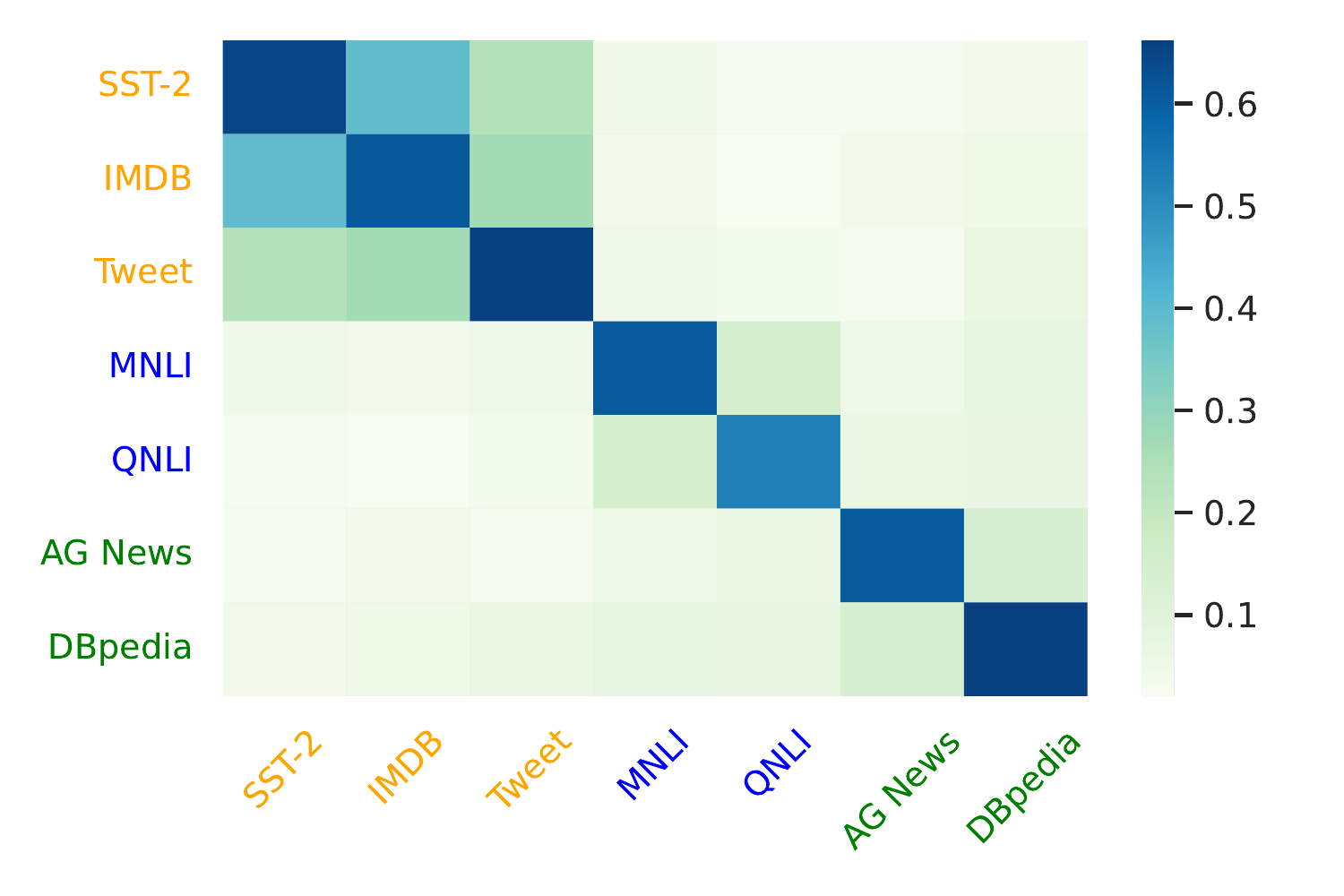}
\caption{Spearman's rank correlations between the neuron predictivity orders of different tasks. Results are averaged over all the layers.}
\label{fig:spearman_neuron}
\end{figure}

First, if skill neurons are task-specific, we shall find similar skill neurons for similar tasks. To verify this, we rank neurons in descending orders of their predictivities for different tasks and see Spearman's rank correlations~\citep{spearman1987proof} between the orders of different tasks. The average results over all the $12$ layers of \RBT are shown in \cref{fig:spearman_neuron}. We can see that the correlations between similar tasks of the same type are obviously higher, which confirms that similar tasks have similar skill neurons. The layer-wise correlations are shown in \cref{app:acc_dist}, from which we can see skill neurons tend to be more task-specific in higher layers, which is consistent with previous probing findings~\citep{Liu2019LinguisticKA}.

\begin{figure}[!t]
\centering
\includegraphics[width=0.48\textwidth]{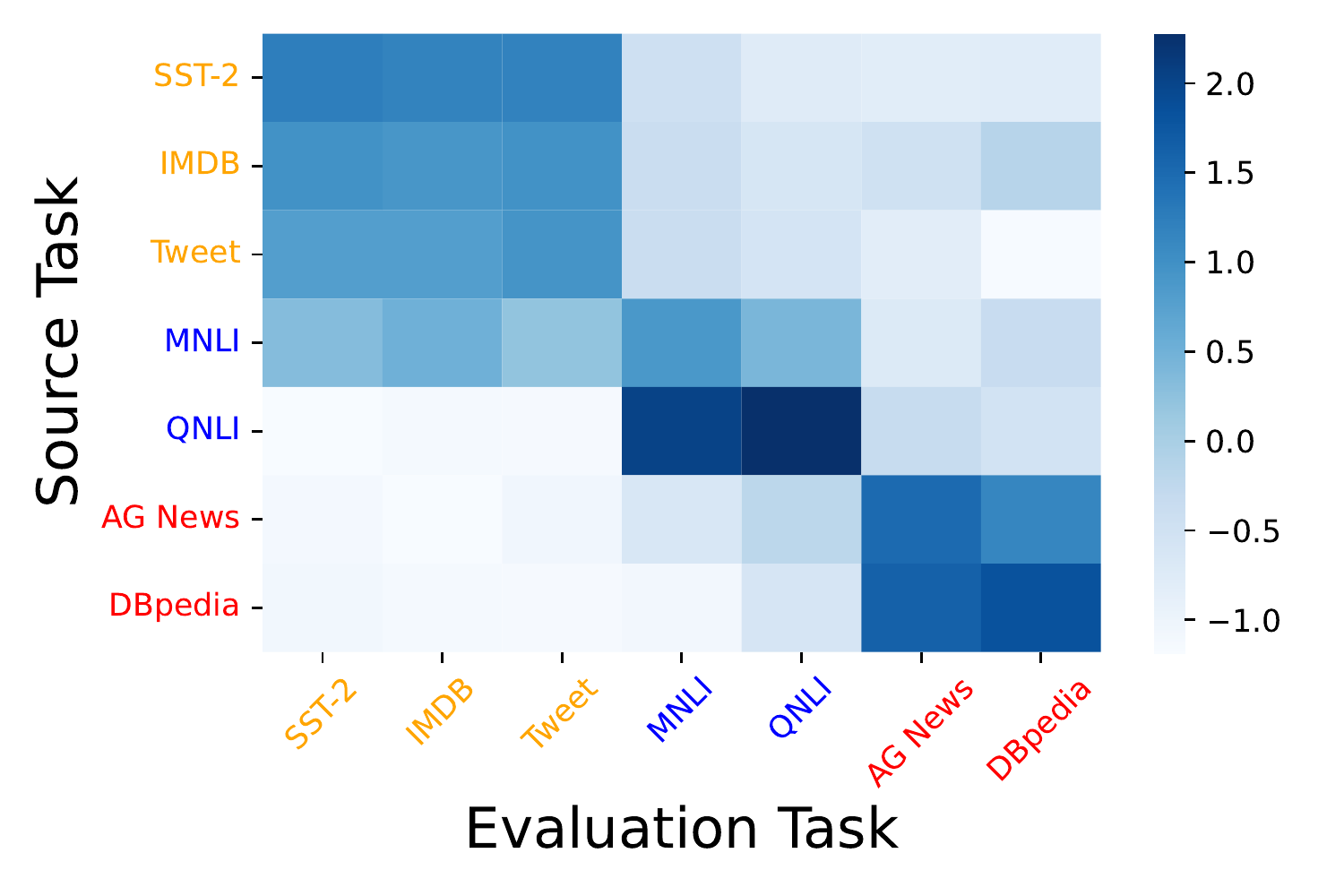}
\caption{Neuronal importances of different task pairs. Results are averaged over $5$ random trials. For an evaluation task, the neuronal importances of different source tasks are normalized as z-scores.}
\label{fig:neuronal_importance}
\end{figure}

Moreover, if skill neurons are task-specific, the skill neurons of same-type tasks shall be more important for handling a specific task. This has been supported by \cref{fig:mask_trend}, which shows that the accuracy on \texttt{Tweet} drops much more significantly when we perturb neurons in the predictivity orders of same-type tasks (\texttt{SST-2}, \texttt{IMDB}). To qualify this effect and comprehensively show this phenomenon in all tasks, we define the \textit{neuronal importance} of a source task to an evaluation task as the area between the accuracy curves obtained by perturbing neurons in the predictivity order of the source task and in random order. For instance, in \cref{fig:mask_trend}, the neuronal importance of \texttt{SST-2} to \texttt{Tweet} is the area between the blue curve and the gray curve. The overall neuronal importance is shown in \cref{fig:neuronal_importance}, from which we can see the skill neurons of same-type tasks are obviously more important, which strongly supports that the found skill neurons encode task-specific skills again.

\subsection{Skill Neurons are not from Word Selectivity}
\label{sec:word_select}
Previous works~\citep{dai2021knowledge,suau2020finding} show that neurons in Transformers may selectively activate on some words or concepts. To confirm that skill neurons encode skills, we show that skill neurons are not from these selectivities.

\begin{table}[t!]
\small
\begin{adjustbox}{max width=1\linewidth}
{
\begin{tabular}{lc}
\toprule
\multicolumn{2}{c}{Cosine Similarity}                                                                                                                                                                                                     \\ \midrule
Top    & \begin{tabular}[c]{@{}c@{}}\texttt{AGES}, \texttt{GES}, \texttt{ITIES}, \texttt{ause}, \texttt{UNCH}, \\ \texttt{AGE}, \texttt{ORK}, \texttt{STE}, \texttt{TING}, \texttt{FE}\end{tabular}                                       \\ \midrule
Bottom & \begin{tabular}[c]{@{}c@{}}\texttt{sham}, \texttt{Nicol}, \texttt{bogus}, \texttt{Rox}, \texttt{Nay}, \texttt{contro}, \\ \texttt{guy}, \texttt{uneven}, \texttt{arbitrarily}, \texttt{unnatural}\end{tabular}                   \\ \midrule \midrule
\multicolumn{2}{c}{Average Activation}                                                                                                                                                                                                            \\ \midrule
Top    & \begin{tabular}[c]{@{}c@{}}\texttt{starters}, \texttt{village}, \texttt{oster}, \texttt{iddled}, \texttt{af},\\ \texttt{mafia}, \texttt{aley}, \texttt{tired}, \texttt{dep}, \texttt{ophobic}\end{tabular}                       \\ \midrule
Bottom & \begin{tabular}[c]{@{}c@{}}\texttt{official}, \texttt{repression}, \texttt{illegal},\\ \texttt{called}, \texttt{ensible}, \texttt{regime}, \texttt{abusers}, \\ \texttt{should}, \texttt{creation}, \texttt{refuse}\end{tabular} \\ \bottomrule
\end{tabular}
}
\end{adjustbox}
\caption{Related words for \texttt{SST-2}'s top skill neuron.}
\label{tab:words_SST2}
\end{table}

We first do case studies on the related words of the top skill neurons, including the words with top and bottom cosine similarities between their input embeddings and the neuron weight vectors~\citep{dai2021knowledge}, and the words with top and bottom average activations~\citep{suau2020finding}. The results of \texttt{SST-2} are shown in \cref{tab:words_SST2}. We can see these related words do not convey sentiments, which demonstrates the skill neurons are not from keyword selectivities. Results of the other tasks are shown in \cref{app:word_selectivity}.

Furthermore, considering the prompt tuning method does predictions by decoding label tokens, we need to check whether skill neurons depend on the label words used. If so, it indicates that the skill neurons do not encode the skills for handling tasks but encode the skills for selectively decoding some words. We rule out this possibility by finding that if we use different random words as label words, the achieved predictivity orders of neurons are pretty consistent. Specifically, for all the tasks, the average Spearman's correlation between the neuron predictivity orders of $5$ random label words is 0.87.

\section{Where do Skill Neurons Come from?}
In \cref{sec:do_encode}, we confirm that skill neurons do encode task-specific skills. Then a natural question is where skill neurons come from, i.e., do skill neurons acquire these skills in pre-training or prompt tuning? We find that skill neurons are most likely \textbf{generated in pre-training} with empirical evidence.

\begin{table}[t!]
\small
\centering
\begin{tabular}{lrrrr}
\toprule
Task             & \multicolumn{1}{c}{\begin{tabular}[c]{@{}c@{}}Random\\ Guess\end{tabular}} & \multicolumn{1}{c}{\begin{tabular}[c]{@{}c@{}}Random\\ Model\end{tabular}} & \multicolumn{1}{c}{\begin{tabular}[c]{@{}c@{}}Random\\ Prompt\end{tabular}} & \multicolumn{1}{c}{\begin{tabular}[c]{@{}c@{}}Hard\\ Prompt\end{tabular}} \\ \midrule
\texttt{SST-2}   & $50.0$  & $52.8_{\pm0.4}$ & $78.1_{\pm0.4}$                                                                & $83.3$                                                                    \\
\texttt{IMDB}    & $50.0$  & $58.0_{\pm0.7}$                                                                   & $76.7_{\pm2.0}$                                                                & $75.1$                                                                    \\
\texttt{Tweet}   & $33.3$  & $48.3_{\pm0.0}$                                                                   & $48.2_{\pm1.8}$                                                                & $48.6$                                                                    \\
\texttt{MNLI}    & $33.3$ & $32.2_{\pm0.4}$                                                                    & $39.8_{\pm1.1}$                                                                & $40.5$                                                                    \\
\texttt{QNLI}    & $50.0$ & $54.3_{\pm0.8}$                                                                    & $69.5_{\pm0.5}$                                                                & $65.2$                                                                    \\
\texttt{AG News} & $50.0$ & $62.7_{\pm0.3}$                                                                    & $96.0_{\pm0.3}$                                                                & $95.9$                                                                    \\
\texttt{DBpedia} & $50.0$ & $60.9_{\pm0.4}$ & $98.8_{\pm0.1}$ & $99.2$ \\ \bottomrule
\end{tabular}
\caption{Accuracies (\%) on various tasks of top skill neurons found with random prompts and untuned hard prompts, compared to random guess and random model. We also report standard deviations over $5$ random trials.}
\label{tab:perf_random}
\end{table}

We first try to find skill neurons with tuning-free prompts, including random prompts, which are randomly generated embeddings, and human-written hard prompts. The predictivities of the found neurons are shown in \cref{tab:perf_random}. We can see that even without tuning, we can still find neurons with non-trivial predictivities. \citet{malach2020proving} shows that randomly initialized neural networks may have predictive subnetworks. Hence we also compare with randomly initialized models using random prompts. It can be observed that the neurons in random models are predictive to some extent, but their predictivities are far below the neurons in pre-trained models. These results imply that the skill neurons are generated in pre-training, and prompt tuning only serves as an effective tool to observe the specificity of these neurons.
\begin{figure}[!t]
\centering
\includegraphics[width=0.48\textwidth]{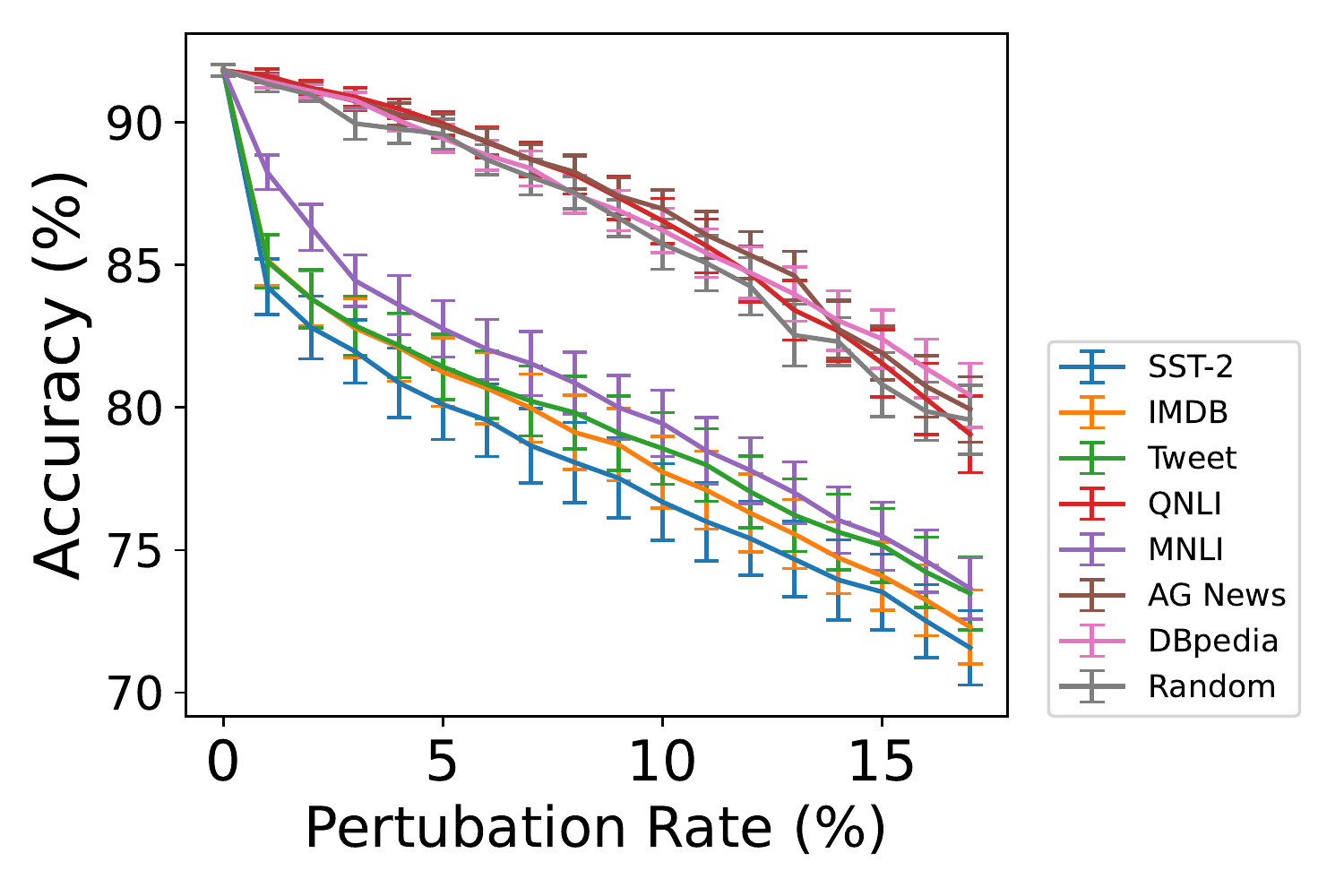}
\caption{BitFit accuracy on \texttt{IMDB} drops along with the neuron perturbation rate. Error bars indicate $\pm 1$ s.e.m. over $5$ random trials. The perturbations are conducted in predictivity orders obtained with prompt tuning.}
\label{fig:mask_trend_bias}
\end{figure}

\begin{figure}[!t]
\centering
\includegraphics[width=0.48\textwidth]{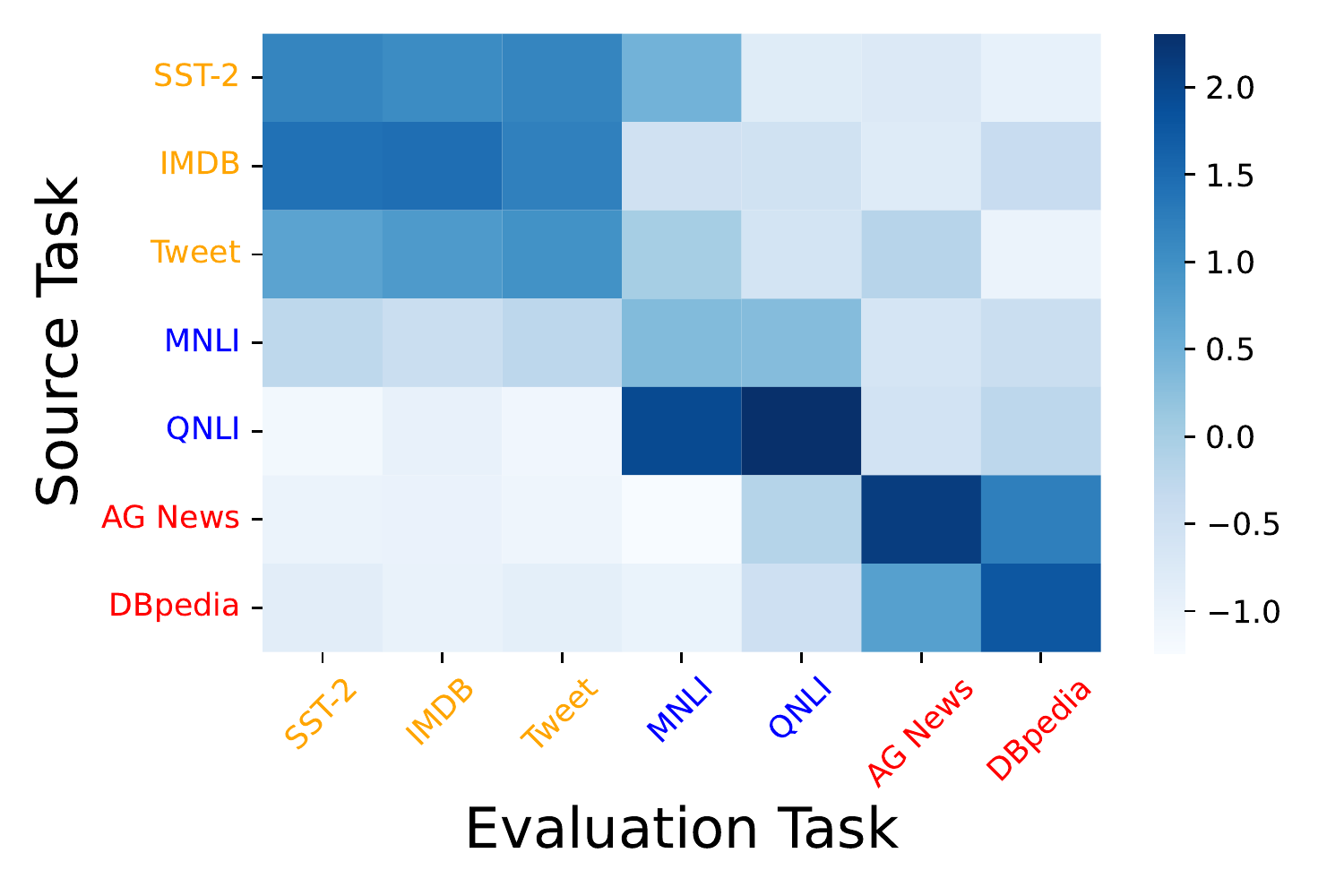}
\caption{Average neuronal importance over models trained with adapter-based tuning and BitFit.}
\label{fig:neuronal_importance_more}
\end{figure}
To provide stronger evidence, we explore whether the skill neurons found with prompt tuning are also important for other fine-tuning methods with different dynamics. We explore two parameter-efficient fine-tuning methods, including adapter-based tuning~\citep{houlsby2019parameter}, which only tunes the additional adapter layers plugged in Transformers, and BitFit~\citep{BenZaken2021BitFitSP}, which only tunes the bias vectors. The two tuning methods both keep neuron weights fixed, which ensures that the skill neurons are unchanged during tuning. BitFit model's performances on \texttt{IMDB} when neurons are perturbed in the descending orders of predictivities obtained with prompts are shown in \cref{fig:mask_trend_bias}, and the results for other tasks and adapter models are shown in \cref{app:perturbation}. We can see the highly-predictive skill neurons found with prompts are still crucial for models fine-tuned with other methods. To comprehensively show this effect, similar to \cref{sec:exp_task_specific}, we visualize the average neuronal importance over models trained with adapter-based tuning and BitFit in \cref{fig:neuronal_importance_more}. The skill neurons found with prompt tuning also exhibit task-specific importance, which again supports that skill neurons are generated in pre-training rather than manufactured by prompt tuning.

\section{Application}
\label{sec:application}
We further explore the applications of our skill neuron finding. We show two preliminary use cases: network pruning and transferability indicator.
\subsection{Network Pruning}
\looseness=-1 First, we apply our skill neuron finding to network pruning~\citep{anwar2017structured,dalvi2020analyzing}, which is to reduce memory cost and accelerate inference by removing redundant parameters in neural networks. Existing works have explored prune PLMs with weight magnitude~\citep{han2015learning,gordon2020compressing} and loss attribution~\citep{michel2019sixteen}. Here we explore prune PLMs by only keeping the top $2\%$ skill neurons active for each task and set the activations of the $98\%$ frozen neurons always as their baseline activations. Considering that the frozen neurons are fixed, we merge them into bias terms. We apply this pruning method to the top $9$ layers of \RBT and reduce it to $66.6\%$ of its original parameters. The performances of prompt tuning on pruned models and vanilla prompt tuning on the original model are shown in \cref{tab:prune}. Our pruning based on skill neurons generally performs comparably to vanilla prompt tuning and can achieve about $1.4$ inference speedup.

\begin{table}[t!]
\small
\centering
\begin{tabular}{lrrr}
\toprule
Task             & \multicolumn{1}{c}{\begin{tabular}[c]{@{}c@{}}Prompt\\ Tuning\end{tabular}} & \multicolumn{1}{c}{\begin{tabular}[c]{@{}c@{}}Pruned\\ Model\end{tabular}} & \multicolumn{1}{c}{Speedup} \\ \midrule
\texttt{SST-2}   & $91.8_{\pm0.5}$                                                             & $89.3_{\pm2.0}$                                                            & $1.34$                      \\
\texttt{IMDB}    & $91.6_{\pm0.5}$                                                             & $87.6_{\pm3.0}$                                                            & $1.34$                      \\
\texttt{Tweet}   & $70.0_{\pm0.2}$                                                             & $69.0_{\pm0.9}$                                                            & $1.34$                        \\
\texttt{MNLI}    & $76.8_{\pm1.8}$                                                             & $70.0_{\pm 1.1}$                                                              & $1.38$                        \\
\texttt{QNLI}    & $85.7_{\pm0.7}$                                                             & $81.0_{\pm1.0}$                                                            & $1.36$                      \\
\texttt{AG News} & $98.8_{\pm0.1}$                                                             & $99.8_{\pm0.1}$                                                            & $1.32$                      \\
\texttt{DBpedia} & $99.7_{\pm0.1}$                                                             & $99.0_{\pm0.1}$                                                            & $1.33$                      \\ \bottomrule
\end{tabular}
\caption{Accuracies (\%) on various tasks of vanilla prompt tuning and prompt tuning on pruned models, along with standard deviations over $5$ random trials. We also report the achieved inference speedups on the tasks. Speedups are evaluated on a single CPU since it is widely used for model inference~\citep{mittal2021survey}.}
\label{tab:prune}
\end{table}

\subsection{Transferability Indicator}
\looseness=-1 Previous works~\citep{su2021transferability,vu2021spot} explore improving prompt tuning with cross-task prompt transfer. \citet{su2021transferability} propose that the \textit{overlapping rate of activated neurons} ($\mathrm{ON}$) between soft prompts can serve as a prompt transferability indicator, which has good correlations with zero-shot prompt transferability and can help to qualify task similarities and improve prompt transfer. \citet{su2021transferability} take all neurons into $\mathrm{ON}$ calculation, but the redundant neurons without task-specific skills may bring noisy signals. Here we only take the top $20\%$ skill neurons of target tasks into the calculation. This improves the average Spearman's correlation between $\mathrm{ON}$ and prompt transferability over our tasks from $0.53$ to $0.71$.
\section{Related Work}
\label{sec:related}
\paragraph{Selective Neurons in Artificial Neural Networks}
There have long been findings about selective neurons in artificial neural networks. Many computer vision works~\citep{Coates2012EmergenceOO,Le2013BuildingHF,zeiler2014visualizing,Agrawal2014AnalyzingTP,Zhou2015ObjectDE,bau2020understanding} find that both supervised and unsupervised models can have units selectively respond to specific visual objects and concepts. \citet{Radford2017LearningTG} also find neurons corresponding to sentiments in unsupervised long short-term memory networks. Interestingly, there are similar selective neurons in human brains~\citep{barlow1972single,quiroga2005invariant}. The widespread emergence of these neuronal selectivities implies that there may be common learning mechanisms among intelligent systems, which is extremely worthwhile to explore in the future.

\citet{Bau2017NetworkDQ} and \citet{Mu2020CompositionalEO} find that selective neurons are more important, which is consistent with our findings. However, \citet{morcos2018importance} draw opposite conclusions. We discuss this with experiments in \cref{app:single_direction}.

\paragraph{Analyzing Pre-trained Transformers}
After the success of Transformer-based PLMs~\citep{devlin-etal-2019-bert,yang2019xlnet,raffel2020T5}, many efforts have been devoted to analyzing how PLMs work, such as probing the knowledge of PLMs~\citep{Liu2019LinguisticKA,hewitt2019structural,petroni-etal-2019-language} and understanding the behaviors of PLMs' parameters~\citep{voita-etal-2019-analyzing,clark-etal-2019-bert}. Among these, some works~\citep{dalvi2019one,durrani2020analyzing,antverg2021pitfalls} find that individual neurons capture linguistic properties, but they define neurons as dimensions in contextualized representations. Other works~\citep{suau2020finding,geva2021transformer,dai2021knowledge} study the same group of neurons as us and find that some neurons encode specific information like concepts, facts, and word patterns. Inspired by them, we study whether neurons encode high-level skills for handling tasks in this work and demonstrate that we can observe skill neurons with the help of prompts. We believe it is promising to explore whether and how skill neurons collaborate with the neurons encoding information in future works.

\section{Conclusion and Future Work}
\label{sec:conclusion}
In this paper, we find some special neurons in pre-trained Transformers whose activations on soft prompts are highly predictive of the task labels of inputs. We dub these neurons skill neurons and develop a method to find them via prompt tuning. With extensive experiments, we confirm that skill neurons encode task-specific skills required to handle these tasks and find empirical evidence showing that skill neurons are most likely generated in pre-training rather than fine-tuning. We also demonstrate some practical applications of our skill neuron finding. In the future, we will extend our prompt-based skill neuron finding method to more scenarios, such as covering non-classification tasks and other parameters in Transformers like attention heads. We will also explore more fundamental problems about skill neurons and the working mechanisms of PLMs, including how the skill neurons emerge in pre-training, as well as the relationships between skill neurons and neurons encoding specific information found in previous works.

\section*{Limitations}

Although we conducted extensive experiments, the exploration scope of this work has some limitations: (1) The experimental analyses are all based on \RBT{}. Whether the skill neuron phenomenon widely exists for other Transformer-based pre-trained language models is unclear and more explorations are needed to verify it. (2) The datasets used in our experiments are all English, which limits the linguistic features covered in our analyses, and the evaluation tasks are limited to classification tasks. We choose English just because of its rich resource. Although we intuitively believe the observed phenomena are not dependent on the English language, experiments on more diverse languages are needed in future works. (3) Following previous works~\citep{geva2021transformer,dai2021knowledge}, the analyzed neurons in our work all distribute in the feed-forward layers of Transformers. Deeper analyses may require considering other parameters like the attention heads. We encourage future works to address these limitations and get more comprehensive analysis results.

\section*{Acknowledgements}
This work is supported by the New Generation Artificial Intelligence of China (2020AAA0106501), the Institute for Guo Qiang, Tsinghua University (2019GQB0003), and Huawei Noah's Ark Lab. We thank anonymous reviewers for their suggestions.
\bibliography{custom}

\begin{thebibliography}{69}
\expandafter\ifx\csname natexlab\endcsname\relax\def\natexlab#1{#1}\fi

\bibitem[{Agrawal et~al.(2014)Agrawal, Girshick, and
  Malik}]{Agrawal2014AnalyzingTP}
Pulkit Agrawal, Ross~B. Girshick, and Jitendra Malik. 2014.
\newblock \href
  {https://www2.eecs.berkeley.edu/Research/Projects/CS/vision/papers/PulkitECCV2014.pdf}
  {Analyzing the performance of multilayer neural networks for object
  recognition}.
\newblock In \emph{Proceedings of ECCV}, pages 329--344.

\bibitem[{Antverg and Belinkov(2022)}]{antverg2021pitfalls}
Omer Antverg and Yonatan Belinkov. 2022.
\newblock \href {https://arxiv.org/pdf/2110.07483.pdf} {On the pitfalls of
  analyzing individual neurons in language models}.
\newblock In \emph{Proceedings of ICLR}.

\bibitem[{Anwar et~al.(2017)Anwar, Hwang, and Sung}]{anwar2017structured}
Sajid Anwar, Kyuyeon Hwang, and Wonyong Sung. 2017.
\newblock \href {https://doi.org/10.1145/3005348} {Structured pruning of deep
  convolutional neural networks}.
\newblock \emph{ACM Journal on Emerging Technologies in Computing Systems
  (JETC)}, 13(3):1--18.

\bibitem[{Arora et~al.(2018)Arora, Ge, Neyshabur, and
  Zhang}]{Arora2018StrongerGB}
Sanjeev Arora, Rong Ge, Behnam Neyshabur, and Yi~Zhang. 2018.
\newblock \href {http://proceedings.mlr.press/v80/arora18b/arora18b.pdf}
  {Stronger generalization bounds for deep nets via a compression approach}.
\newblock In \emph{Proceedings of ICML}, pages 254--263.

\bibitem[{Auer et~al.(2007)Auer, Bizer, Kobilarov, Lehmann, Cyganiak, and
  Ives}]{auer2007dbpedia}
S{\"o}ren Auer, Christian Bizer, Georgi Kobilarov, Jens Lehmann, Richard
  Cyganiak, and Zachary Ives. 2007.
\newblock \href {https://doi.org/10.1007/978-3-540-76298-0_52} {{DBpedia}: A
  nucleus for a web of open data}.
\newblock In \emph{Proceedings of ISWC/ASWC}, pages 722--735.

\bibitem[{Barbieri et~al.(2020)Barbieri, Camacho-Collados, Espinosa~Anke, and
  Neves}]{barbieri-etal-2020-Tweeteval}
Francesco Barbieri, Jose Camacho-Collados, Luis Espinosa~Anke, and Leonardo
  Neves. 2020.
\newblock \href {https://doi.org/10.18653/v1/2020.findings-emnlp.148}
  {{T}weet{E}val: Unified benchmark and comparative evaluation for tweet
  classification}.
\newblock In \emph{Findings of EMNLP}, pages 1644--1650.

\bibitem[{Barlow(1972)}]{barlow1972single}
Horace~B Barlow. 1972.
\newblock \href {https://doi.org/10.1068/p010371} {Single units and sensation:
  {A} neuron doctrine for perceptual psychology?}
\newblock \emph{Perception}, 1(4):371--394.

\bibitem[{Bau et~al.(2018)Bau, Belinkov, Sajjad, Durrani, Dalvi, and
  Glass}]{bau2018identifying}
Anthony Bau, Yonatan Belinkov, Hassan Sajjad, Nadir Durrani, Fahim Dalvi, and
  James Glass. 2018.
\newblock \href {https://openreview.net/pdf?id=H1z-PsR5KX} {Identifying and
  controlling important neurons in neural machine translation}.
\newblock In \emph{Proceedings of ICLR}.

\bibitem[{Bau et~al.(2017)Bau, Zhou, Khosla, Oliva, and
  Torralba}]{Bau2017NetworkDQ}
David Bau, Bolei Zhou, Aditya Khosla, Aude Oliva, and Antonio Torralba. 2017.
\newblock \href
  {https://openaccess.thecvf.com/content_cvpr_2017/papers/Bau_Network_Dissection_Quantifying_CVPR_2017_paper.pdf}
  {Network dissection: Quantifying interpretability of deep visual
  representations}.
\newblock \emph{Proceedings of CVPR}, pages 3319--3327.

\bibitem[{Bau et~al.(2020)Bau, Zhu, Strobelt, Lapedriza, Zhou, and
  Torralba}]{bau2020understanding}
David Bau, Jun-Yan Zhu, Hendrik Strobelt, Agata Lapedriza, Bolei Zhou, and
  Antonio Torralba. 2020.
\newblock \href {https://doi.org/10.1073/pnas.1907375117} {Understanding the
  role of individual units in a deep neural network}.
\newblock \emph{Proceedings of the National Academy of Sciences},
  117(48):30071--30078.

\bibitem[{Ben-Zaken et~al.(2022)Ben-Zaken, Ravfogel, and
  Goldberg}]{BenZaken2021BitFitSP}
Elad Ben-Zaken, Shauli Ravfogel, and Yoav Goldberg. 2022.
\newblock \href {https://arxiv.org/pdf/2106.10199.pdf} {{BitFit}: Simple
  parameter-efficient fine-tuning for transformer-based masked
  language-models}.
\newblock In \emph{Proceedings of ACL}.

\bibitem[{Brown et~al.(2020)Brown, Mann, Ryder, Subbiah, Kaplan, Dhariwal,
  Neelakantan, Shyam, Sastry, Askell, Agarwal, Herbert{-}Voss, Krueger,
  Henighan, Child, Ramesh, Ziegler, Wu, Winter, Hesse, Chen, Sigler, Litwin,
  Gray, Chess, Clark, Berner, McCandlish, Radford, Sutskever, and
  Amodei}]{brown2020GPT3}
Tom~B. Brown, Benjamin Mann, Nick Ryder, Melanie Subbiah, Jared Kaplan,
  Prafulla Dhariwal, Arvind Neelakantan, Pranav Shyam, Girish Sastry, Amanda
  Askell, Sandhini Agarwal, Ariel Herbert{-}Voss, Gretchen Krueger, Tom
  Henighan, Rewon Child, Aditya Ramesh, Daniel~M. Ziegler, Jeffrey Wu, Clemens
  Winter, Christopher Hesse, Mark Chen, Eric Sigler, Mateusz Litwin, Scott
  Gray, Benjamin Chess, Jack Clark, Christopher Berner, Sam McCandlish, Alec
  Radford, Ilya Sutskever, and Dario Amodei. 2020.
\newblock \href
  {https://proceedings.neurips.cc/paper/2020/hash/1457c0d6bfcb4967418bfb8ac142f64a-Abstract.html}
  {Language models are few-shot learners}.
\newblock In \emph{Proceedings of NeurIPS}, pages 1877--1901.

\bibitem[{Clark et~al.(2019)Clark, Khandelwal, Levy, and
  Manning}]{clark-etal-2019-bert}
Kevin Clark, Urvashi Khandelwal, Omer Levy, and Christopher~D. Manning. 2019.
\newblock \href {https://doi.org/10.18653/v1/W19-4828} {What does {BERT} look
  at? an analysis of {BERT}{'}s attention}.
\newblock In \emph{Proceedings of the 2019 ACL Workshop BlackboxNLP: Analyzing
  and Interpreting Neural Networks for NLP}, pages 276--286.

\bibitem[{Coates et~al.(2012)Coates, Karpathy, and Ng}]{Coates2012EmergenceOO}
Adam Coates, Andrej Karpathy, and A.~Ng. 2012.
\newblock \href
  {http://citeseerx.ist.psu.edu/viewdoc/download?doi=10.1.1.269.5798&rep=rep1&type=pdf}
  {Emergence of object-selective features in unsupervised feature learning}.
\newblock In \emph{Proceedings of NeurIPS}, pages 2681--2689.

\bibitem[{Dai et~al.(2021)Dai, Dong, Hao, Sui, and Wei}]{dai2021knowledge}
Damai Dai, Li~Dong, Yaru Hao, Zhifang Sui, and Furu Wei. 2021.
\newblock \href {https://arxiv.org/abs/2104.08696} {Knowledge neurons in
  pretrained transformers}.
\newblock \emph{arXiv preprint, arXiv:2104.08696}.

\bibitem[{Dalvi et~al.(2019)Dalvi, Durrani, Sajjad, Belinkov, Bau, and
  Glass}]{dalvi2019one}
Fahim Dalvi, Nadir Durrani, Hassan Sajjad, Yonatan Belinkov, Anthony Bau, and
  James Glass. 2019.
\newblock \href {https://doi.org/10.1609/aaai.v33i01.33016309} {What is one
  grain of sand in the desert? analyzing individual neurons in deep nlp
  models}.
\newblock In \emph{Proceedings of AAAI}, pages 6309--6317.

\bibitem[{Dalvi et~al.(2020)Dalvi, Sajjad, Durrani, and
  Belinkov}]{dalvi2020analyzing}
Fahim Dalvi, Hassan Sajjad, Nadir Durrani, and Yonatan Belinkov. 2020.
\newblock \href {https://aclanthology.org/2020.emnlp-main.398.pdf} {Analyzing
  redundancy in pretrained transformer models}.
\newblock In \emph{Proceedings of EMNLP}, pages 4908--4926.

\bibitem[{Devlin et~al.(2019)Devlin, Chang, Lee, and
  Toutanova}]{devlin-etal-2019-bert}
Jacob Devlin, Ming-Wei Chang, Kenton Lee, and Kristina Toutanova. 2019.
\newblock \href {https://doi.org/10.18653/v1/N19-1423} {{BERT}: {P}re-training
  of deep bidirectional transformers for language understanding}.
\newblock In \emph{Proceedings of NAACL-HLT}, pages 4171--4186.

\bibitem[{Dong et~al.(2021)Dong, Cordonnier, and Loukas}]{dong2021attention}
Yihe Dong, Jean-Baptiste Cordonnier, and Andreas Loukas. 2021.
\newblock \href {https://proceedings.mlr.press/v139/dong21a.html} {Attention is
  not all you need: pure attention loses rank doubly exponentially with depth}.
\newblock In \emph{Proceedings of ICML}, pages 2793--2803.

\bibitem[{Durrani et~al.(2020)Durrani, Sajjad, Dalvi, and
  Belinkov}]{durrani2020analyzing}
Nadir Durrani, Hassan Sajjad, Fahim Dalvi, and Yonatan Belinkov. 2020.
\newblock \href {https://aclanthology.org/2020.emnlp-main.395.pdf} {Analyzing
  individual neurons in pre-trained language models}.
\newblock In \emph{Proceedings of EMNLP}, pages 4865--4880.

\bibitem[{Geva et~al.(2021)Geva, Schuster, Berant, and
  Levy}]{geva2021transformer}
Mor Geva, Roei Schuster, Jonathan Berant, and Omer Levy. 2021.
\newblock \href {https://doi.org/10.18653/v1/2021.emnlp-main.446} {Transformer
  feed-forward layers are key-value memories}.
\newblock In \emph{Proceedings of EMNLP}, pages 5484--5495.

\bibitem[{Gordon et~al.(2020)Gordon, Duh, and Andrews}]{gordon2020compressing}
Mitchell~A Gordon, Kevin Duh, and Nicholas Andrews. 2020.
\newblock \href {https://arxiv.org/pdf/2002.08307.pdf} {Compressing bert:
  Studying the effects of weight pruning on transfer learning}.
\newblock \emph{arXiv preprint arXiv:2002.08307}.

\bibitem[{Han et~al.(2015)Han, Pool, Tran, and Dally}]{han2015learning}
Song Han, Jeff Pool, John Tran, and William Dally. 2015.
\newblock \href {10.5555/2969239.2969366} {Learning both weights and
  connections for efficient neural network}.
\newblock In \emph{Proceedings of NeurIPS}, pages 1135--1143.

\bibitem[{Han et~al.(2021)Han, Zhang, Ding, Gu, Liu, Huo, Qiu, Zhang, Han,
  Huang et~al.}]{han2021pre}
Xu~Han, Zhengyan Zhang, Ning Ding, Yuxian Gu, Xiao Liu, Yuqi Huo, Jiezhong Qiu,
  Liang Zhang, Wentao Han, Minlie Huang, et~al. 2021.
\newblock \href {https://doi.org/10.1016/j.aiopen.2021.08.002} {Pre-trained
  models: Past, present and future}.
\newblock \emph{AI Open}, pages 225--250.

\bibitem[{Hennigen et~al.(2020)Hennigen, Williams, and
  Cotterell}]{hennigen2020intrinsic}
Lucas~Torroba Hennigen, Adina Williams, and Ryan Cotterell. 2020.
\newblock \href {https://aclanthology.org/2020.emnlp-main.15.pdf} {Intrinsic
  probing through dimension selection}.
\newblock In \emph{Proceedings of EMNLP}, pages 197--216.

\bibitem[{Hewitt and Manning(2019)}]{hewitt2019structural}
John Hewitt and Christopher~D Manning. 2019.
\newblock \href {https://doi.org/10.18653/v1/n19-1419} {A structural probe for
  finding syntax in word representations}.
\newblock In \emph{Proceedings of NACCL-HLT}, pages 4129--4138.

\bibitem[{Houlsby et~al.(2019)Houlsby, Giurgiu, Jastrzebski, Morrone,
  de~Laroussilhe, Gesmundo, Attariyan, and Gelly}]{houlsby2019parameter}
Neil Houlsby, Andrei Giurgiu, Stanislaw Jastrzebski, Bruna Morrone, Quentin
  de~Laroussilhe, Andrea Gesmundo, Mona Attariyan, and Sylvain Gelly. 2019.
\newblock \href {http://proceedings.mlr.press/v97/houlsby19a.html}
  {Parameter-efficient transfer learning for {NLP}}.
\newblock In \emph{Proceedings of ICML}, pages 2790--2799.

\bibitem[{Jiang et~al.(2020)Jiang, Xu, Araki, and
  Neubig}]{jiang-etal-2020-know}
Zhengbao Jiang, Frank~F. Xu, Jun Araki, and Graham Neubig. 2020.
\newblock \href {https://doi.org/10.1162/tacl_a_00324} {How can we know what
  language models know?}
\newblock \emph{Transactions of the Association for Computational Linguistics},
  8:423--438.

\bibitem[{Karpathy et~al.(2015)Karpathy, Johnson, and
  Fei-Fei}]{karpathy2015visualizing}
Andrej Karpathy, Justin Johnson, and Li~Fei-Fei. 2015.
\newblock \href {https://arxiv.org/pdf/1506.02078.pdf} {Visualizing and
  understanding recurrent networks}.
\newblock \emph{arXiv preprint arXiv:1506.02078}, pages 818--833.

\bibitem[{Kingma and Ba(2015)}]{Kingma2015AdamAM}
Diederik~P. Kingma and Jimmy Ba. 2015.
\newblock \href {https://arxiv.org/abs/1412.6980} {Adam: A method for
  stochastic optimization}.
\newblock In \emph{Proceedings of ICLR}.

\bibitem[{Le et~al.(2013)Le, Ranzato, Monga, Devin, Corrado, Chen, Dean, and
  Ng}]{Le2013BuildingHF}
Quoc~V. Le, Marc'Aurelio Ranzato, Rajat Monga, Matthieu Devin, Gregory~S.
  Corrado, Kai Chen, Jeffrey Dean, and A.~Ng. 2013.
\newblock \href {https://doi.org/10.1109/ICASSP.2013.6639343} {Building
  high-level features using large scale unsupervised learning}.
\newblock In \emph{Proceedings of ICASSP}, pages 8595--8598.

\bibitem[{Lester et~al.(2021)Lester, Al-Rfou, and Constant}]{lester2021power}
Brian Lester, Rami Al-Rfou, and Noah Constant. 2021.
\newblock \href {https://doi.org/10.18653/v1/2021.emnlp-main.243} {The power of
  scale for parameter-efficient prompt tuning}.
\newblock In \emph{Proceedings of EMNLP}, pages 3045--3059.

\bibitem[{Lhoest et~al.(2021)Lhoest, Villanova~del Moral, Jernite, Thakur, von
  Platen, Patil, Chaumond, Drame, Plu, Tunstall, Davison, {\v{S}}a{\v{s}}ko,
  Chhablani, Malik, Brandeis, Le~Scao, Sanh, Xu, Patry, McMillan-Major, Schmid,
  Gugger, Delangue, Matussi{\`e}re, Debut, Bekman, Cistac, Goehringer, Mustar,
  Lagunas, Rush, and Wolf}]{lhoest-etal-2021-datasets}
Quentin Lhoest, Albert Villanova~del Moral, Yacine Jernite, Abhishek Thakur,
  Patrick von Platen, Suraj Patil, Julien Chaumond, Mariama Drame, Julien Plu,
  Lewis Tunstall, Joe Davison, Mario {\v{S}}a{\v{s}}ko, Gunjan Chhablani,
  Bhavitvya Malik, Simon Brandeis, Teven Le~Scao, Victor Sanh, Canwen Xu,
  Nicolas Patry, Angelina McMillan-Major, Philipp Schmid, Sylvain Gugger,
  Cl{\'e}ment Delangue, Th{\'e}o Matussi{\`e}re, Lysandre Debut, Stas Bekman,
  Pierric Cistac, Thibault Goehringer, Victor Mustar, Fran{\c{c}}ois Lagunas,
  Alexander Rush, and Thomas Wolf. 2021.
\newblock \href {https://aclanthology.org/2021.emnlp-demo.21} {Datasets: A
  community library for natural language processing}.
\newblock In \emph{Proceedings of EMNLP}, pages 175--184.

\bibitem[{Li and Liang(2021)}]{li-liang-2021-prefix}
Xiang~Lisa Li and Percy Liang. 2021.
\newblock \href {https://doi.org/10.18653/v1/2021.acl-long.353} {Prefix-tuning:
  Optimizing continuous prompts for generation}.
\newblock In \emph{Proceedings of ACL}, pages 4582--4597.

\bibitem[{Liu et~al.(2019{\natexlab{a}})Liu, Gardner, Belinkov, Peters, and
  Smith}]{Liu2019LinguisticKA}
Nelson~F. Liu, Matt Gardner, Yonatan Belinkov, Matthew~E. Peters, and Noah~A.
  Smith. 2019{\natexlab{a}}.
\newblock \href {https://doi.org/10.18653/v1/N19-1112} {Linguistic knowledge
  and transferability of contextual representations}.
\newblock In \emph{Proceedings of NAACL-HLT}, pages 1073--1094.

\bibitem[{Liu et~al.(2022)Liu, Ji, Fu, Du, Yang, and Tang}]{liu2021ptuningv2}
Xiao Liu, Kaixuan Ji, Yicheng Fu, Zhengxiao Du, Zhilin Yang, and Jie Tang.
  2022.
\newblock \href {https://arxiv.org/abs/2110.07602} {{P-Tuning v2}: Prompt
  tuning can be comparable to fine-tuning universally across scales and tasks}.
\newblock In \emph{Proceedings of ACL}.

\bibitem[{Liu et~al.(2019{\natexlab{b}})Liu, Ott, Goyal, Du, Joshi, Chen, Levy,
  Lewis, Zettlemoyer, and Stoyanov}]{liu2019roberta}
Yinhan Liu, Myle Ott, Naman Goyal, Jingfei Du, Mandar Joshi, Danqi Chen, Omer
  Levy, Mike Lewis, Luke Zettlemoyer, and Veselin Stoyanov. 2019{\natexlab{b}}.
\newblock \href {https://arxiv.org/pdf/1907.11692.pdf} {{RoBERTa}: A robustly
  optimized bert pretraining approach}.
\newblock \emph{arXiv preprint arXiv:1907,11692}.

\bibitem[{Maas et~al.(2011)Maas, Daly, Pham, Huang, Ng, and
  Potts}]{maas-EtAl:2011:ACL-HLT2011}
Andrew~L. Maas, Raymond~E. Daly, Peter~T. Pham, Dan Huang, Andrew~Y. Ng, and
  Christopher Potts. 2011.
\newblock \href {https://aclanthology.org/P11-1015} {Learning word vectors for
  sentiment analysis}.
\newblock In \emph{Proceedings of ACL-HLT}, pages 142--150.

\bibitem[{Malach et~al.(2020)Malach, Yehudai, Shalev-shwartz, and
  Shamir}]{malach2020proving}
Eran Malach, Gilad Yehudai, Shai Shalev-shwartz, and Ohad Shamir. 2020.
\newblock Proving the lottery ticket hypothesis: Pruning is all you need.
\newblock In \emph{Proceedings of ICML}.

\bibitem[{Michel et~al.(2019)Michel, Levy, and Neubig}]{michel2019sixteen}
Paul Michel, Omer Levy, and Graham Neubig. 2019.
\newblock \href
  {https://proceedings.neurips.cc/paper/2019/file/2c601ad9d2ff9bc8b282670cdd54f69f-Paper.pdf}
  {Are sixteen heads really better than one?}
\newblock In \emph{Proceedings of NeurIPS}, pages 14014--14024.

\bibitem[{Mitchell et~al.(2021)Mitchell, Lin, Bosselut, Finn, and
  Manning}]{mitchell2021fast}
Eric Mitchell, Charles Lin, Antoine Bosselut, Chelsea Finn, and Christopher~D
  Manning. 2021.
\newblock \href {https://openreview.net/pdf?id=0DcZxeWfOPt} {Fast model editing
  at scale}.
\newblock In \emph{Proceedings of ICLR}.

\bibitem[{Mittal et~al.(2021)Mittal, Rajput, and Subramoney}]{mittal2021survey}
Sparsh Mittal, Poonam Rajput, and Sreenivas Subramoney. 2021.
\newblock \href {https://doi.org/10.1109/TNNLS.2021.3071762} {A survey of deep
  learning on {CPU}s: opportunities and co-optimizations}.
\newblock \emph{IEEE Transactions on Neural Networks and Learning Systems},
  pages 1--21.

\bibitem[{Morcos et~al.(2018)Morcos, Barrett, Rabinowitz, and
  Botvinick}]{morcos2018importance}
Ari~S Morcos, David~GT Barrett, Neil~C Rabinowitz, and Matthew Botvinick. 2018.
\newblock \href {https://arxiv.org/pdf/1803.06959.pdf} {On the importance of
  single directions for generalization}.
\newblock In \emph{Proceedings of ICLR}.

\bibitem[{Mu and Andreas(2020)}]{Mu2020CompositionalEO}
Jesse Mu and Jacob Andreas. 2020.
\newblock \href
  {https://proceedings.neurips.cc/paper/2020/file/c74956ffb38ba48ed6ce977af6727275-Paper.pdf}
  {Compositional explanations of neurons}.
\newblock In \emph{Proceedings of NeurIPS}, pages 17153--17163.

\bibitem[{Petroni et~al.(2019)Petroni, Rockt{\"a}schel, Riedel, Lewis, Bakhtin,
  Wu, and Miller}]{petroni-etal-2019-language}
Fabio Petroni, Tim Rockt{\"a}schel, Sebastian Riedel, Patrick Lewis, Anton
  Bakhtin, Yuxiang Wu, and Alexander Miller. 2019.
\newblock \href {https://doi.org/10.18653/v1/D19-1250} {Language models as
  knowledge bases?}
\newblock In \emph{Proceedings of EMNLP-IJCNLP}, pages 2463--2473.

\bibitem[{Press et~al.(2020)Press, Smith, and Levy}]{press-etal-2020-improving}
Ofir Press, Noah~A. Smith, and Omer Levy. 2020.
\newblock \href {https://doi.org/10.18653/v1/2020.acl-main.270} {Improving
  transformer models by reordering their sublayers}.
\newblock In \emph{Proceedings of ACL}, pages 2996--3005.

\bibitem[{Qin and Eisner(2021)}]{qin-eisner-2021-learning}
Guanghui Qin and Jason Eisner. 2021.
\newblock \href {https://doi.org/10.18653/v1/2021.naacl-main.410} {Learning how
  to ask: Querying {LM}s with mixtures of soft prompts}.
\newblock In \emph{Proceedings of NAACL-HLT}, pages 5203--5212.

\bibitem[{Quiroga et~al.(2005)Quiroga, Reddy, Kreiman, Koch, and
  Fried}]{quiroga2005invariant}
R~Quian Quiroga, Leila Reddy, Gabriel Kreiman, Christof Koch, and Itzhak Fried.
  2005.
\newblock \href {https://doi.org/10.1038/nature03687} {Invariant visual
  representation by single neurons in the human brain}.
\newblock \emph{Nature}, 435(7045):1102--1107.

\bibitem[{Radford et~al.(2017)Radford, J{\'o}zefowicz, and
  Sutskever}]{Radford2017LearningTG}
Alec Radford, Rafal J{\'o}zefowicz, and Ilya Sutskever. 2017.
\newblock \href {https://arxiv.org/pdf/1704.01444.pdf} {Learning to generate
  reviews and discovering sentiment}.
\newblock \emph{arXiv preprint arXiv:1704.01444}.

\bibitem[{Raffel et~al.(2020)Raffel, Shazeer, Roberts, Lee, Narang, Matena,
  Zhou, Li, and Liu}]{raffel2020T5}
Colin Raffel, Noam Shazeer, Adam Roberts, Katherine Lee, Sharan Narang, Michael
  Matena, Yanqi Zhou, Wei Li, and Peter~J Liu. 2020.
\newblock \href {https://jmlr.org/papers/v21/20-074.html} {Exploring the limits
  of transfer learning with a unified text-to-text transformer}.
\newblock \emph{Journal of Machine Learning Research}, 21:1--67.

\bibitem[{Rosenthal et~al.(2017)Rosenthal, Farra, and
  Nakov}]{rosenthal-etal-2017-semeval}
Sara Rosenthal, Noura Farra, and Preslav Nakov. 2017.
\newblock \href {https://doi.org/10.18653/v1/S17-2088} {{S}em{E}val-2017 task
  4: Sentiment analysis in {T}witter}.
\newblock In \emph{Proceedings of {S}em{E}val}, pages 502--518.

\bibitem[{Rudy et~al.(2011)Rudy, Fishell, Lee, and
  Hjerling-Leffler}]{rudy2011three}
Bernardo Rudy, Gordon Fishell, SooHyun Lee, and Jens Hjerling-Leffler. 2011.
\newblock \href {https://doi.org/10.1002/dneu.20853} {Three groups of
  interneurons account for nearly 100\% of neocortical gabaergic neurons}.
\newblock \emph{Developmental neurobiology}, 71(1):45--61.

\bibitem[{Schick and Sch{\"u}tze(2021)}]{schick-schutze-2021-exploiting}
Timo Schick and Hinrich Sch{\"u}tze. 2021.
\newblock \href {https://aclanthology.org/2021.eacl-main.20} {Exploiting
  cloze-questions for few-shot text classification and natural language
  inference}.
\newblock In \emph{Proceedings of EACL}, pages 255--269.

\bibitem[{Socher et~al.(2013)Socher, Perelygin, Wu, Chuang, Manning, Ng, and
  Potts}]{socher-etal-2013-recursive}
Richard Socher, Alex Perelygin, Jean Wu, Jason Chuang, Christopher~D. Manning,
  Andrew Ng, and Christopher Potts. 2013.
\newblock \href {https://aclanthology.org/D13-1170} {Recursive deep models for
  semantic compositionality over a sentiment treebank}.
\newblock In \emph{Proceedings of EMNLP}, pages 1631--1642.

\bibitem[{Spearman(1987)}]{spearman1987proof}
Charles Spearman. 1987.
\newblock \href {https://www.jstor.org/stable/1422689} {The proof and
  measurement of association between two things}.
\newblock In \emph{Proceedings of AJP}, 3/4, pages 441--471.

\bibitem[{Su et~al.(2021)Su, Wang, Qin, Chan, Lin, Liu, Li, Li, Hou, Sun
  et~al.}]{su2021transferability}
Yusheng Su, Xiaozhi Wang, Yujia Qin, Chi-Min Chan, Yankai Lin, Zhiyuan Liu,
  Peng Li, Juanzi Li, Lei Hou, Maosong Sun, et~al. 2021.
\newblock \href {https://arxiv.org/pdf/2111.06719.pdf} {On transferability of
  prompt tuning for natural language understanding}.
\newblock \emph{arXiv preprint arXiv:2111.06719}.

\bibitem[{Suau et~al.(2020)Suau, Zappella, and Apostoloff}]{suau2020finding}
Xavier Suau, Luca Zappella, and Nicholas Apostoloff. 2020.
\newblock \href {https://arxiv.org/pdf/2005.07647.pdf} {Finding experts in
  transformer models}.
\newblock \emph{arXiv preprint arXiv:2005.07647}.

\bibitem[{Vaswani et~al.(2017)Vaswani, Shazeer, Parmar, Uszkoreit, Jones,
  Gomez, Kaiser, and Polosukhin}]{vaswani2017attention}
Ashish Vaswani, Noam Shazeer, Niki Parmar, Jakob Uszkoreit, Llion Jones,
  Aidan~N. Gomez, Lukasz Kaiser, and Illia Polosukhin. 2017.
\newblock \href
  {https://proceedings.neurips.cc/paper/2017/hash/3f5ee243547dee91fbd053c1c4a845aa-Abstract.html}
  {Attention is all you need}.
\newblock In \emph{Proceedings of NeurIPS}, pages 5998--6008.

\bibitem[{Voita et~al.(2019)Voita, Talbot, Moiseev, Sennrich, and
  Titov}]{voita-etal-2019-analyzing}
Elena Voita, David Talbot, Fedor Moiseev, Rico Sennrich, and Ivan Titov. 2019.
\newblock \href {https://doi.org/10.18653/v1/P19-1580} {Analyzing multi-head
  self-attention: Specialized heads do the heavy lifting, the rest can be
  pruned}.
\newblock In \emph{Proceedings of NAACL}, pages 5797--5808.

\bibitem[{Vu et~al.(2021)Vu, Lester, Constant, Al-Rfou, and Cer}]{vu2021spot}
Tu~Vu, Brian Lester, Noah Constant, Rami Al-Rfou, and Daniel Cer. 2021.
\newblock \href {https://arxiv.org/pdf/2110.07904.pdf} {Spot: Better frozen
  model adaptation through soft prompt transfer}.
\newblock \emph{arXiv preprint arxiv:2110.07904}.

\bibitem[{Wang et~al.(2019)Wang, Singh, Michael, Hill, Levy, and
  Bowman}]{wang2018glue}
Alex Wang, Amanpreet Singh, Julian Michael, Felix Hill, Omer Levy, and
  Samuel~R. Bowman. 2019.
\newblock \href {https://openreview.net/pdf?id=rJ4km2R5t7} {{GLUE:} {A}
  multi-task benchmark and analysis platform for natural language
  understanding}.
\newblock In \emph{Proceedings of ICLR}.

\bibitem[{Williams et~al.(2018)Williams, Nangia, and
  Bowman}]{williams-etal-2018-broad}
Adina Williams, Nikita Nangia, and Samuel Bowman. 2018.
\newblock \href {https://doi.org/10.18653/v1/N18-1101} {A broad-coverage
  challenge corpus for sentence understanding through inference}.
\newblock In \emph{Proceedings of NAACL-HLT}, pages 1112--1122.

\bibitem[{Wolf et~al.(2020)Wolf, Debut, Sanh, Chaumond, Delangue, Moi, Cistac,
  Rault, Louf, Funtowicz et~al.}]{wolf2020transformers}
Thomas Wolf, Lysandre Debut, Victor Sanh, Julien Chaumond, Clement Delangue,
  Anthony Moi, Pierric Cistac, Tim Rault, R{\'e}mi Louf, Morgan Funtowicz,
  et~al. 2020.
\newblock \href {https://arxiv.org/pdf/1910.03771.pdf} {Transformers:
  State-of-the-art natural language processing}.
\newblock In \emph{Proceedings of EMNLP}, pages 38--45.

\bibitem[{Yang et~al.(2019)Yang, Dai, Yang, Carbonell, Salakhutdinov, and
  Le}]{yang2019xlnet}
Zhilin Yang, Zihang Dai, Yiming Yang, Jaime~G. Carbonell, Ruslan Salakhutdinov,
  and Quoc~V. Le. 2019.
\newblock \href
  {https://proceedings.neurips.cc/paper/2019/hash/dc6a7e655d7e5840e66733e9ee67cc69-Abstract.html}
  {Xlnet: Generalized autoregressive pretraining for language understanding}.
\newblock In \emph{Proceedings of NeurIPS}, pages 5754--5764.

\bibitem[{Zeiler and Fergus(2014)}]{zeiler2014visualizing}
Matthew~D. Zeiler and Rob Fergus. 2014.
\newblock \href
  {https://link.springer.com/content/pdf/10.1007/978-3-319-10590-1_53.pdf}
  {Visualizing and understanding convolutional networks}.
\newblock In \emph{Proceedings of ECCV}.

\bibitem[{Zhang et~al.(2015)Zhang, Zhao, and LeCun}]{zhang2015character}
Xiang Zhang, Junbo Zhao, and Yann LeCun. 2015.
\newblock \href
  {https://proceedings.neurips.cc/paper/2015/file/250cf8b51c773f3f8dc8b4be867a9a02-Paper.pdf}
  {Character-level convolutional networks for text classification}.
\newblock In \emph{Proceedings of NeurIPS}, pages 649--657.

\bibitem[{Zhang et~al.(2021)Zhang, Lin, Liu, Li, Sun, and
  Zhou}]{zhang2021moefication}
Zhengyan Zhang, Yankai Lin, Zhiyuan Liu, Peng Li, Maosong Sun, and Jie Zhou.
  2021.
\newblock \href {https://arxiv.org/pdf/2110.01786} {Moefication: Conditional
  computation of transformer models for efficient inference}.
\newblock \emph{arXiv preprint arXiv:2110.01786}.

\bibitem[{Zhong et~al.(2021)Zhong, Friedman, and
  Chen}]{zhong-etal-2021-factual}
Zexuan Zhong, Dan Friedman, and Danqi Chen. 2021.
\newblock \href {https://doi.org/10.18653/v1/2021.naacl-main.398} {Factual
  probing is [{MASK}]: Learning vs. learning to recall}.
\newblock In \emph{Proceedings of NAACL}, pages 5017--5033.

\bibitem[{Zhou et~al.(2015)Zhou, Khosla, Lapedriza, Oliva, and
  Torralba}]{Zhou2015ObjectDE}
Bolei Zhou, Aditya Khosla, {\`A}gata Lapedriza, Aude Oliva, and Antonio
  Torralba. 2015.
\newblock \href {https://arxiv.org/pdf/1412.6856.pdf} {Object detectors emerge
  in deep scene cnns}.
\newblock In \emph{Proceedings of ICLR}.

\end{thebibliography}
\bibliographystyle{acl_natbib}

\clearpage
\appendix

\section*{Appendices}
\section{Details about Investigated Tasks}
\label{app:task}
In experiments, we use $7$ established public English NLP datasets, which are licensed and intended for research use. These datasets are all created with public texts, and we believe they do not involve personal information and are well anonymized. The details about the datasets are as follows:

\subsection{Sentiment Analysis}
\textbf{SST-2}~\citep{socher-etal-2013-recursive} requires to classify the sentiments expressed in movie reviews into \textsc{Positive} and \textsc{Negative} sentiments.

\noindent\textbf{IMDB}~\citep{maas-EtAl:2011:ACL-HLT2011} requires to classify the sentiments expressed in reviews from the Internet Movie Database\footnote{\url{https://www.imdb.com}} into \textsc{Positive} and \textsc{Negative} sentiments.

\noindent\textbf{TweetEval}~\citep{barbieri-etal-2020-Tweeteval} is a collection of $7$ Twitter-specific classification tasks. Here we use its sentiment analysis subtask, which is originally from SemEval 2017 Task 4~\citep{rosenthal-etal-2017-semeval}. It requires to recognize if a tweet is \textsc{Positive}, \textsc{Negative} or \textsc{Neutral}. We decompose it to two subtasks: \textsc{Positive} vs. \textsc{Negative}, and \textsc{Neural} vs. \textsc{Non-neutral}.

\subsection{Natural Language Inference}
\textbf{MNLI}~\citep{williams-etal-2018-broad} requires to recognize the relationship between sentence pairs as \textsc{Entailment}, \textsc{Neutral} and \textsc{Contradiction}. We decompose it to two subtasks: \textsc{Entailment} vs. \textsc{Contradiction}, and \textsc{Neural} vs. \textsc{Non-neutral}.

\noindent\textbf{QNLI}~\citep{wang2018glue} requires to classify whether a context sentence contains the answer to a question.

\subsection{Topic Classification}
\textbf{AG News}~\citep{zhang2015character} requires to classify the $4$ topics of news articles in the AG's corpus\footnote{\url{http://groups.di.unipi.it/~gulli/AG_corpus_of_news_articles.html}}.

\noindent\textbf{DBpedia}~\citep{zhang2015character} requires to classify the $14$ topics of articles in DBpedia~\citep{auer2007dbpedia}.

Since recognizing different topics requires essentially different skills, we use the only two similar labels of the two tasks. They are \textsc{Business} and \textsc{Sports} in \texttt{AG News}, and \textsc{Company} and \textsc{Athlete} in \texttt{DBpedia}.

We obtain the datasets from Huggingface's dataset platform~\citep{lhoest-etal-2021-datasets}. For the datasets included in the GLUE collection~\citep{wang2018glue}, since we cannot get their test set, we use the released validation set as our test set, $80\%$ random samples from the original training set as our training set, and the other $20\%$ samples as our validation set. The detailed data statistics are shown in \cref{tab:app_data_stat}.

\begin{table}[t!]
\centering
\small
\begin{tabular}{lrrr}
\toprule
Task             & \multicolumn{1}{c}{Training} & \multicolumn{1}{c}{Validation} & \multicolumn{1}{c}{Test} \\ \midrule
\texttt{SST-2}   & $53,879$                           & $13,470$                             & $872$                       \\
\texttt{IMDB}    & $20,000$                           & $5,000$                             & $25,000$                       \\
\texttt{Tweet}   & $45,615$                           & $2,000$                             & $12,284$                       \\
\texttt{MNLI}    & $314,161$                           & $78,541$                             & $9,815$                       \\
\texttt{QNLI}    & $83,794$                           & $20,949$                             & $5,463$                       \\
\texttt{AG News} & $47,966$                           & $12,034$                             & $3,800$                       \\
\texttt{DBpedia} & $63,899$                           & $16,100$                             & $9,999$                       \\ \bottomrule
\end{tabular}
\caption{Data statistics of the $7$ used datasets.}
\label{tab:app_data_stat}
\end{table}

\section{Implementations Details}
\label{app:pt}

We implement the prompt tuning method introduced in \cref{sec:pre_PT} with $l = 127$ soft prompts. We randomly initialize each soft prompt using a normal distribution with the standard deviation as $0.03$. We then train the model using Adam~\citep{Kingma2015AdamAM} as the optimizer. We set the learning rate as $0.001$ and the batch size as $8$. We do the evaluation on the validation set every $2,000$ iterations and early stop the training if the validation accuracy does not rise for $6$ times. We use label words \texttt{Negative}, \texttt{Positive} for binary classification tasks and \texttt{Negative, Neutral, Positive} for multi-class classification tasks. For the random label words experiment in \cref{sec:word_select}, we uniformly sample the label words from the vocabulary of RoBERTa~\citep{liu2019roberta}.

We conduct all experiments on \RBT model, which has $110$M parameters, and we use Huggingface's Transformers library~\citep{wolf2020transformers} to implement the experiments. We run the experiments on NVIDIA GeForce RTX 2080 Ti and NVIDIA GeForce RTX 3090 GPUs, and it takes about $1000$ GPU hours.
 
\section{More Predictivity Distributions}
\label{app:acc_dist}
\begin{figure*}[!t]
\subcapraggedrighttrue
\subcaphangtrue
    \centering
    \subfigure[\texttt{SST-2}]{
	    \includegraphics[width=0.48\textwidth]{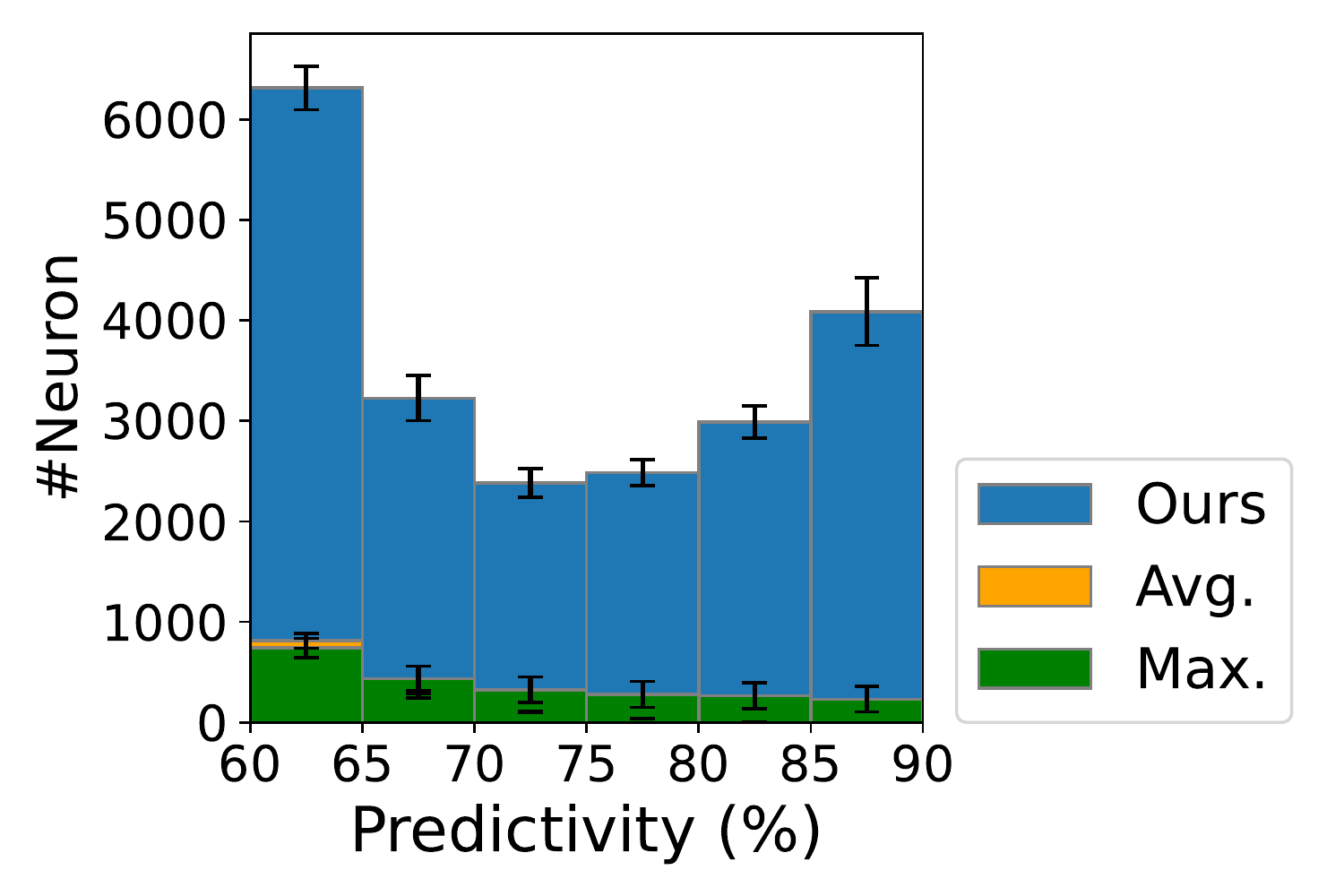}
    }
    \subfigure[\texttt{QNLI}]{
	    \includegraphics[width=0.48\textwidth]{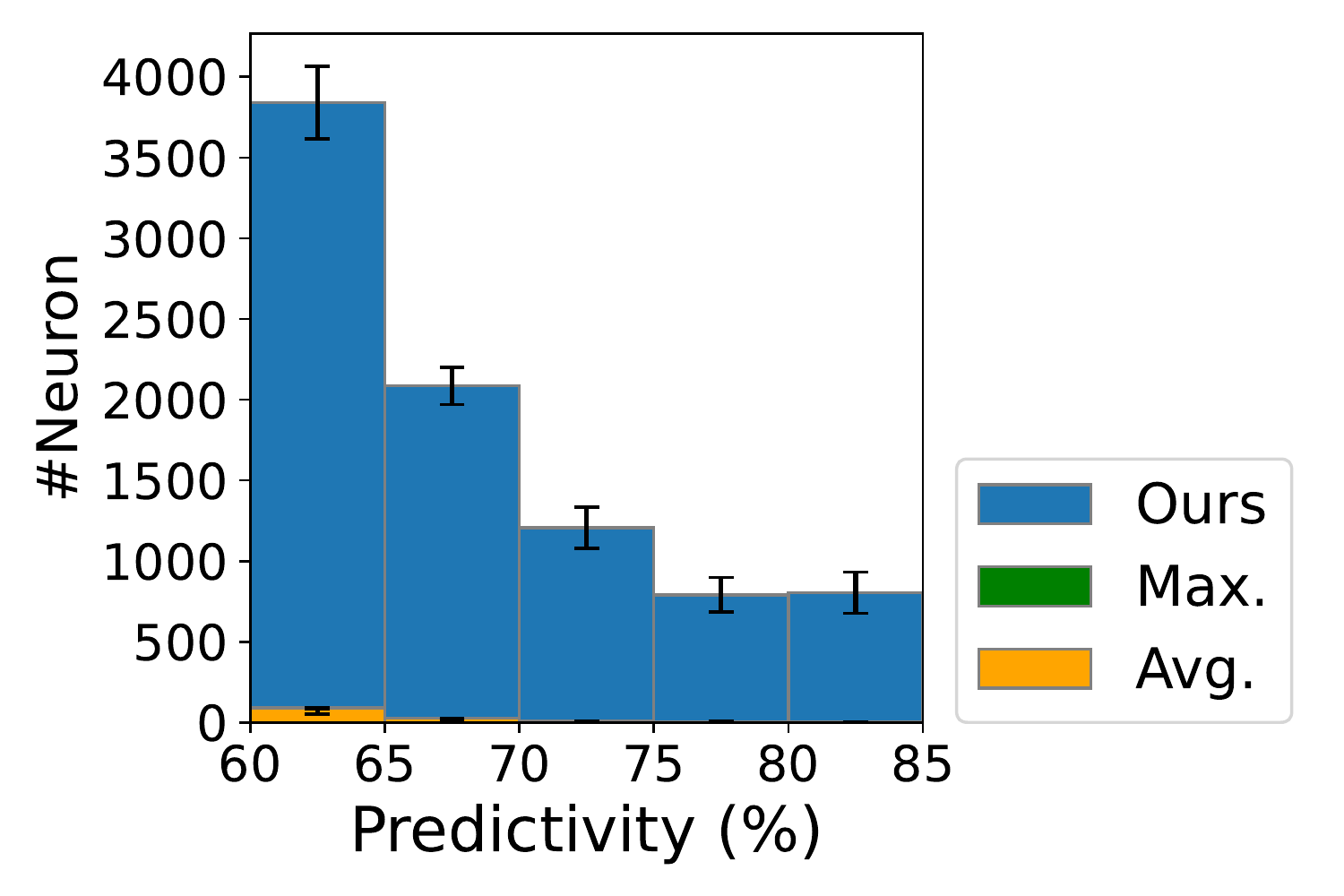}
    }
    \subfigure[\texttt{DBpedia}]{
	    \includegraphics[width=0.48\textwidth]{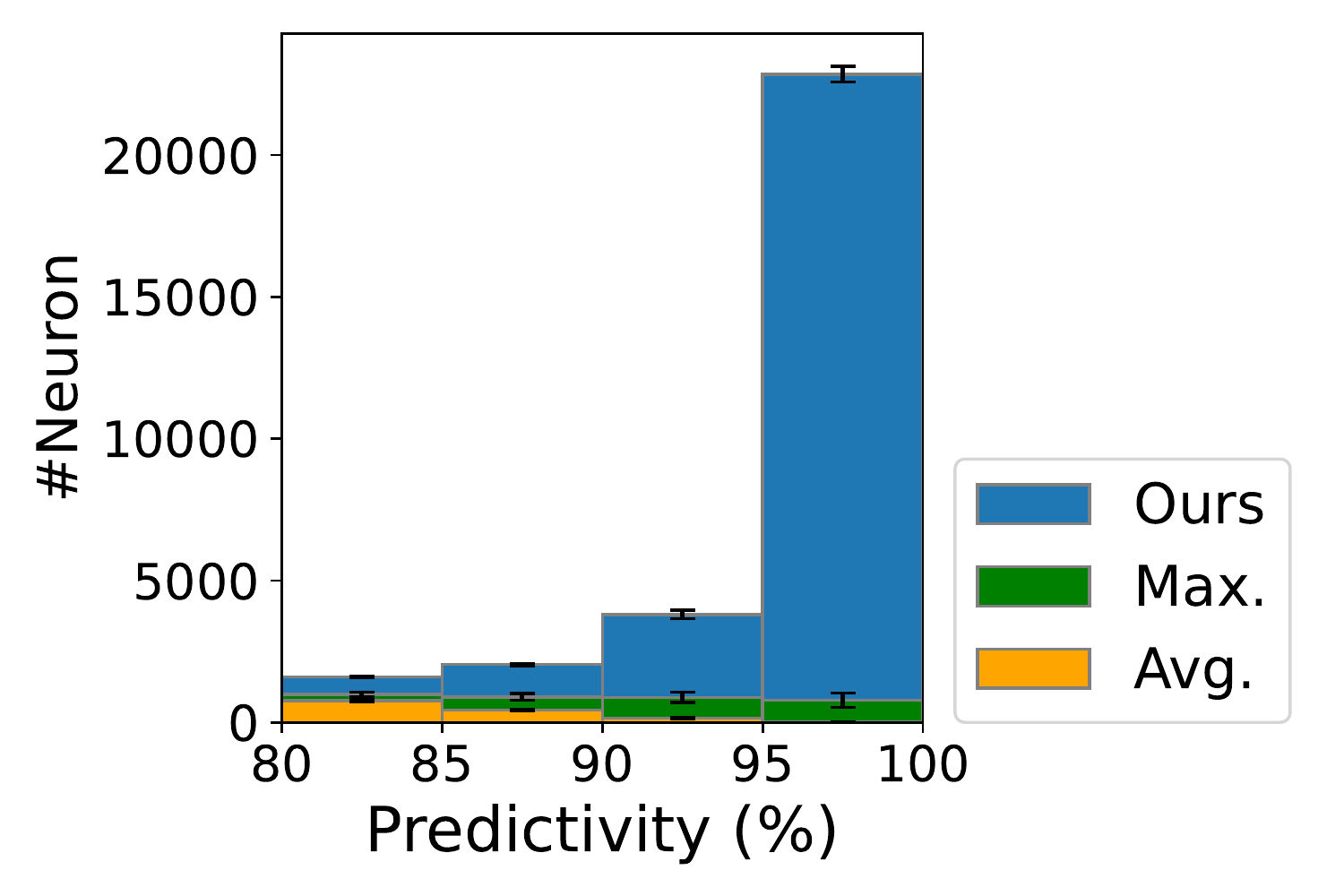}
    }
    \subfigure[\texttt{AG News}]{
	    \includegraphics[width=0.48\textwidth]{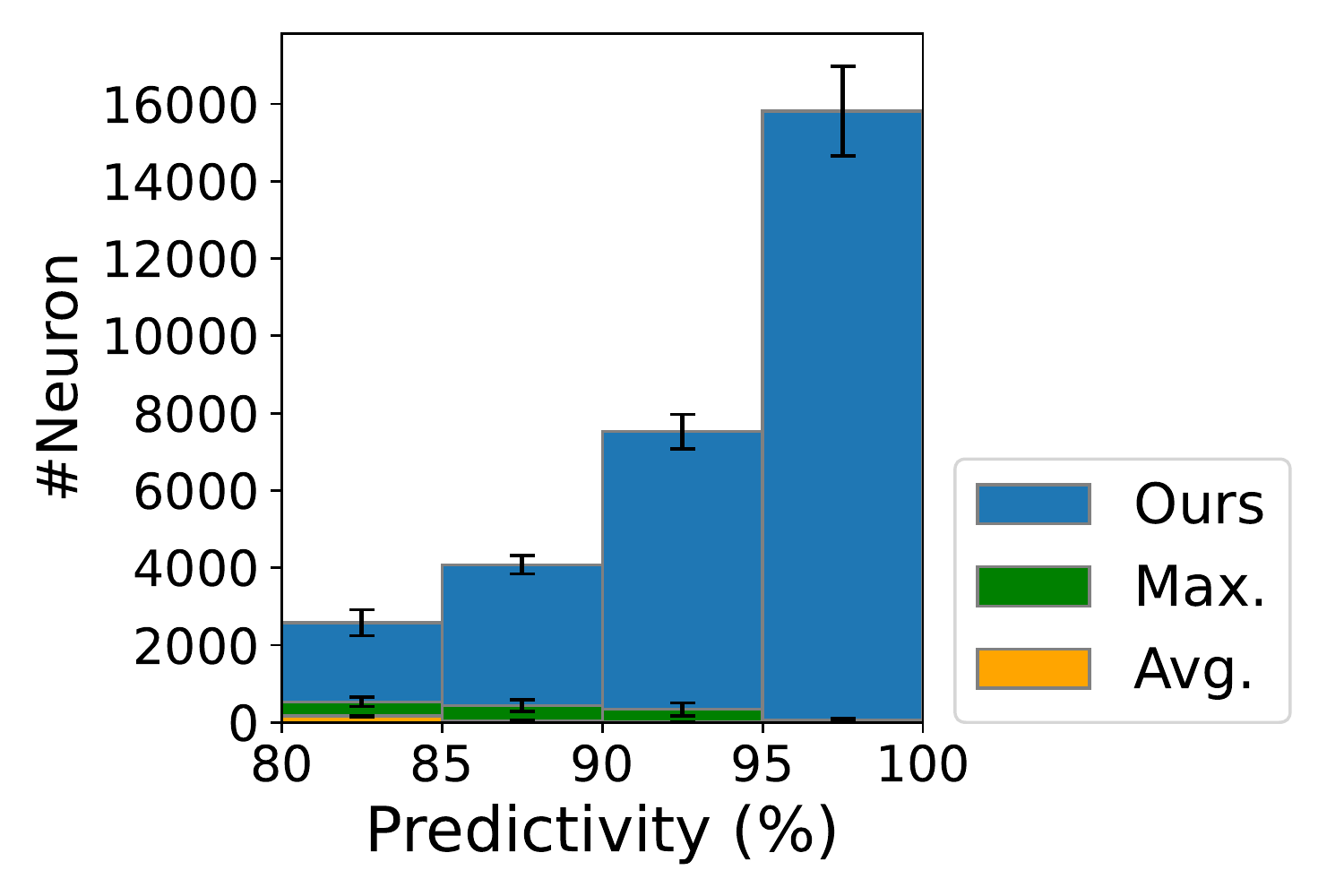}
    }
    
    \caption{Histograms of predictivity for various tasks on neurons within \RBT. Error bars indicate $\pm 1$ s.e.m. over $5$ random trials.}
    \label{fig:app_acc_dist}
\end{figure*}

We report the predictivity distribution for IMDB in \cref{sec:stable} and show the distributions for the other $4$ binary classification tasks in \cref{fig:app_acc_dist}. We can see our method can stably find many highly-predictive skill neurons for all the tasks. For the multi-class classification tasks, since the predictivities are for decomposed subtasks, we cannot draw distributions for the original tasks and do not include them in the results here. 

\begin{figure}[!t]
\small
\includegraphics[width=0.48\textwidth]{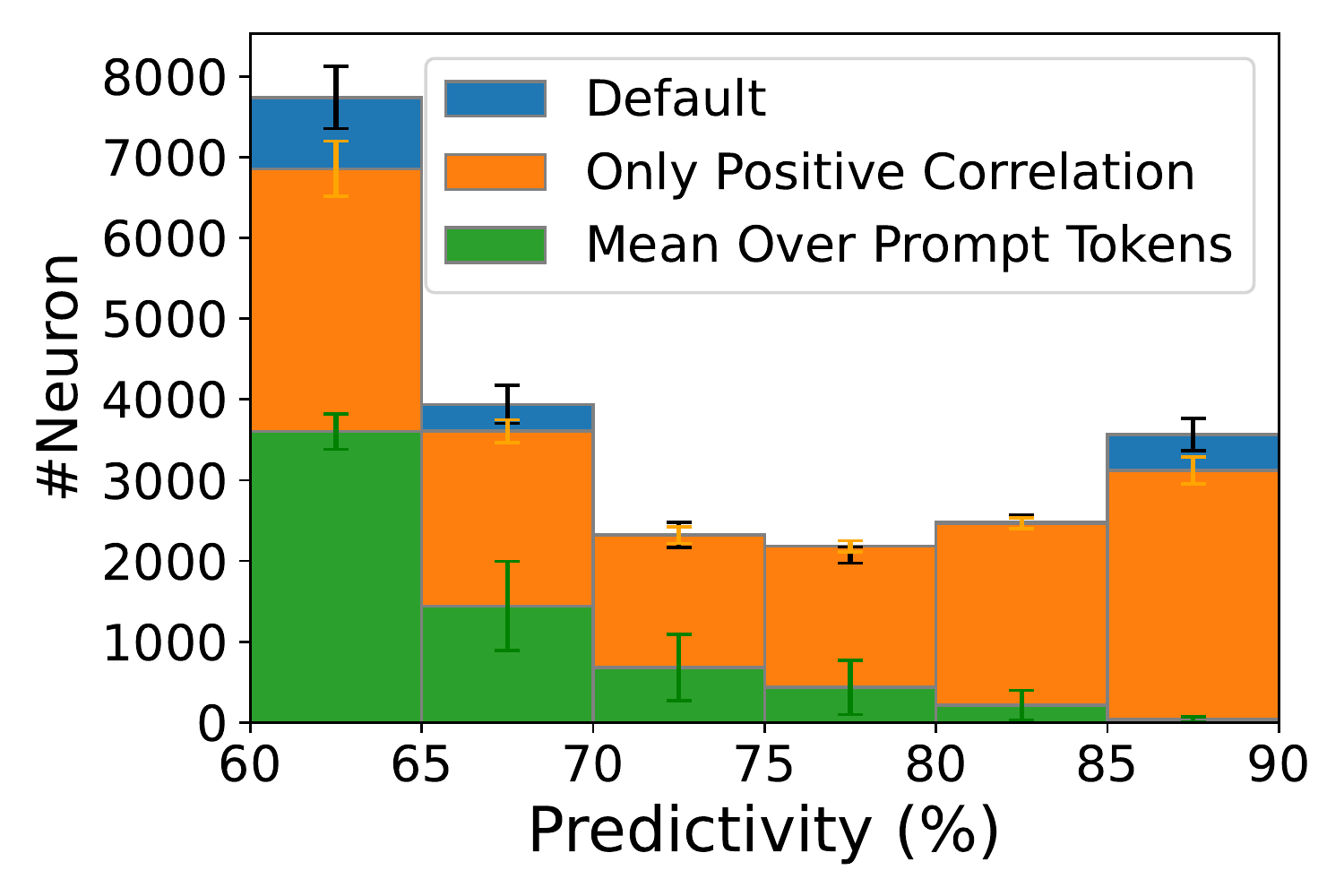}
    \caption{Histogram of neuron's predictivity in different definitions for \texttt{SST-2}. Error bars indicate $\pm1$ s.e.m. over $5$ random trials.}
    \label{fig:design_choice}
\end{figure}

\section{More Neuron Perturbation Results}
\label{app:perturbation}
Here we demonstrate more neuron perturbation experimental results.
\subsection{Performance Dropping Trends for Prompt Tuning}
\label{app:mask_pt}
\begin{figure*}[!t]
\subcapraggedrighttrue
\subcaphangtrue
    \centering
    
    \subfigure[On \texttt{SST-2}]{
	    \includegraphics[width=0.48\textwidth]{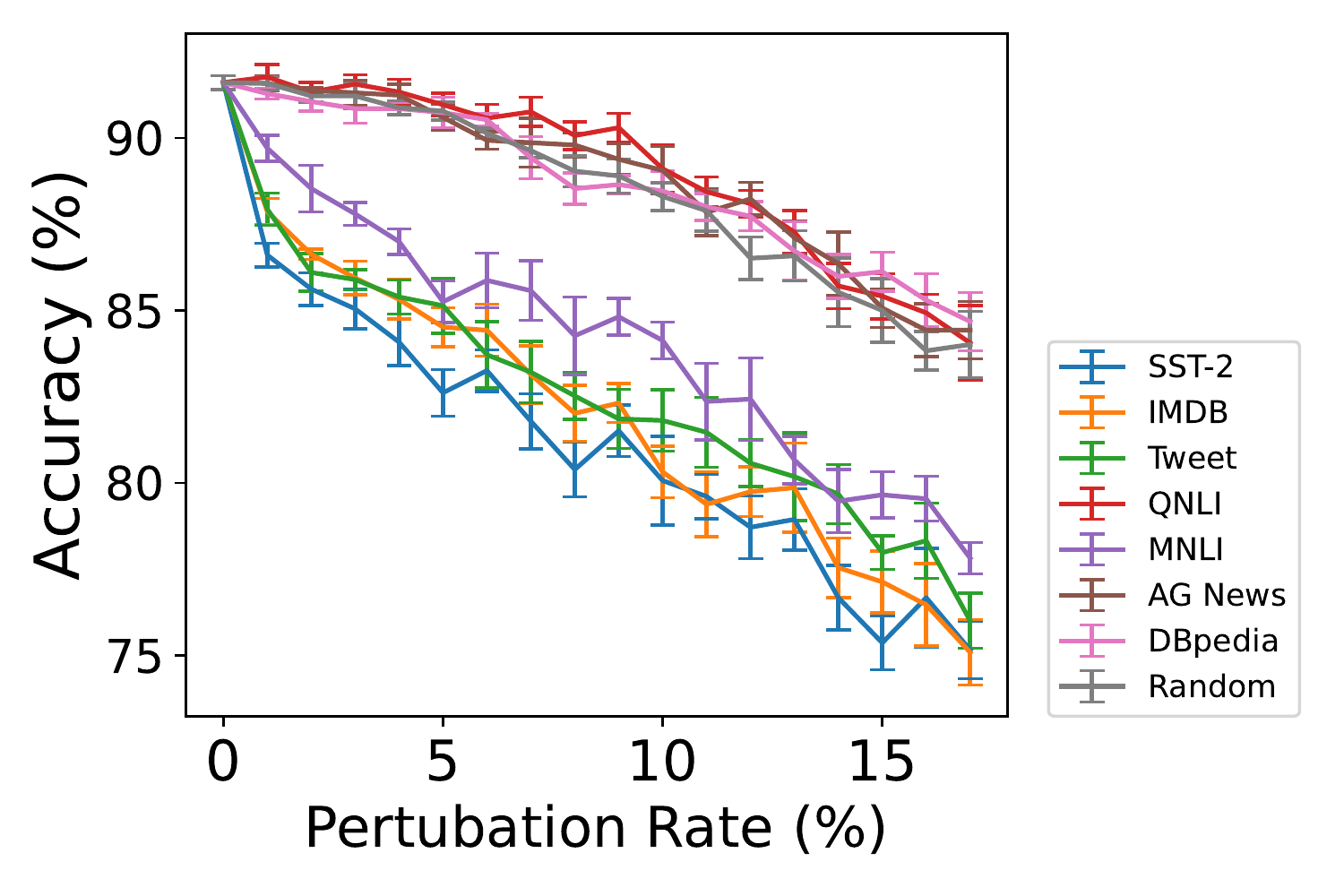}
    }
    \subfigure[On \texttt{IMDB}]{
	    \includegraphics[width=0.48\textwidth]{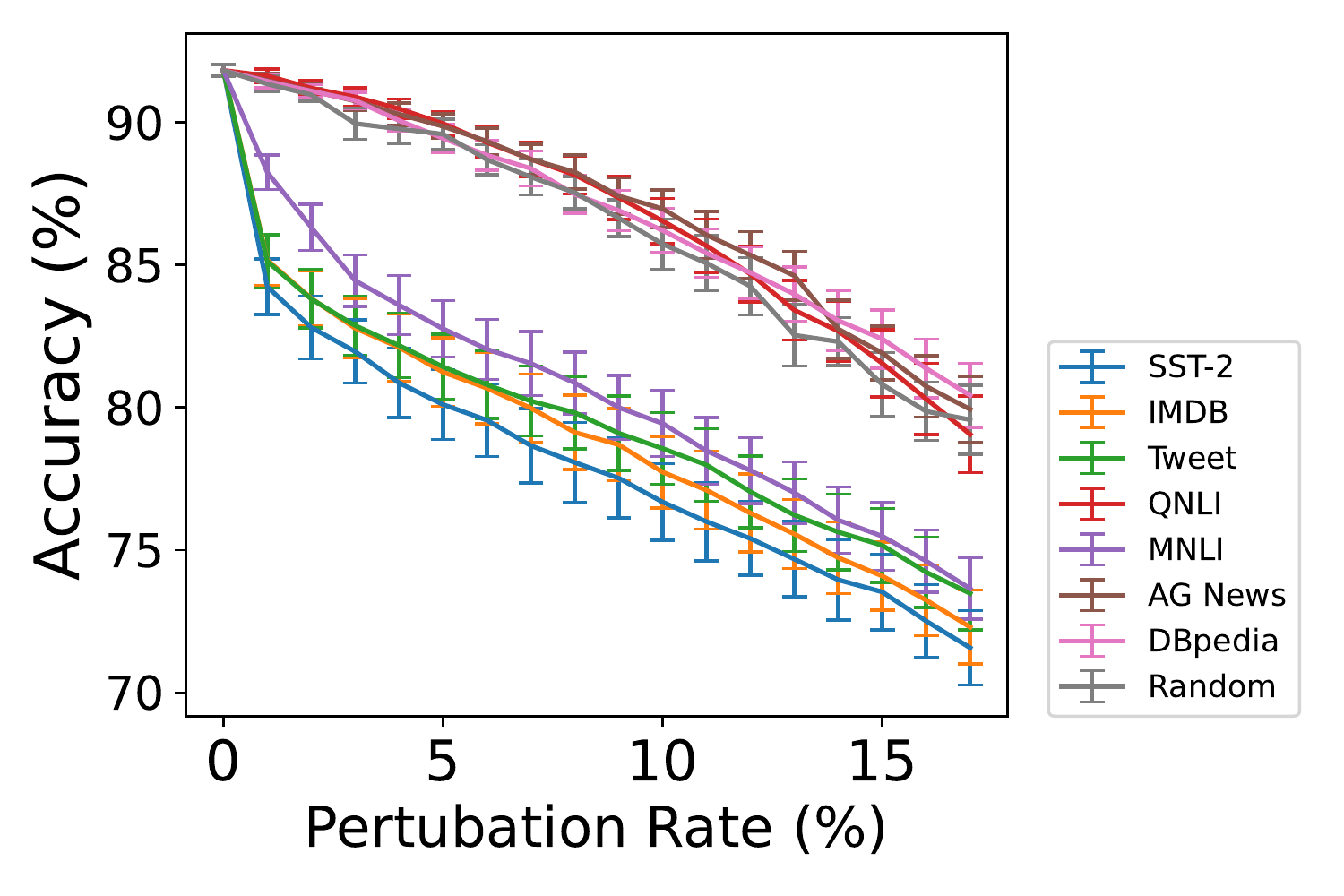}
    }
    \subfigure[On \texttt{MNLI}]{
	    \includegraphics[width=0.48\textwidth]{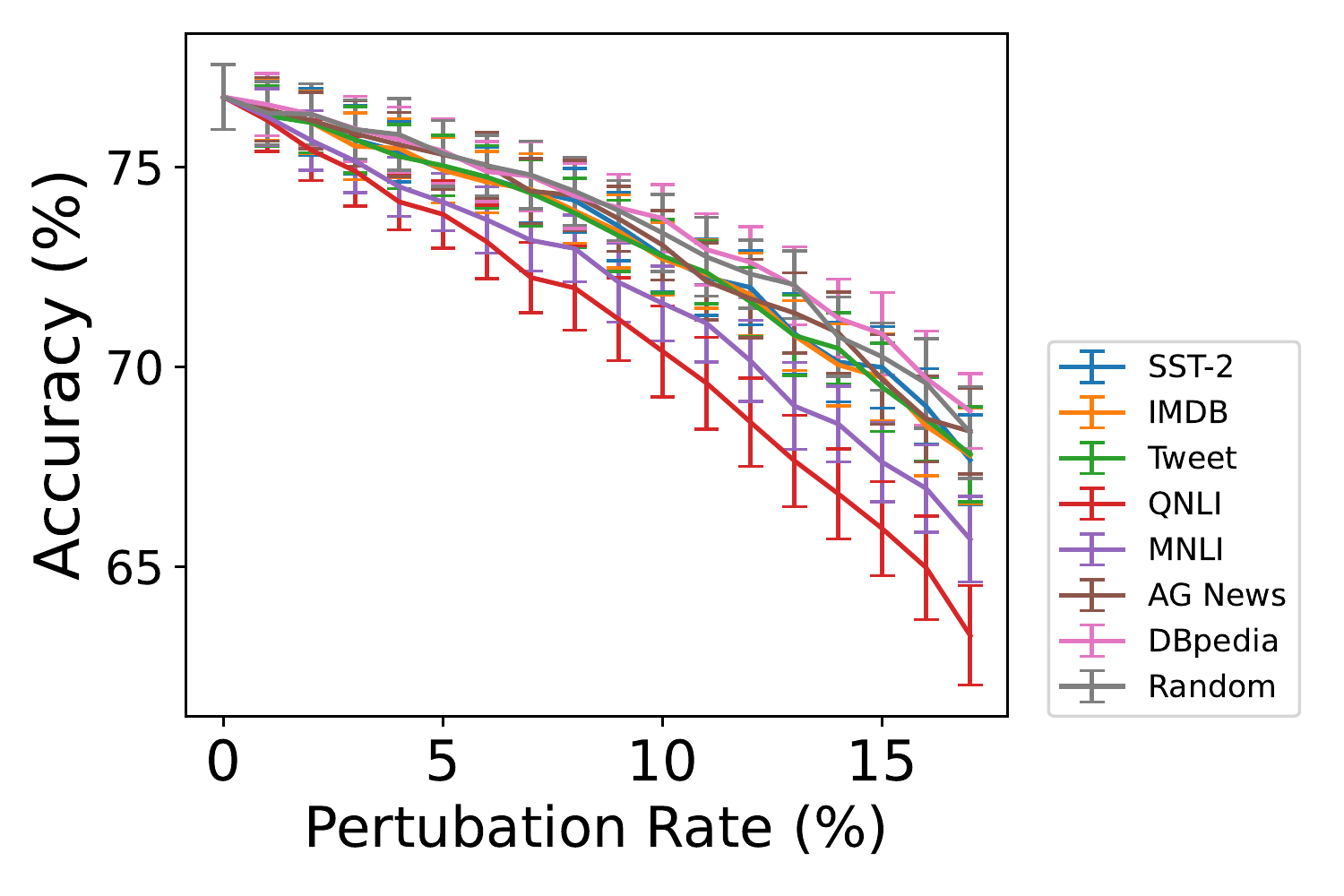}
    }
    \subfigure[On \texttt{QNLI}]{
	    \includegraphics[width=0.48\textwidth]{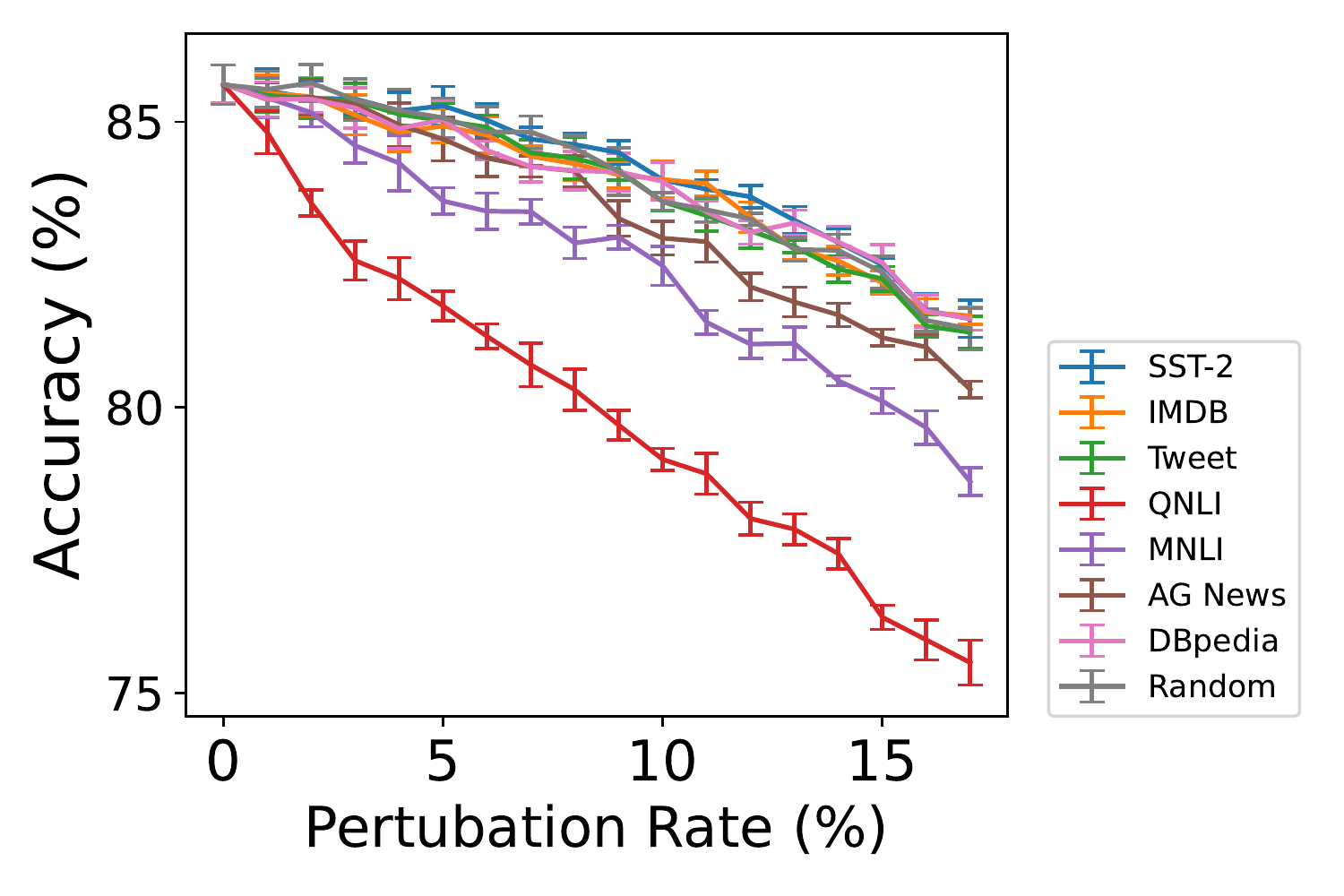}
    }
    \subfigure[On \texttt{AG News}]{
	    \includegraphics[width=0.48\textwidth]{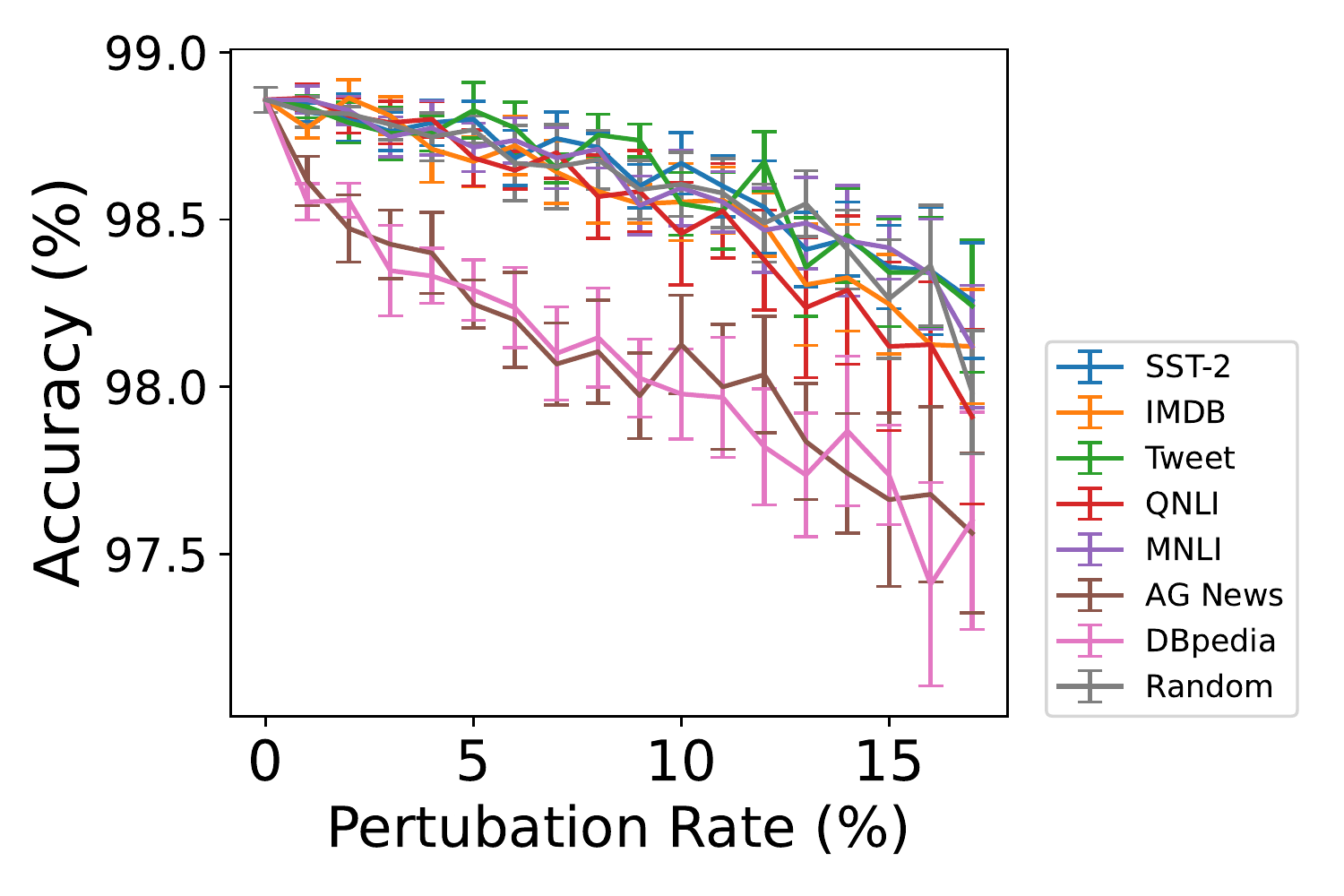}
    }
    \subfigure[On \texttt{DBpedia}]{
	    \includegraphics[width=0.48\textwidth]{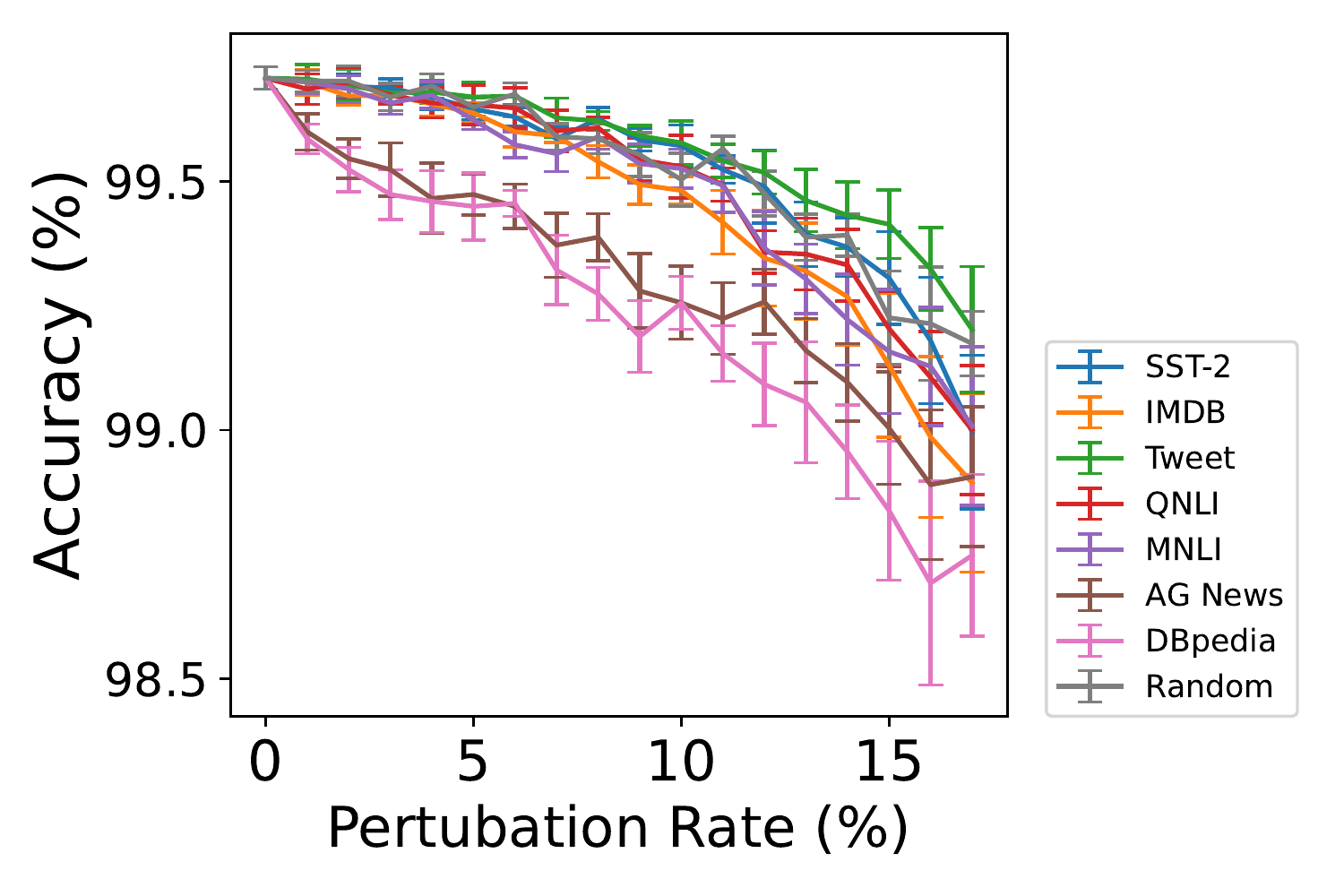}
    }
    \caption{Accuracies on various tasks drop along with the neuron perturbation rates. Error bars indicate $\pm 1$ s.e.m. over $5$ random trials. The perturbations are conducted in descending orders of neurons' predictivities for different tasks or in random order (the ``Random'' curve). }
    \label{fig:app_mask_trend}
\end{figure*}

In \cref{fig:mask_trend}, we show the performance dropping trend on \texttt{Tweet} task. The results on the other tasks are shown in \cref{fig:app_mask_trend}.

\subsection{Performance Dropping Trends for Adapter-based Tuning}
\label{app:mask_adapter}
\begin{figure*}[!t]
\subcapraggedrighttrue
\subcaphangtrue
    \centering
    
    \subfigure[On \texttt{SST-2}]{
	    \includegraphics[width=0.48\textwidth]{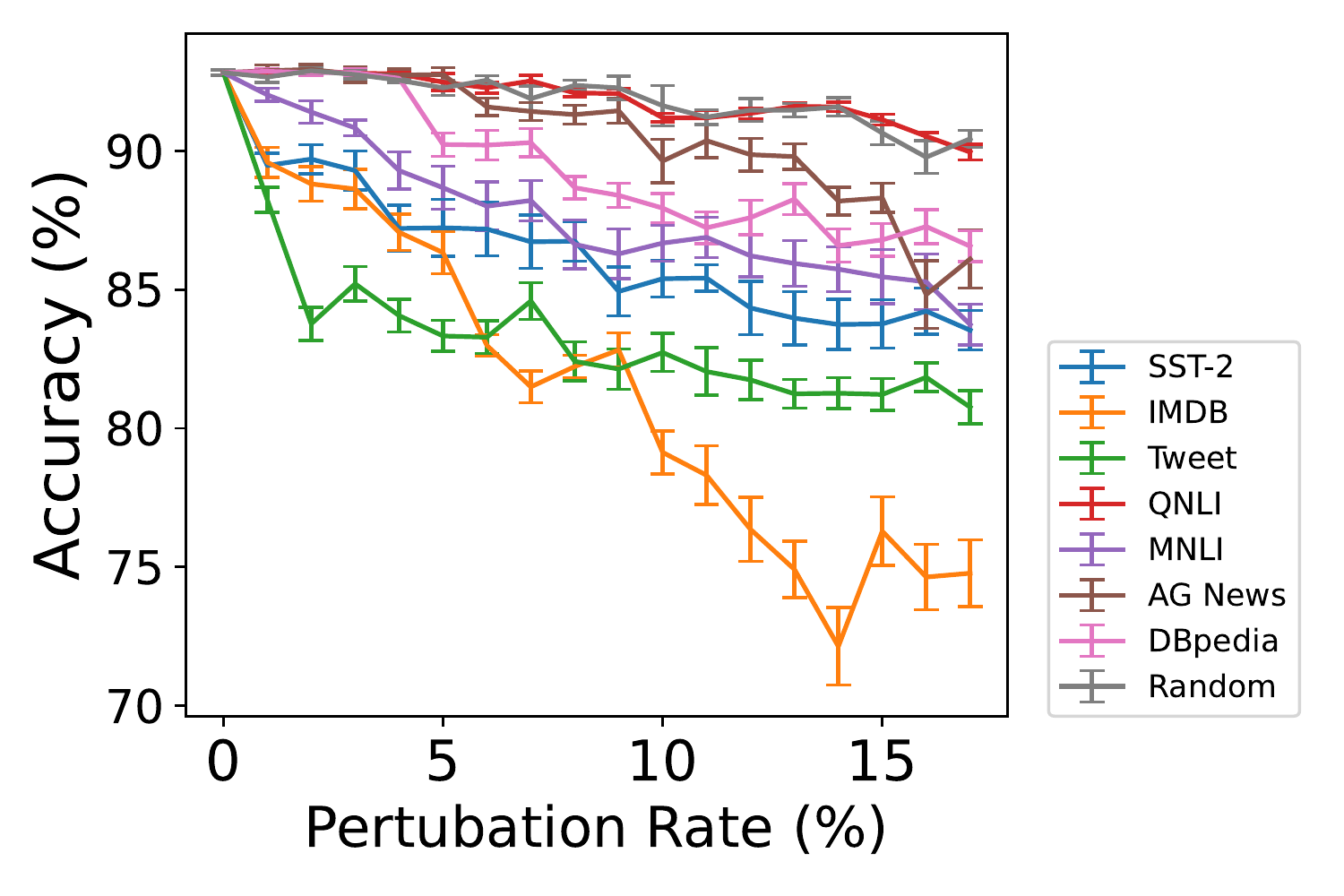}
    }
    \subfigure[On \texttt{IMDB}]{
	    \includegraphics[width=0.48\textwidth]{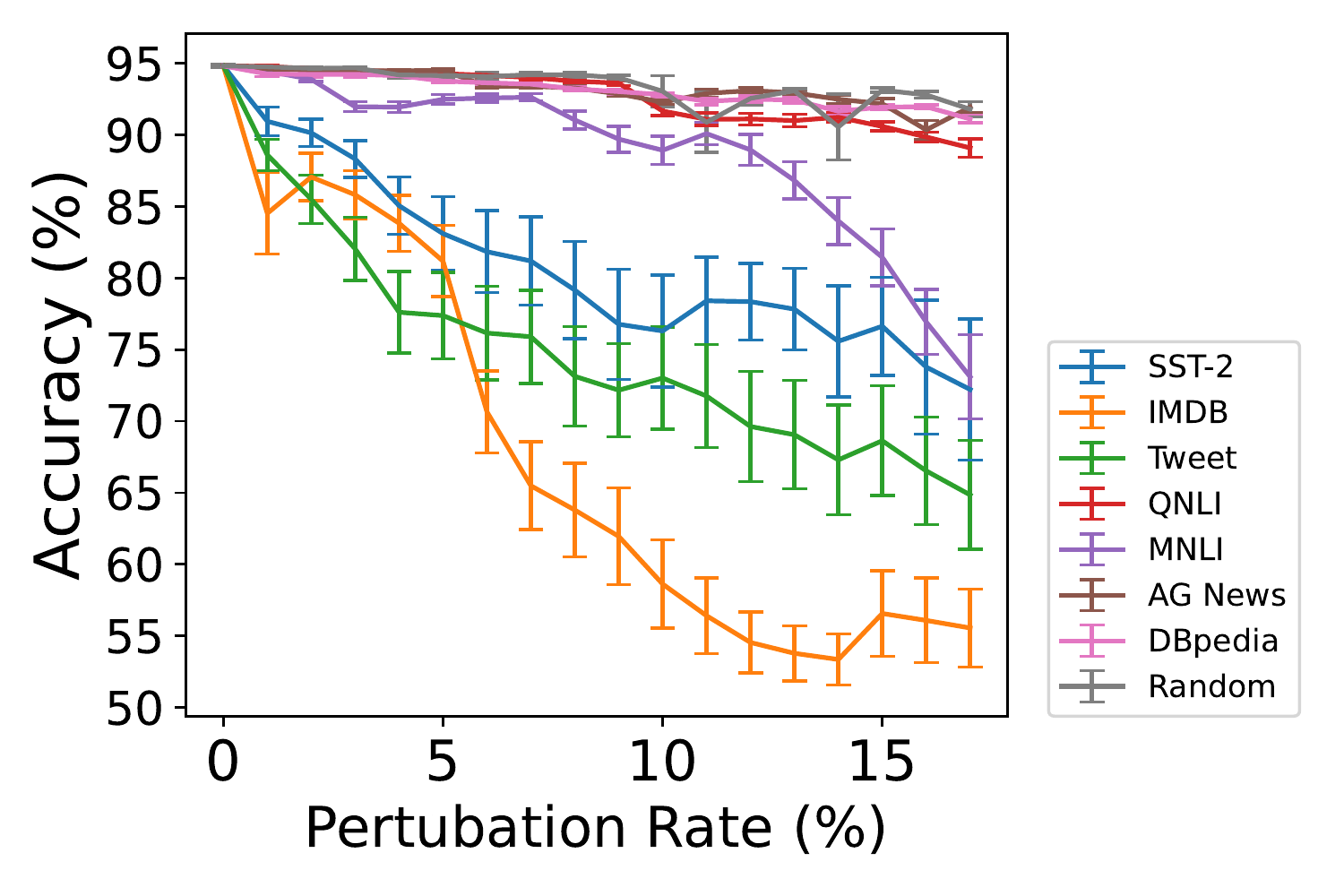}
    }
    \subfigure[On \texttt{Tweet}]{
	    \includegraphics[width=0.48\textwidth]{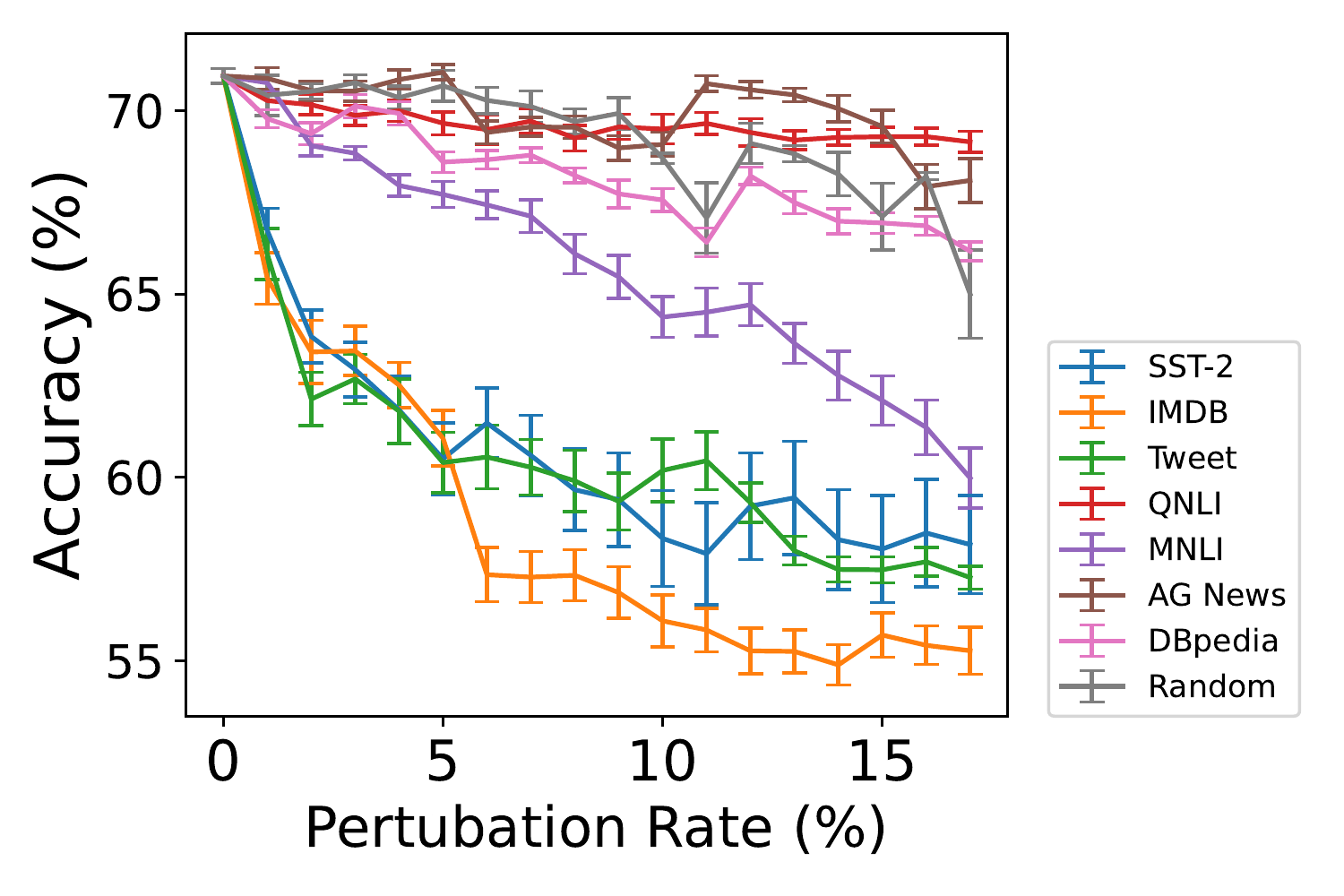}
    }
    \subfigure[On \texttt{MNLI}]{
	    \includegraphics[width=0.48\textwidth]{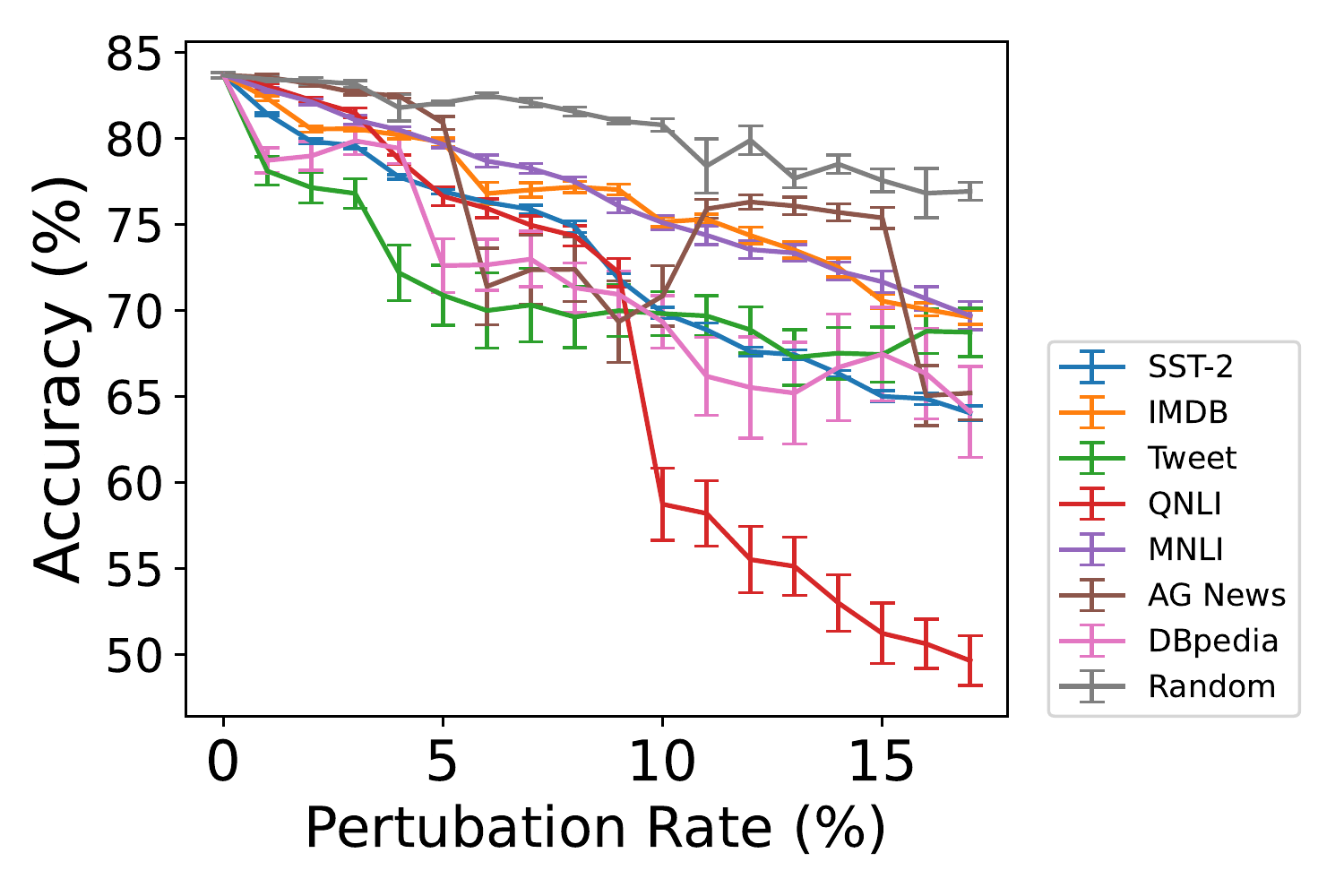}
    }
    \subfigure[On \texttt{QNLI}]{
	    \includegraphics[width=0.48\textwidth]{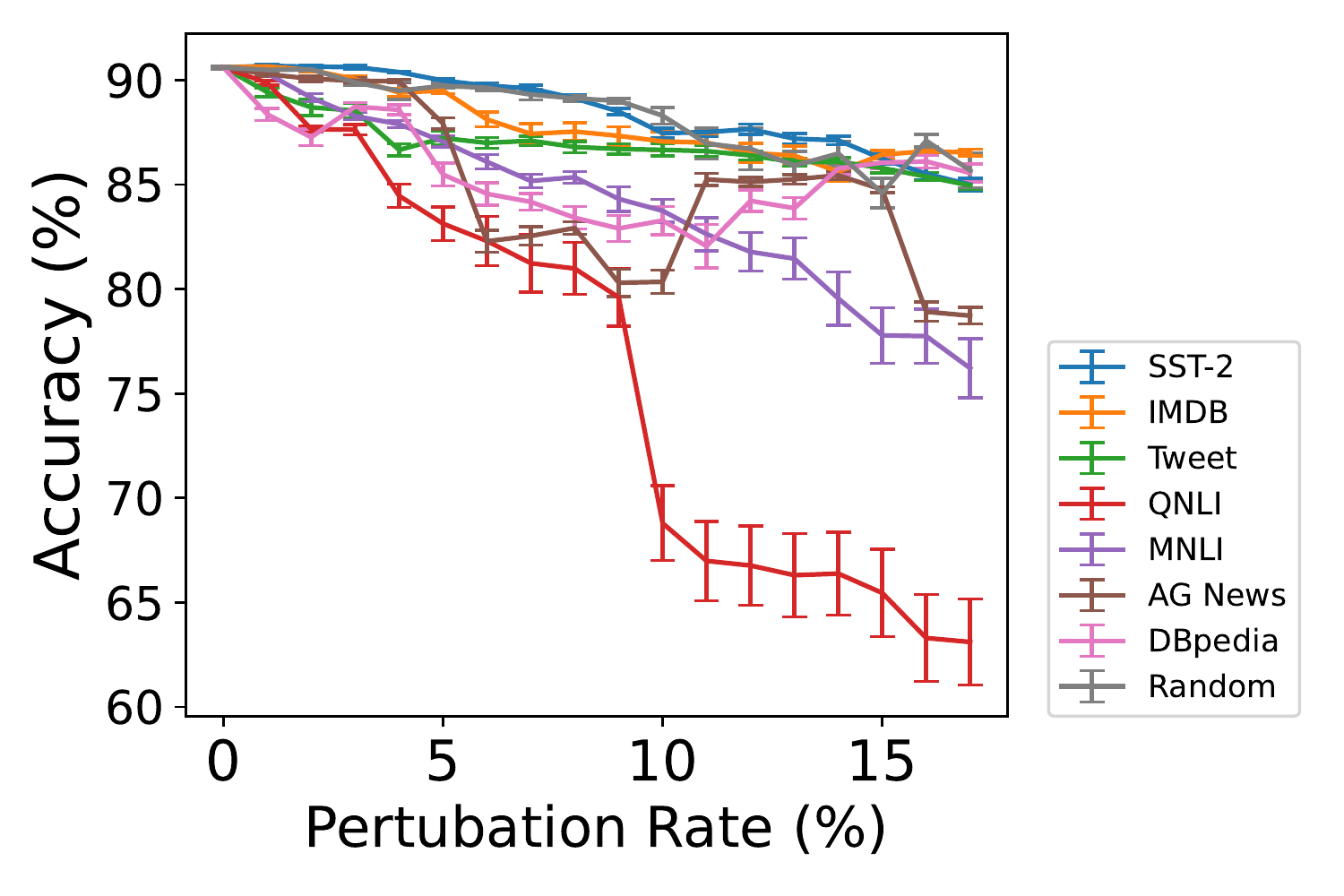}
    }
    \subfigure[On \texttt{AG News}]{
	    \includegraphics[width=0.48\textwidth]{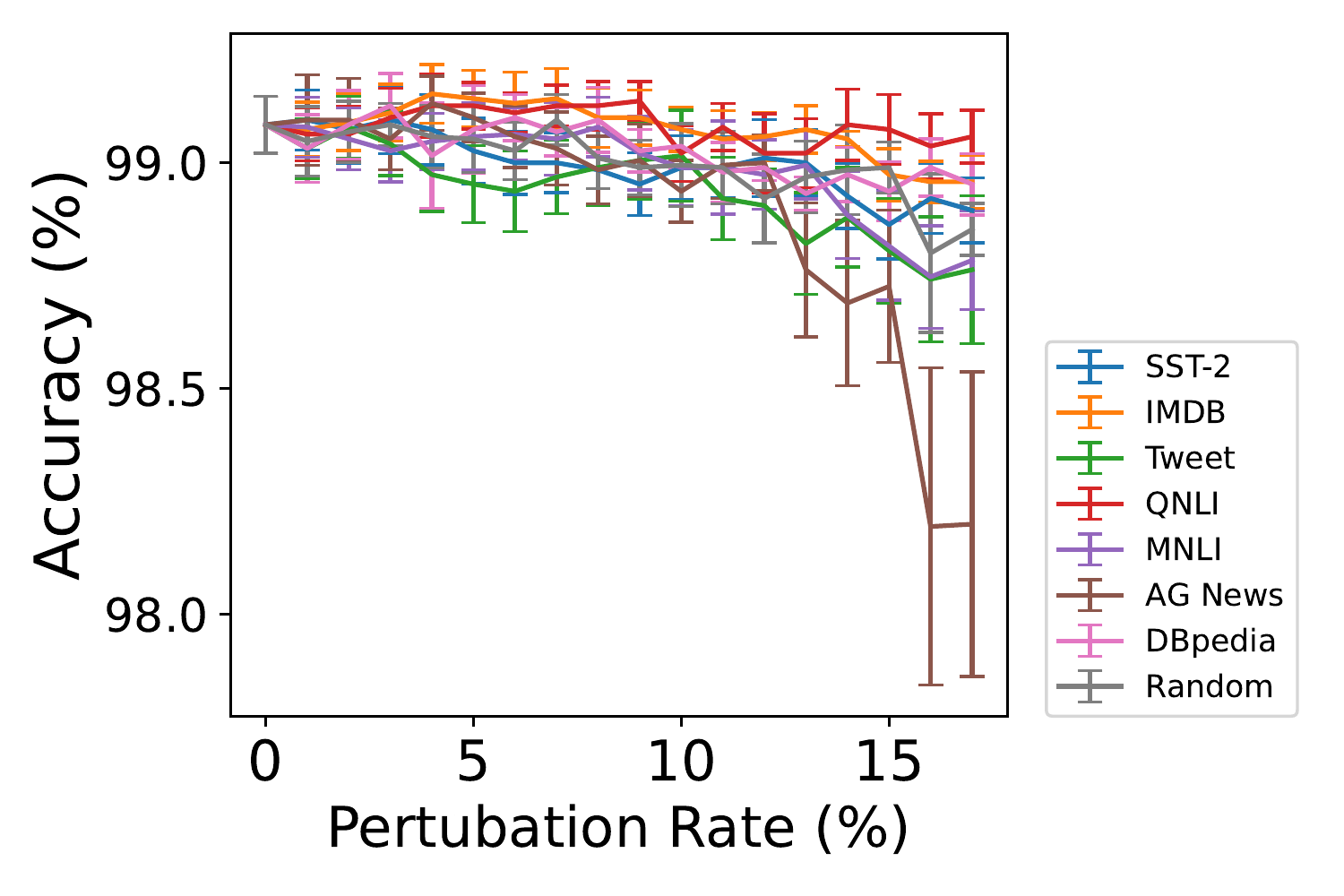}
    }
    \subfigure[On \texttt{DBpedia}]{
	    \includegraphics[width=0.48\textwidth]{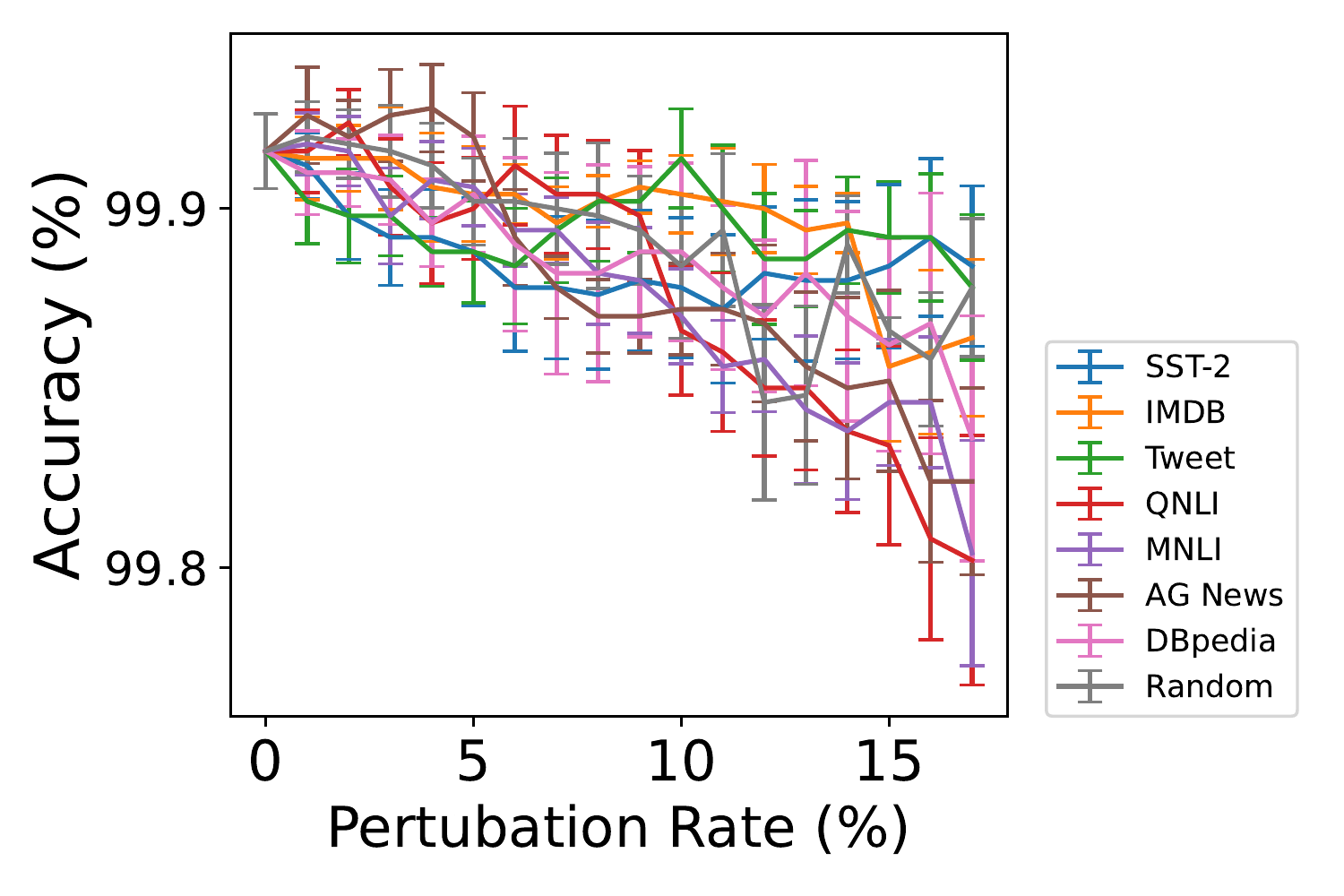}
    }
    \caption{Adapter-based tuning accuracies on various tasks drop along with the neuron perturbation rates. Error bars indicate $\pm 1$ s.e.m. over $5$ random trials. The perturbations are conducted in predictivity orders obtained with prompt tuning.}
    \label{fig:app_mask_trend_adapter}
\end{figure*}

The performance dropping trends of adapter-based tuning models on various tasks are shown in \cref{fig:app_mask_trend_adapter}.

\subsection{Performance Dropping Trends for BitFit}
\label{app:mask_bias}
\begin{figure*}[!t]
\subcapraggedrighttrue
\subcaphangtrue
    \centering
    
    \subfigure[On \texttt{SST-2}]{
	    \includegraphics[width=0.48\textwidth]{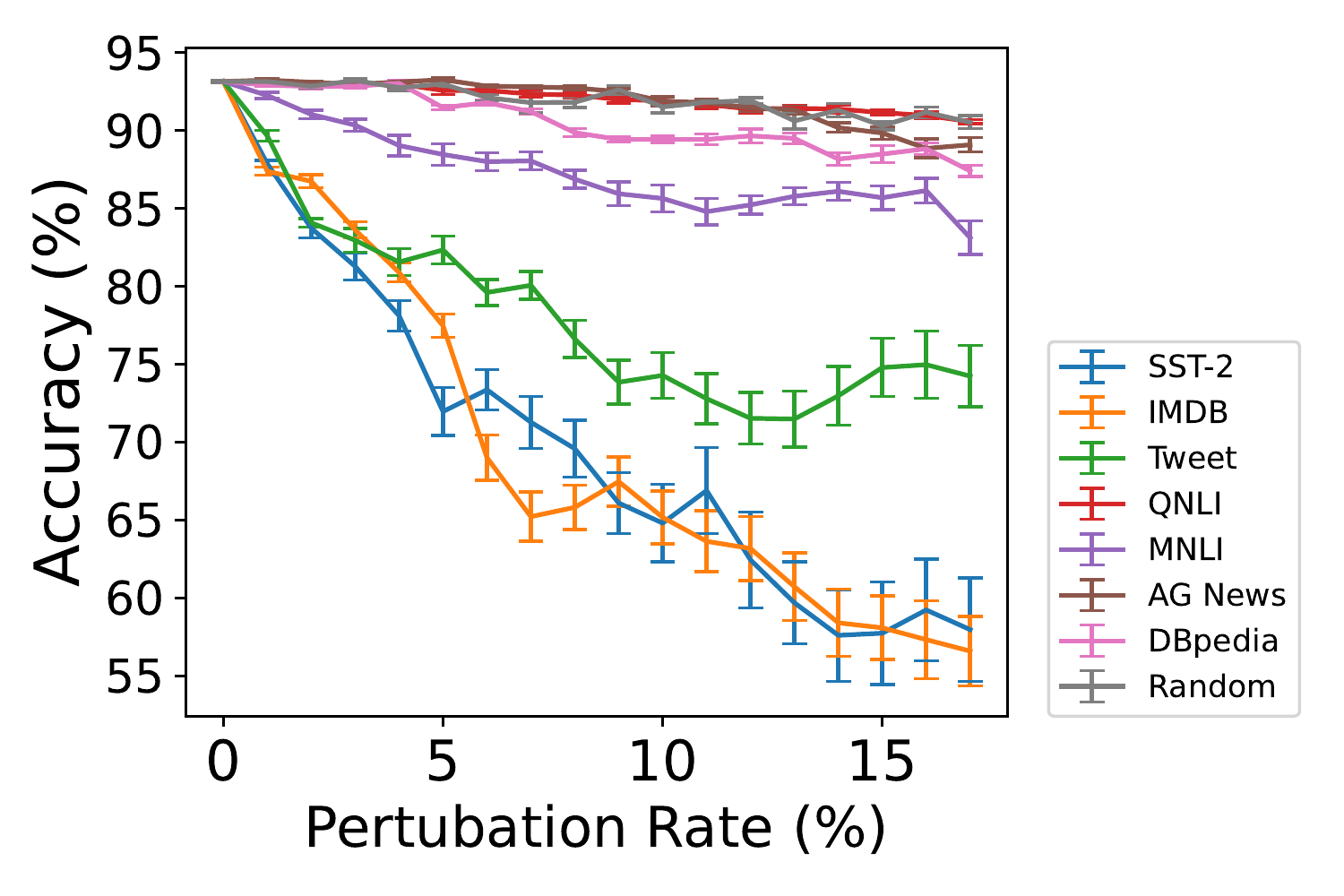}
    }
    \subfigure[On \texttt{Tweet}]{
	    \includegraphics[width=0.48\textwidth]{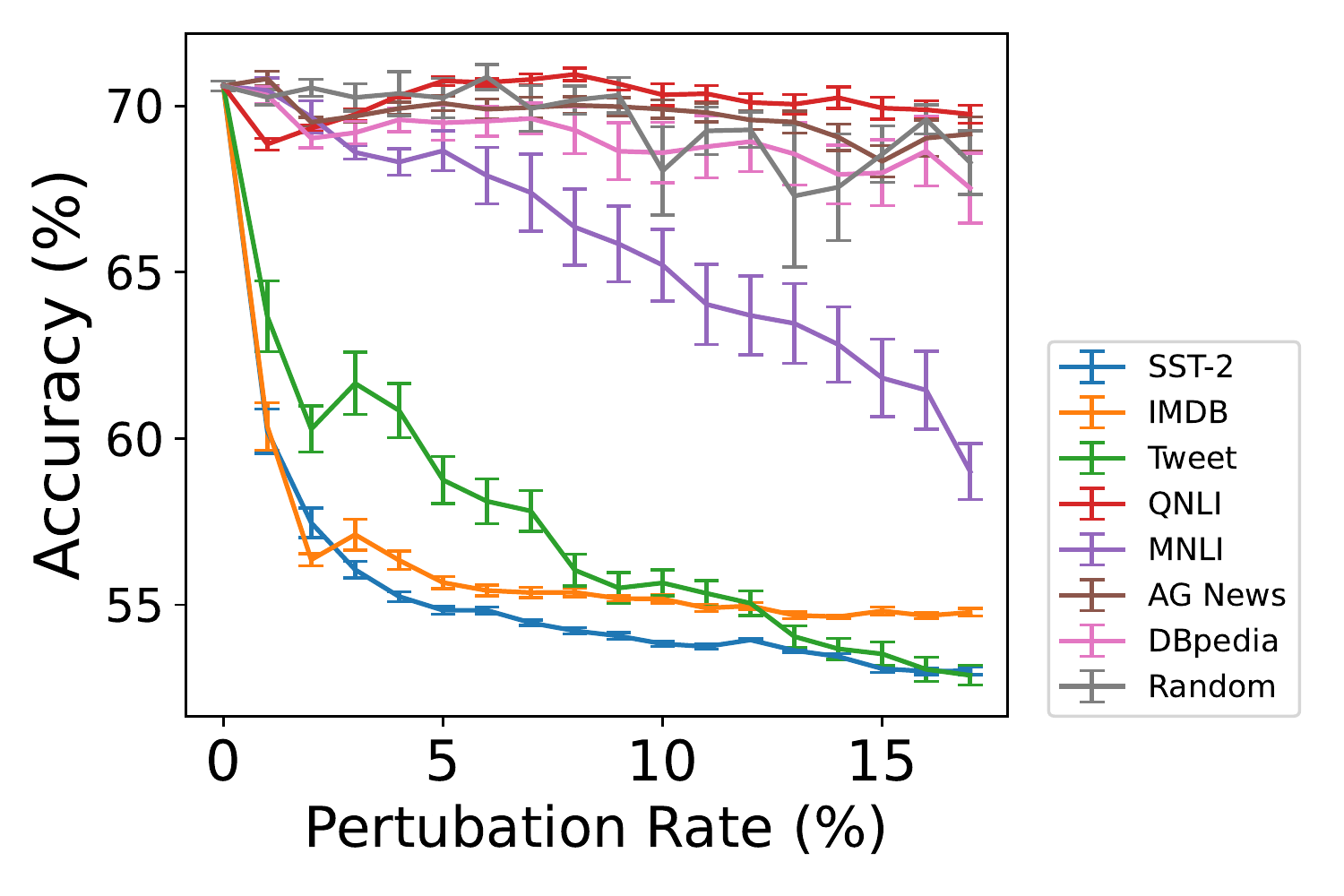}
    }
    \subfigure[On \texttt{MNLI}]{
	    \includegraphics[width=0.48\textwidth]{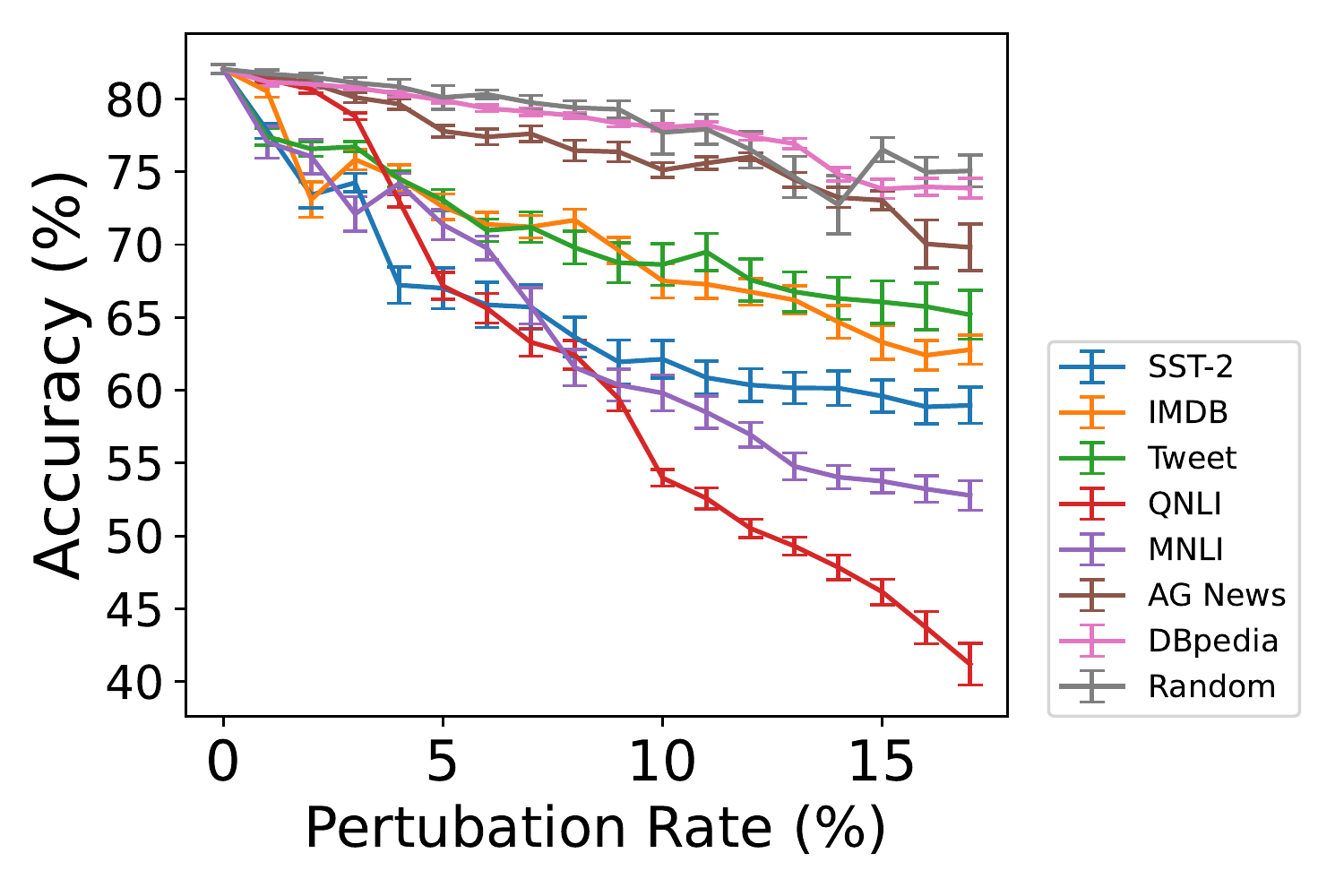}
    }
    \subfigure[On \texttt{QNLI}]{
	    \includegraphics[width=0.48\textwidth]{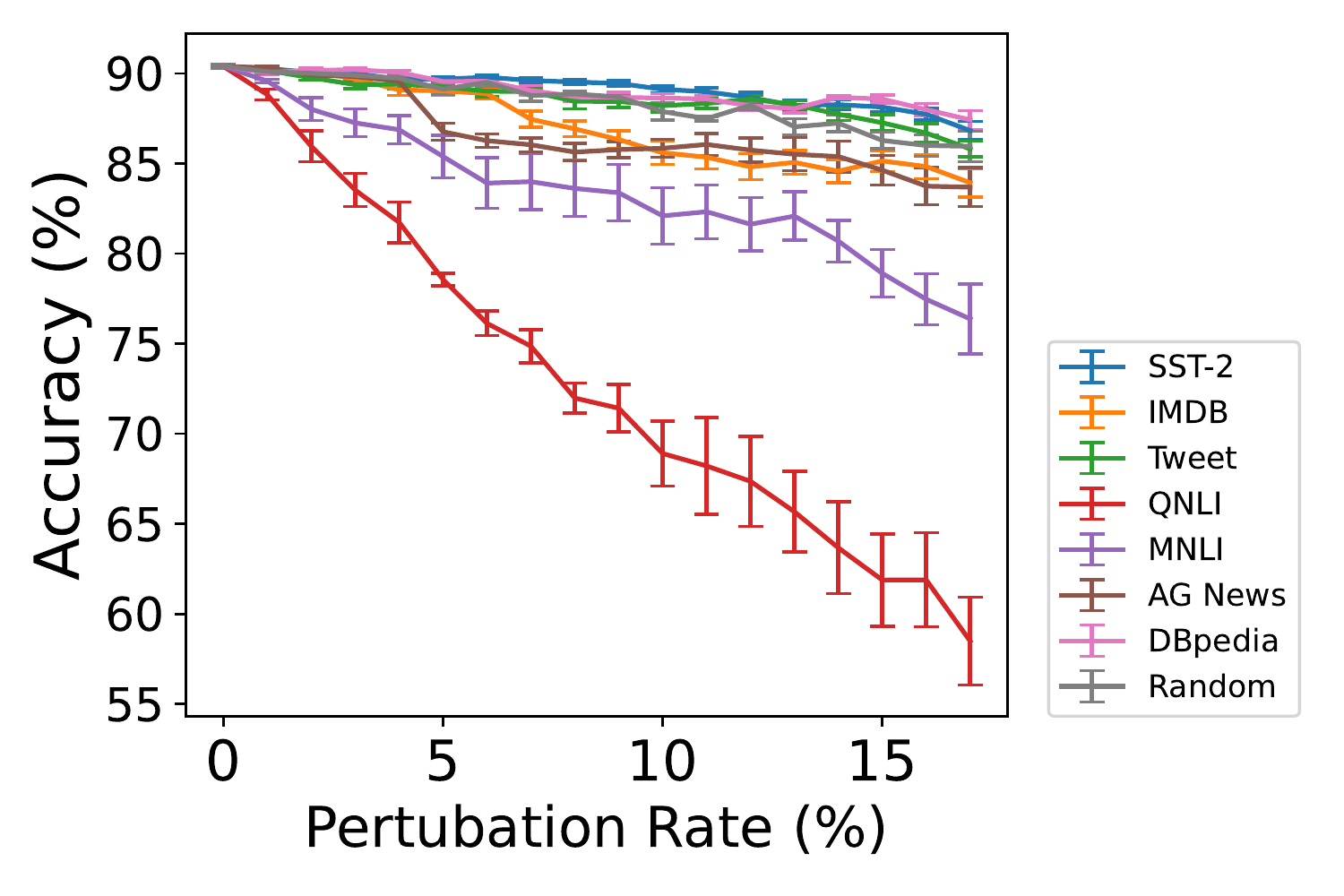}
    }
    \subfigure[On \texttt{AG News}]{
	    \includegraphics[width=0.48\textwidth]{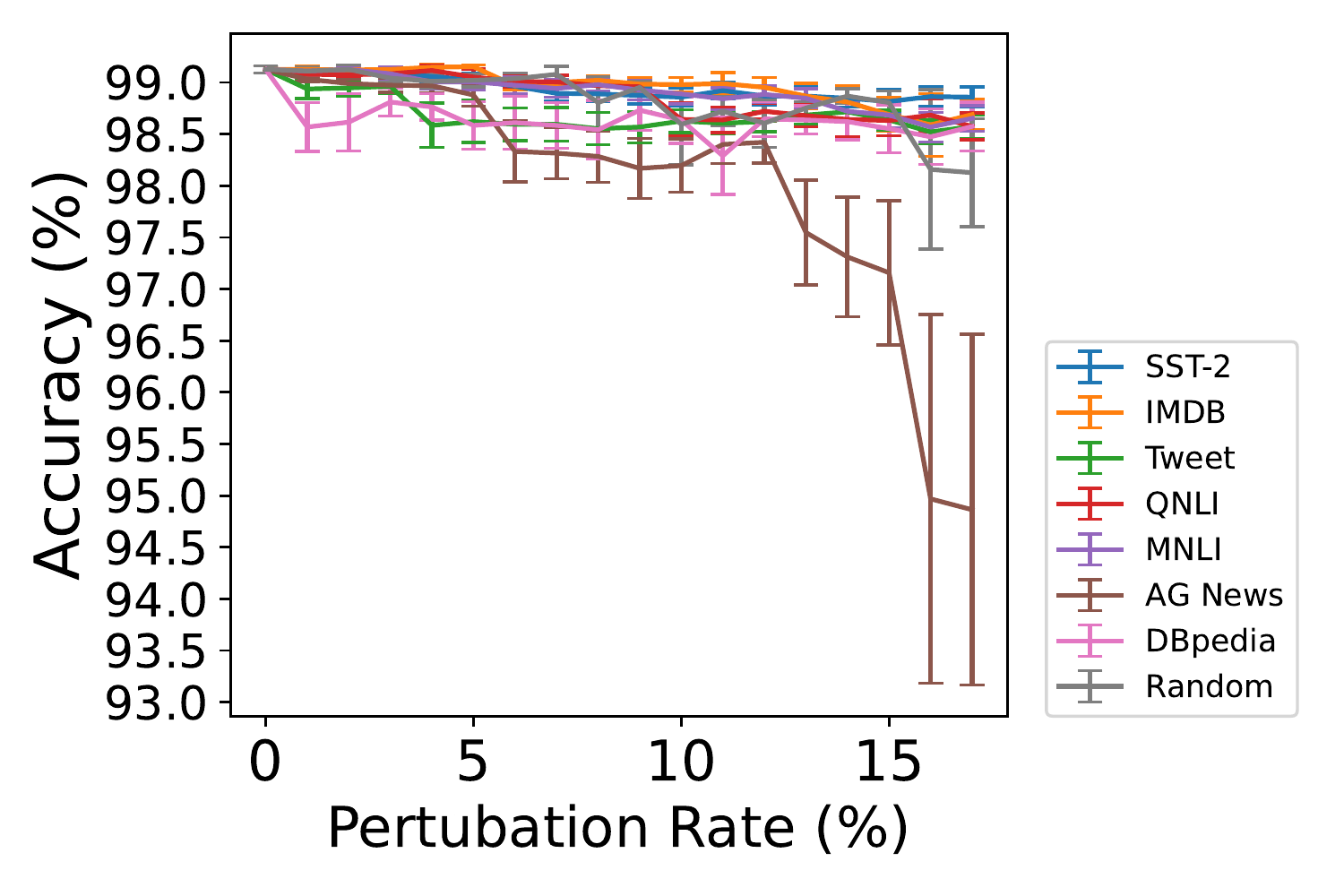}
    }
    \subfigure[On \texttt{DBpedia}]{
	    \includegraphics[width=0.48\textwidth]{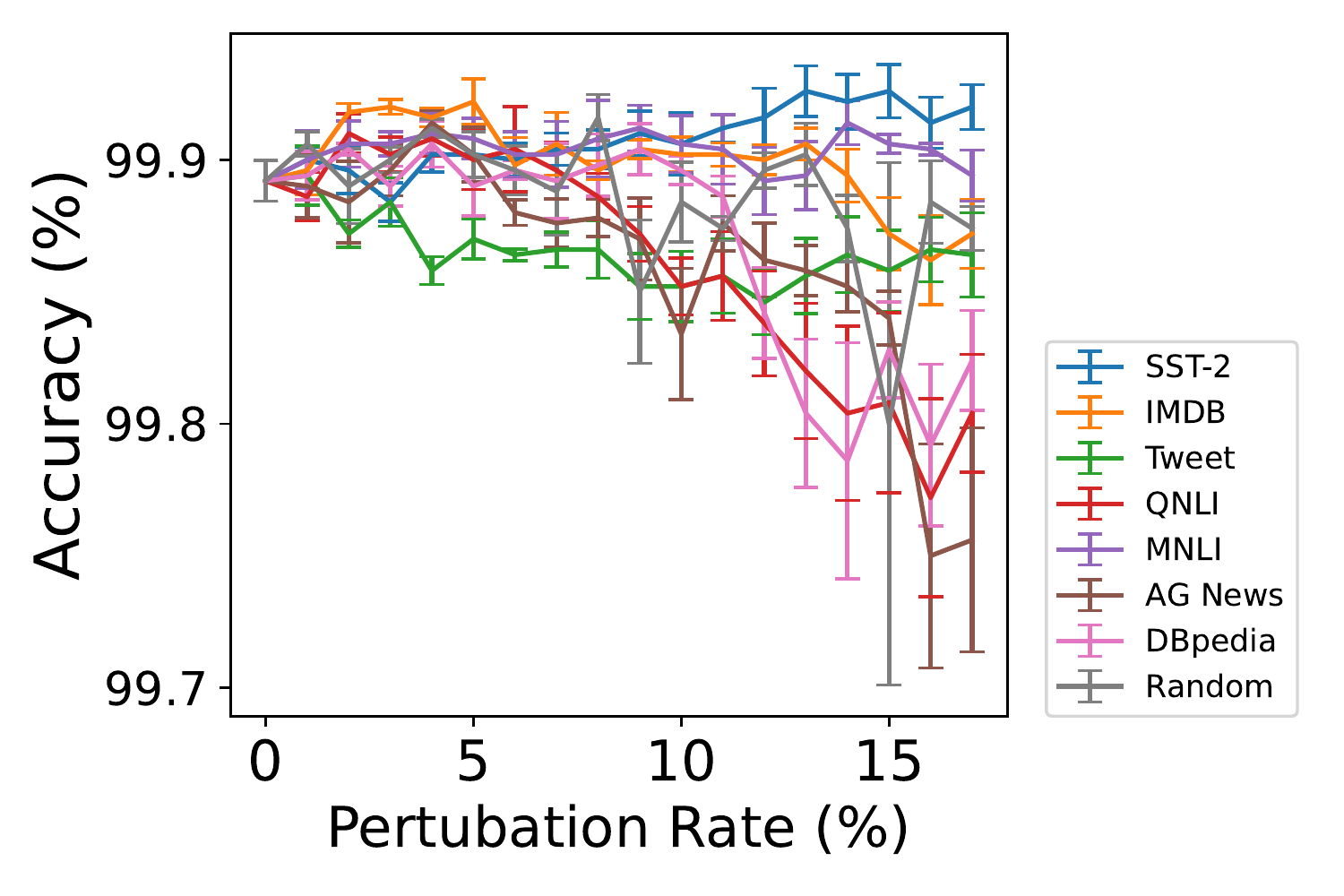}
    }
    \caption{BitFit accuracies on various tasks drop along with the neuron perturbation rates. Error bars indicate $\pm 1$ s.e.m. over $5$ random trials. The perturbations are conducted in predictivity orders obtained with prompt tuning.}
    \label{fig:app_mask_trend_bias}
\end{figure*}

The performance dropping trends of BitFit models on various tasks are shown in \cref{fig:app_mask_trend_bias}.

\section{Layer-wise Correlations between Neuron Predictivity Orders of Different Tasks}

\begin{figure*}[!t]
\subcapraggedrighttrue
\subcaphangtrue
    \centering
    
    \subfigure[Layer $1$]{
	    \includegraphics[width=0.30\textwidth]{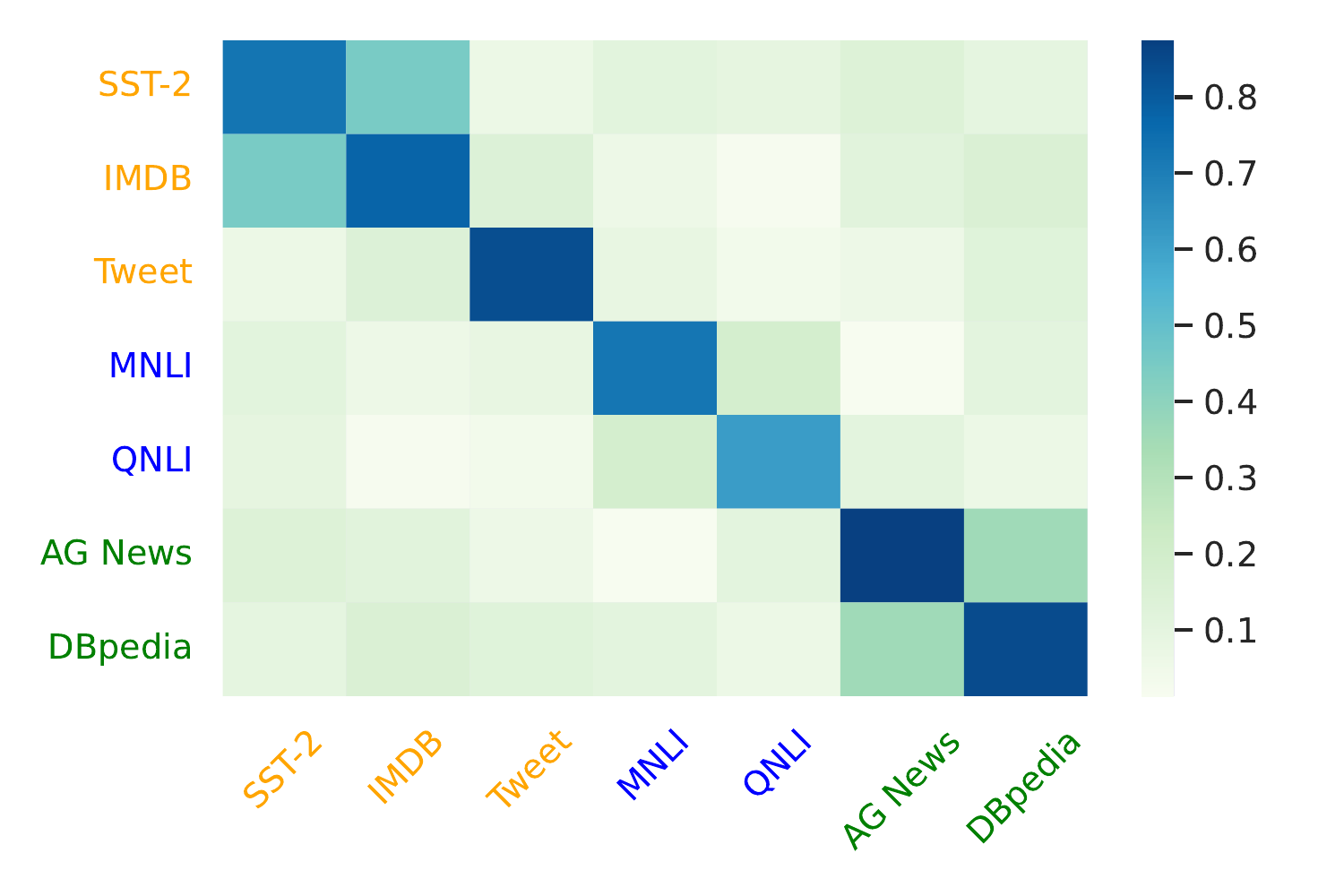}
    }
    \subfigure[Layer $2$]{
	    \includegraphics[width=0.30\textwidth]{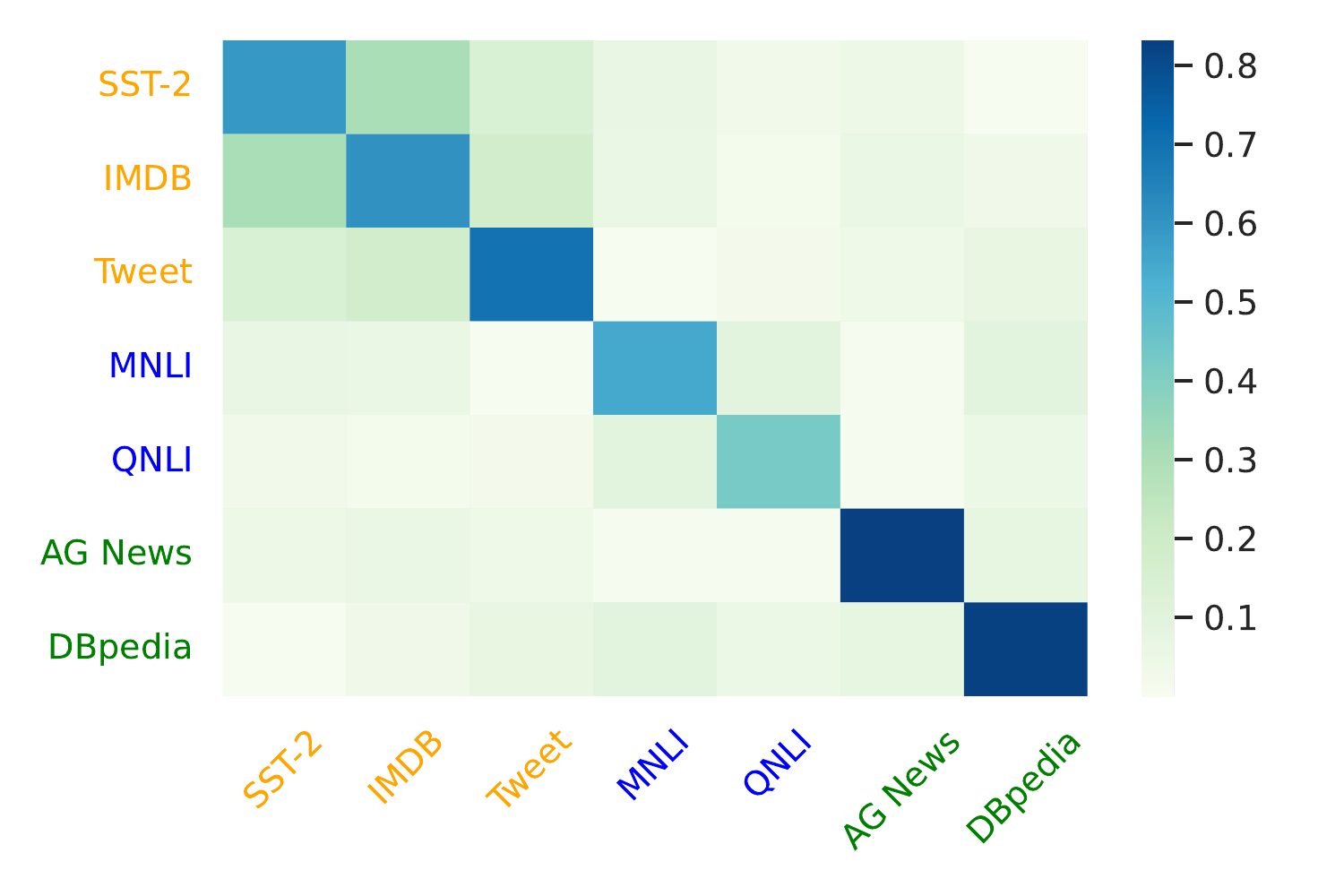}
    }
    \subfigure[Layer $3$]{
	    \includegraphics[width=0.30\textwidth]{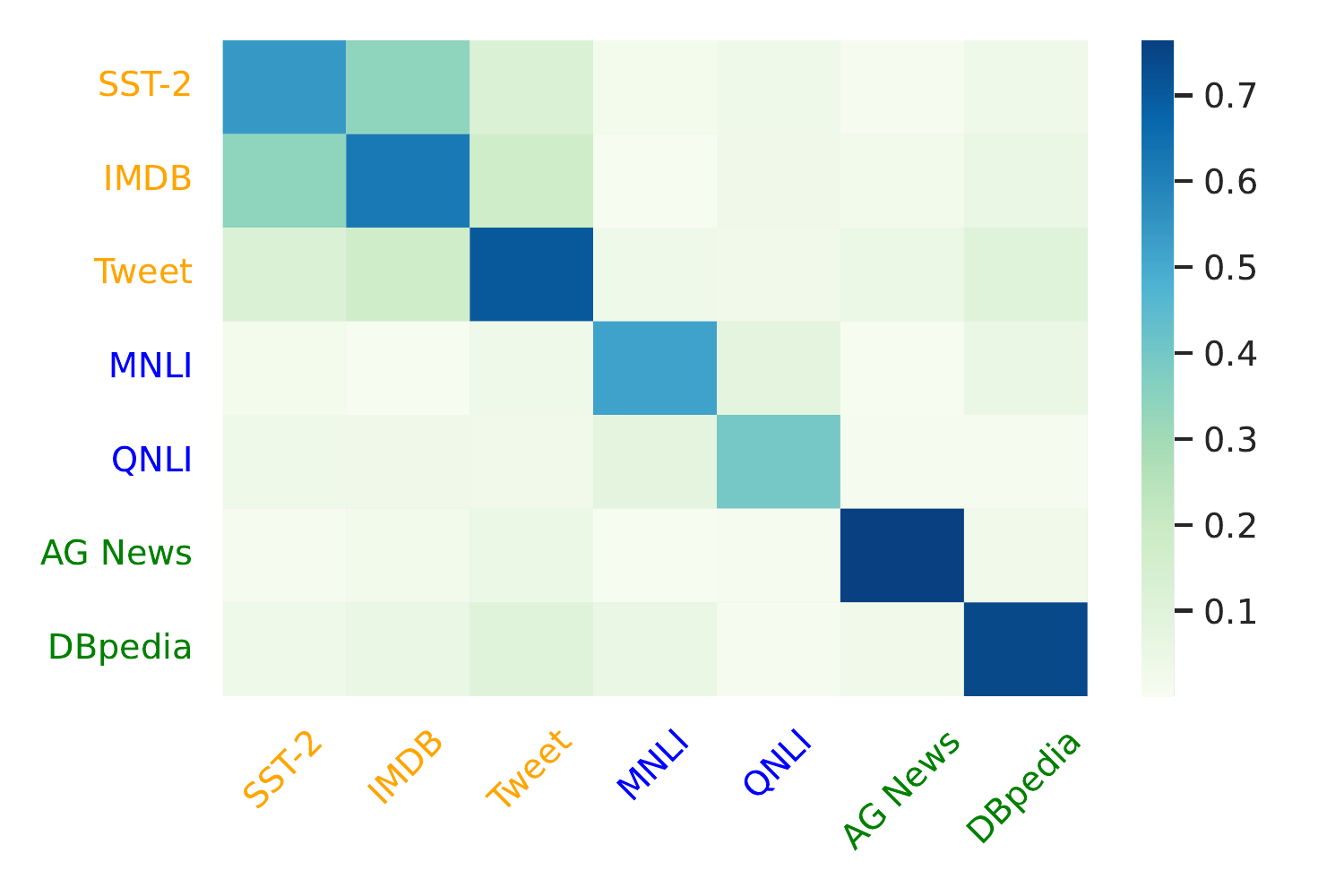}
    }
    \subfigure[Layer $4$]{
	    \includegraphics[width=0.30\textwidth]{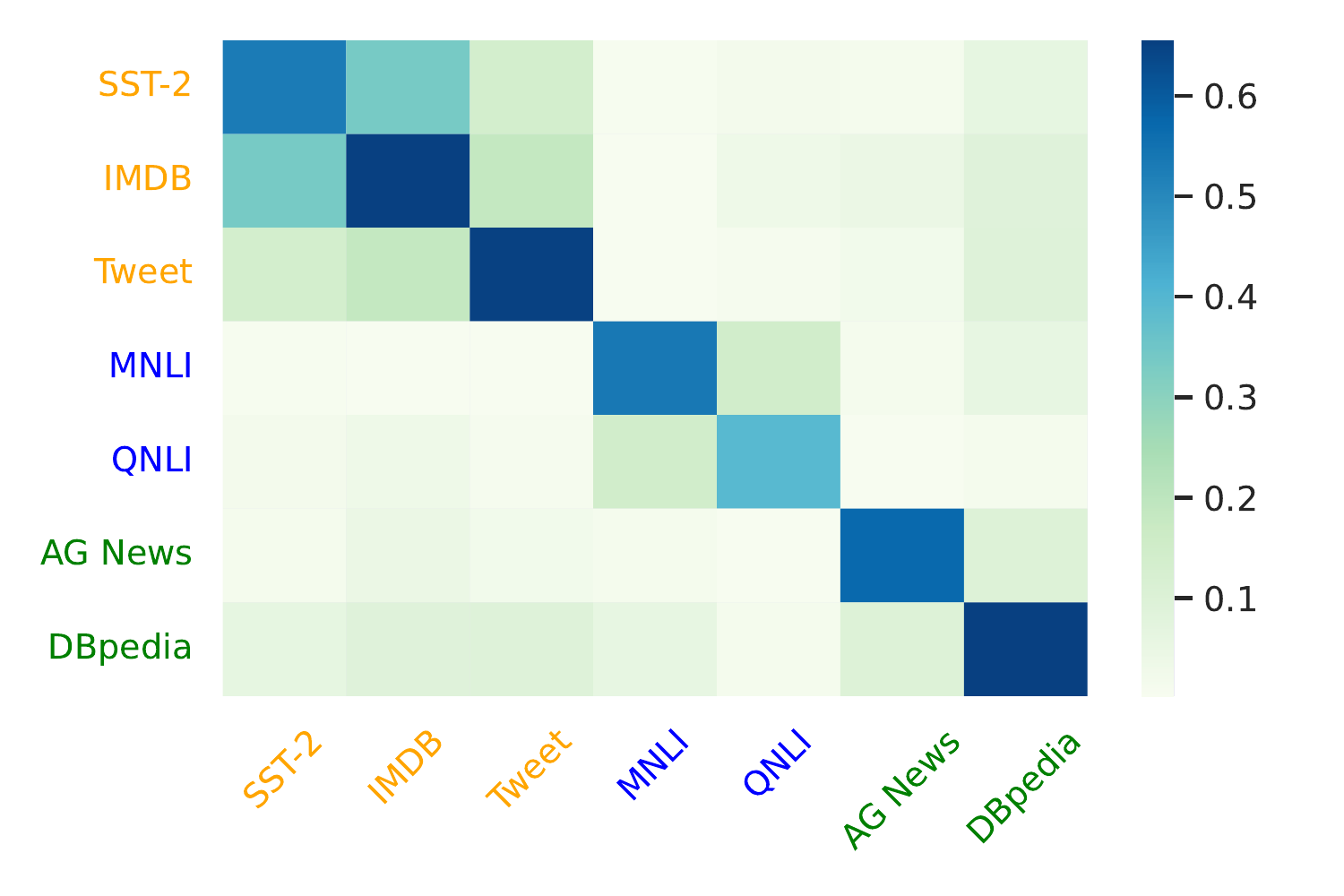}
    }
    \subfigure[Layer $5$]{
	    \includegraphics[width=0.30\textwidth]{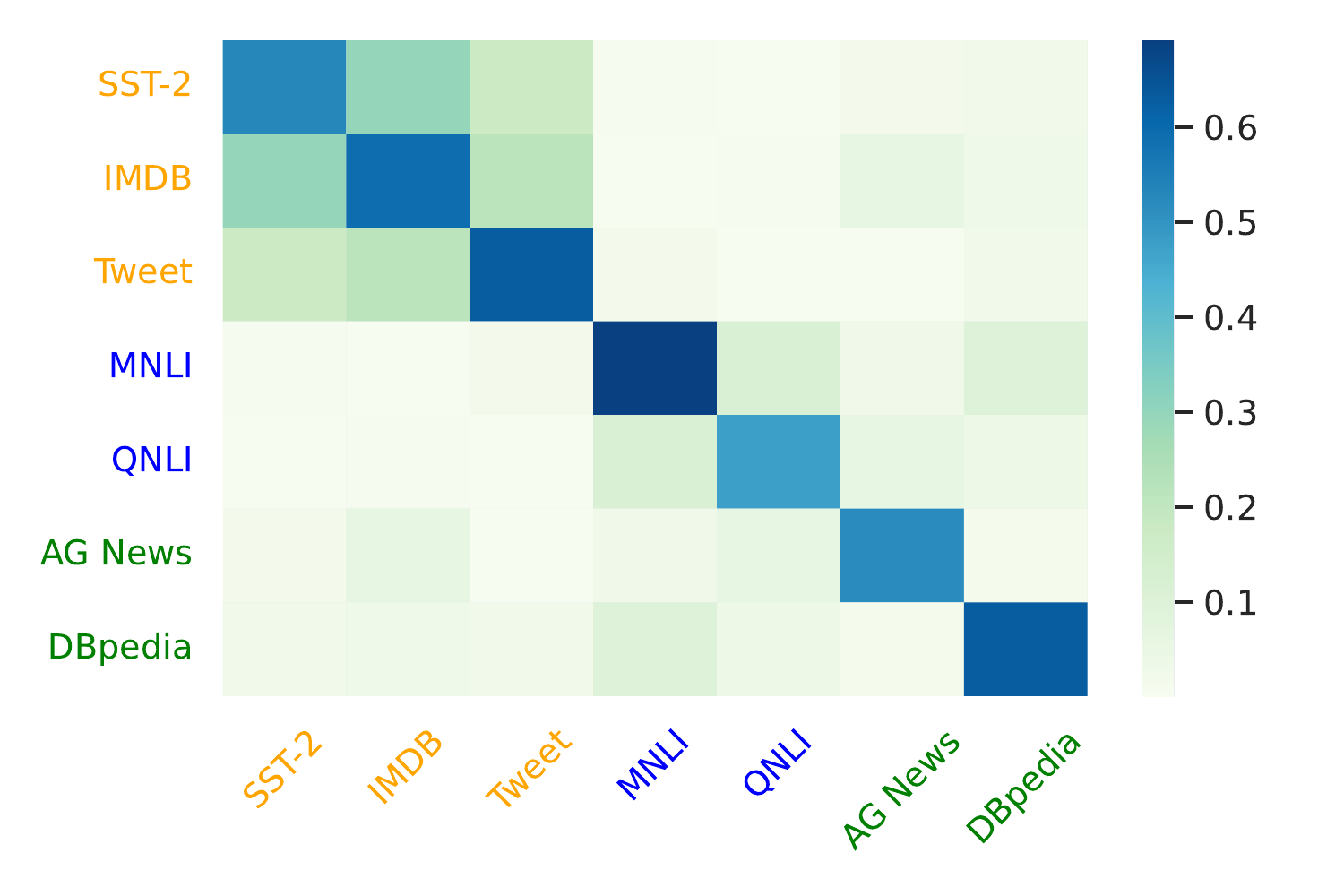}
    }
    \subfigure[Layer $6$]{
	    \includegraphics[width=0.30\textwidth]{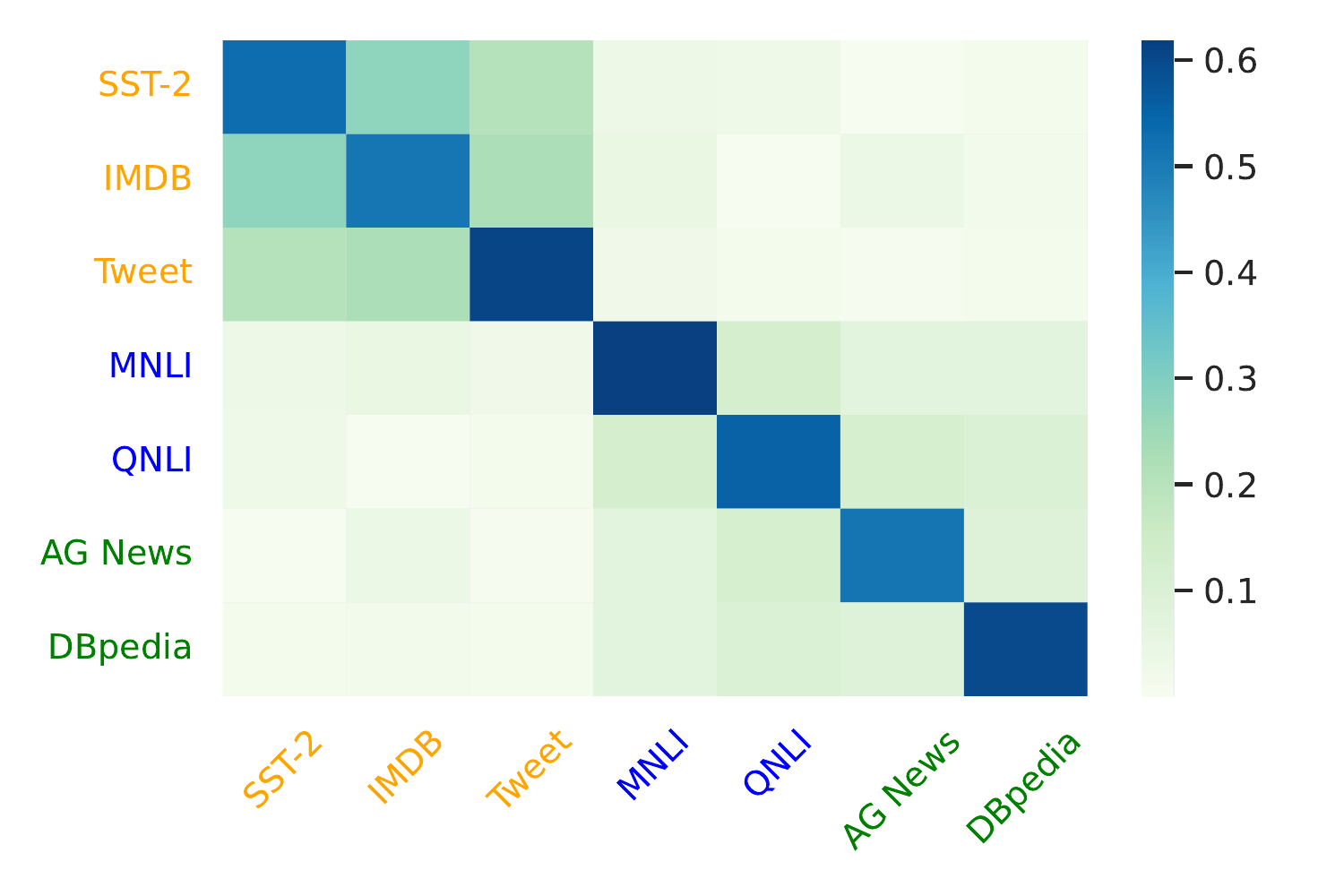}
    }
    \subfigure[Layer $7$]{
	    \includegraphics[width=0.30\textwidth]{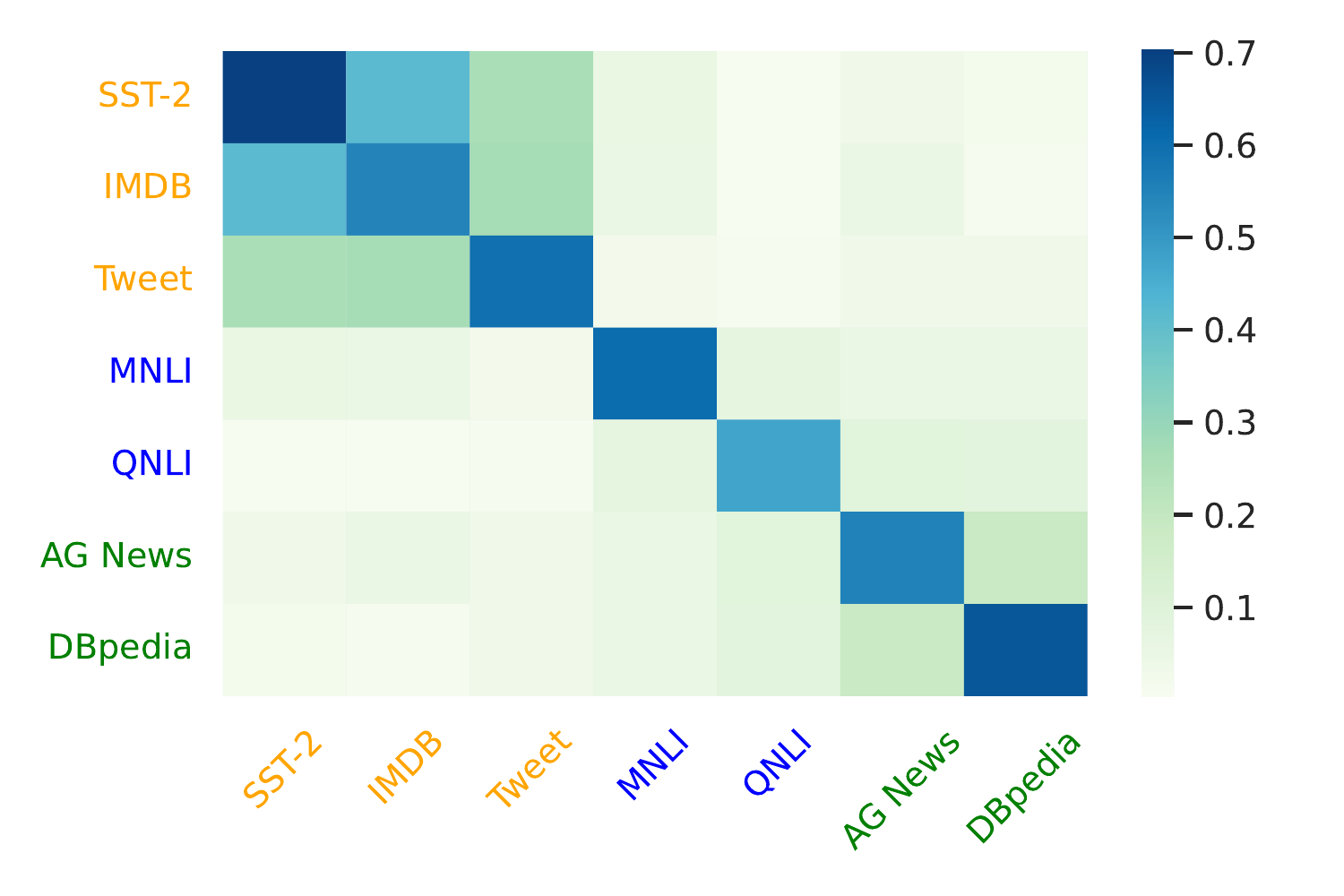}
    }
    \subfigure[Layer $8$]{
	    \includegraphics[width=0.30\textwidth]{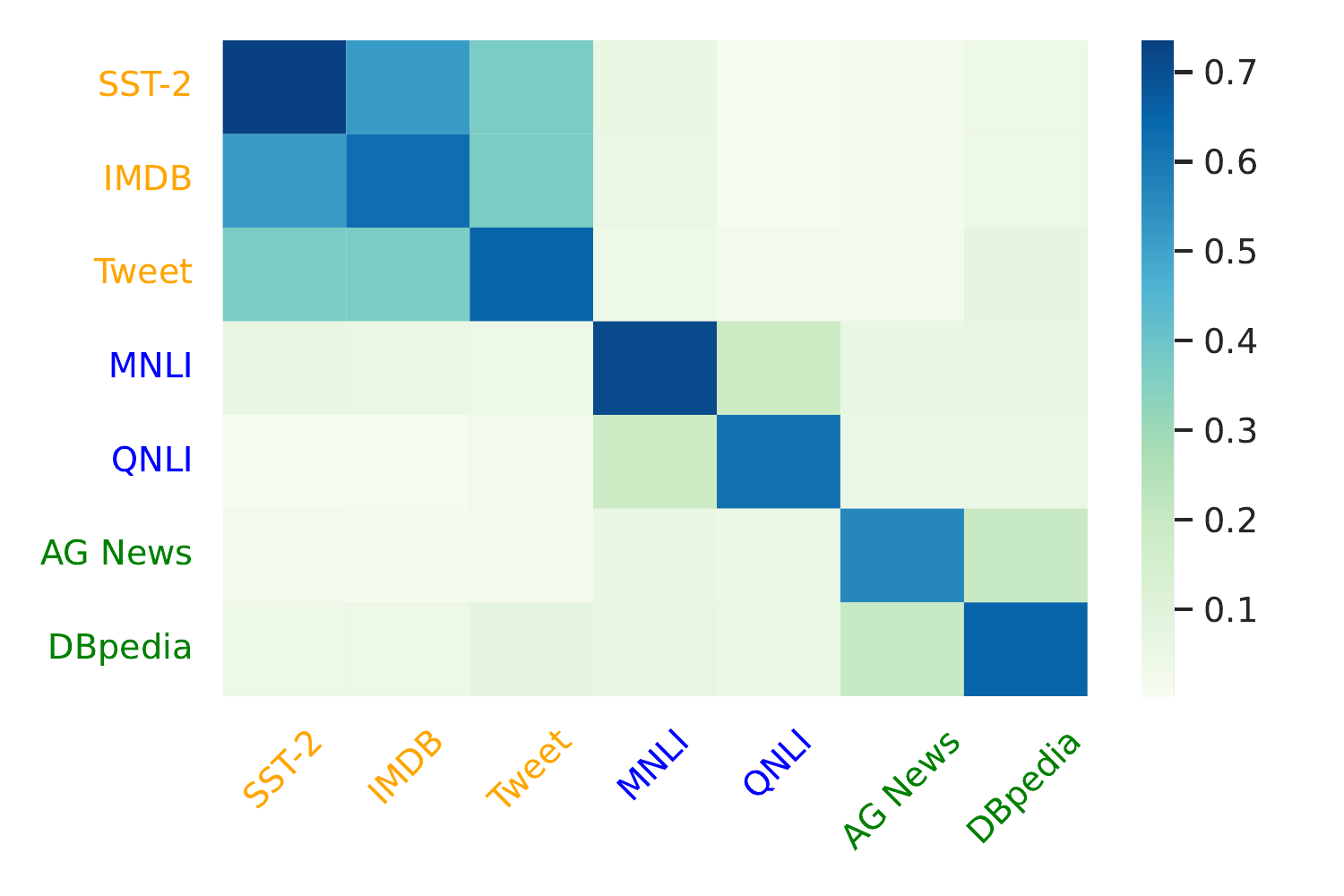}
    }
    \subfigure[Layer $9$]{
	    \includegraphics[width=0.30\textwidth]{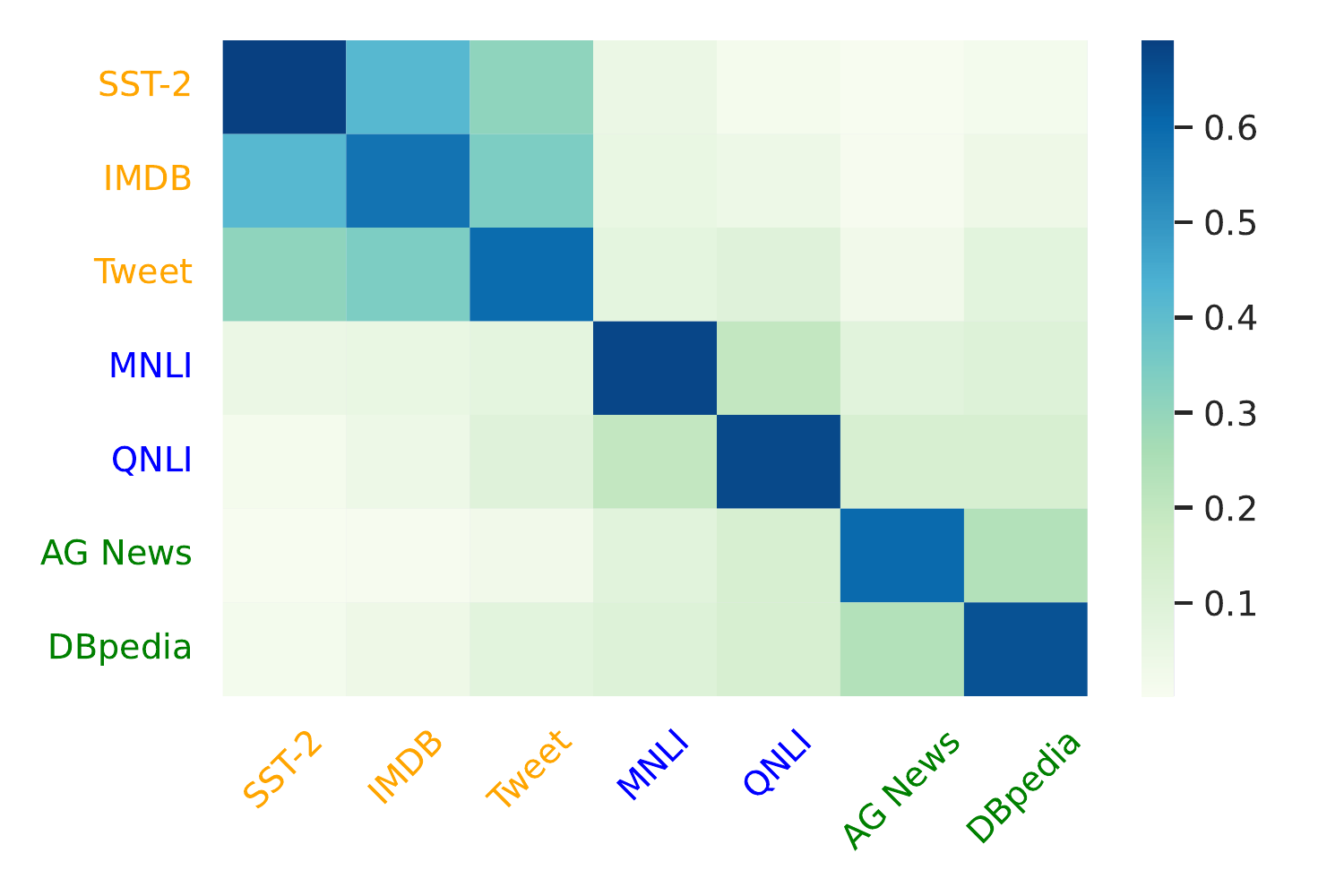}
    }
    \subfigure[Layer $10$]{
	    \includegraphics[width=0.30\textwidth]{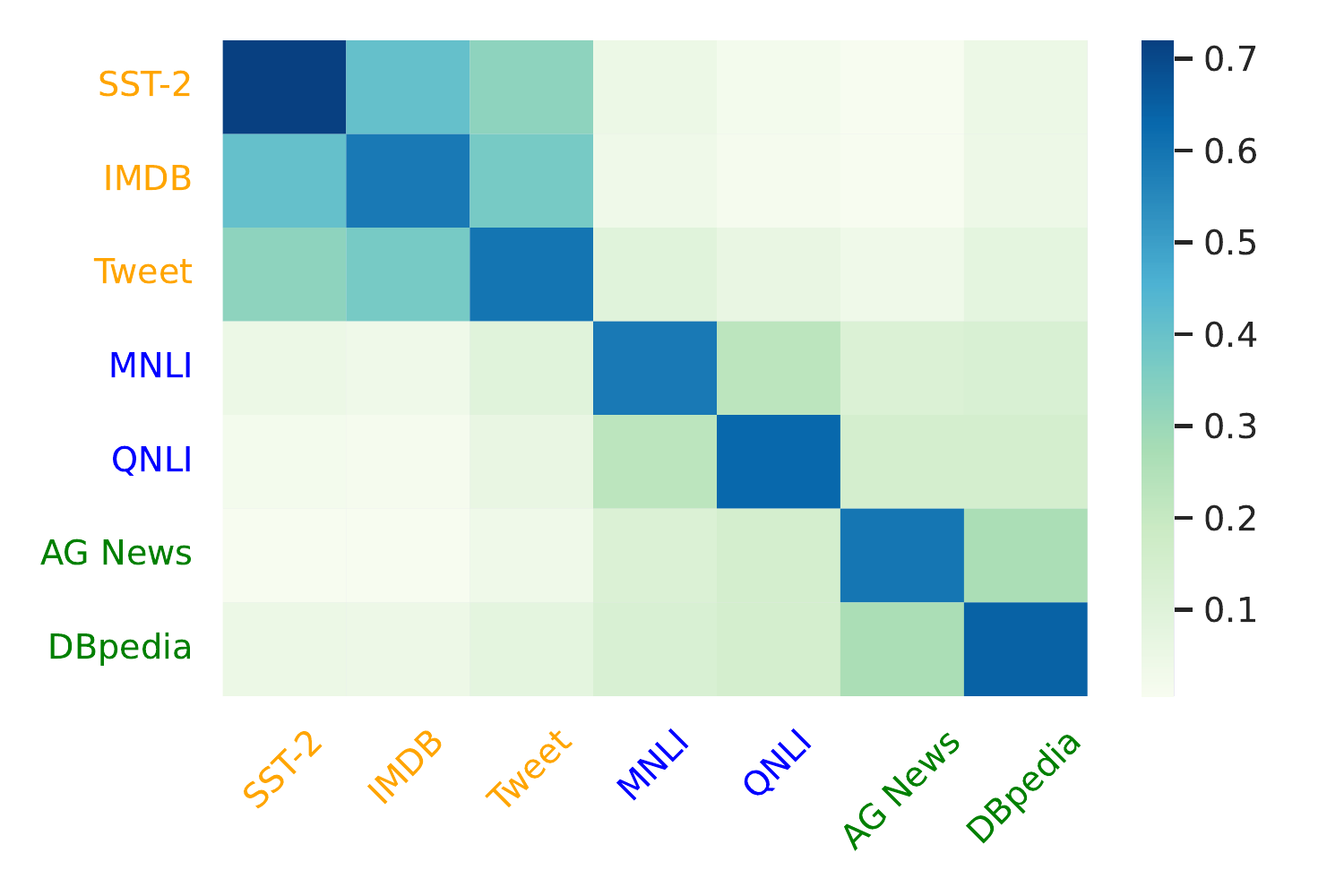}
    }
    \subfigure[Layer $11$]{
	    \includegraphics[width=0.30\textwidth]{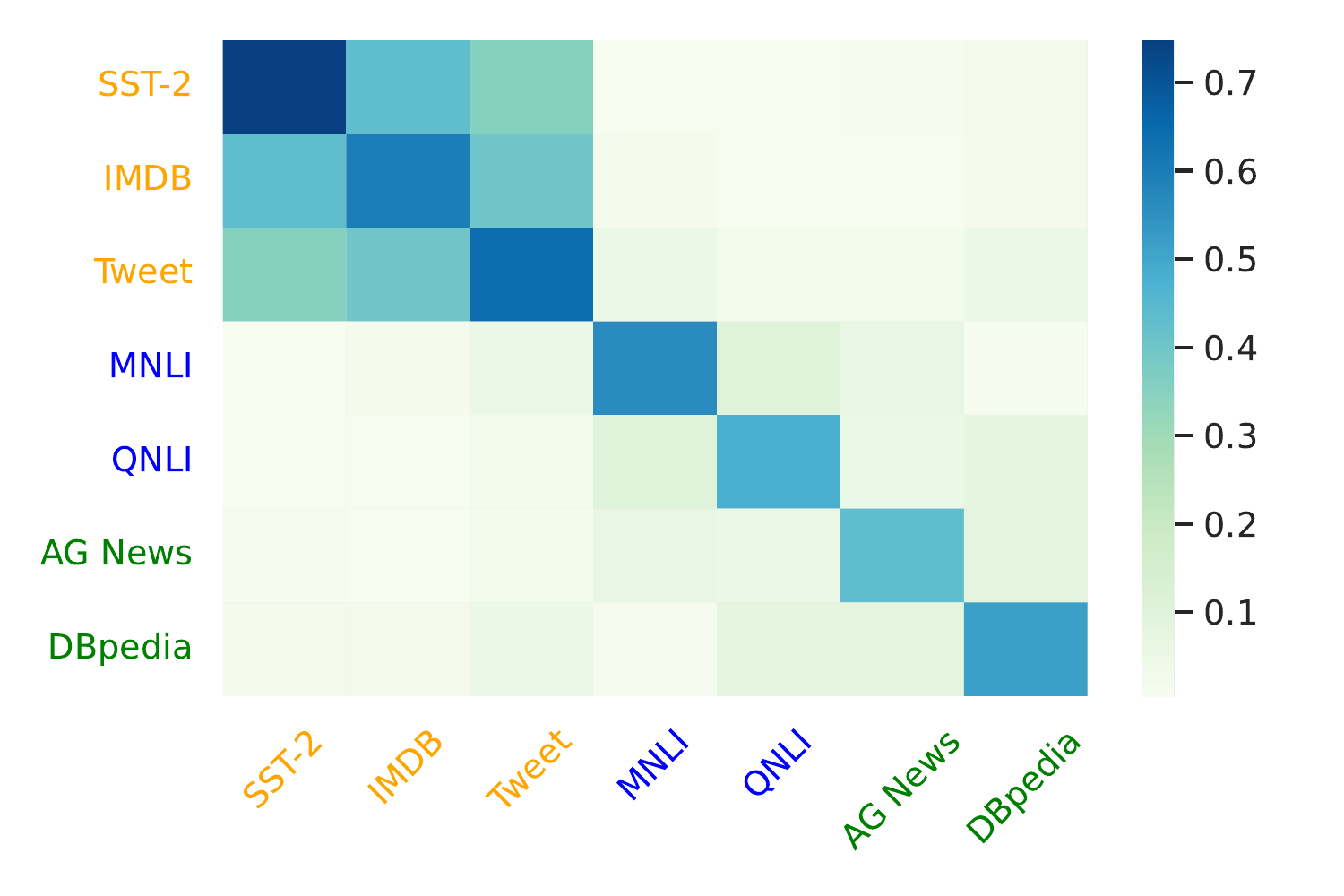}
    }
    \subfigure[Layer $12$]{
	    \includegraphics[width=0.30\textwidth]{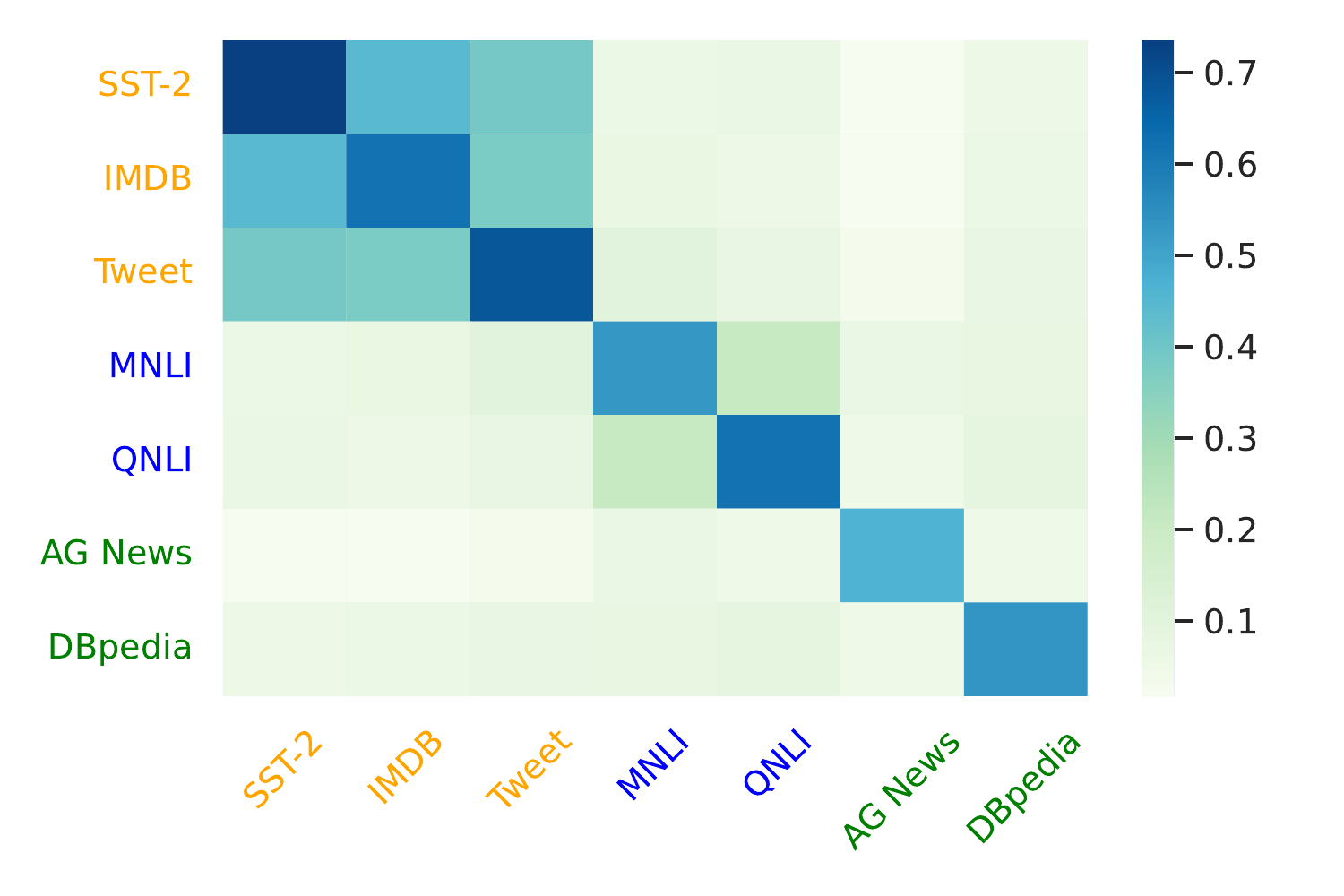}
    }
    \caption{Spearman's rank correlations between the neuron predictivity orders of different tasks on different layers. Layer $1$ is the bottom layer near the inputs, and layer $12$ is the top layer near the outputs.}
    \label{fig:app_spearman_neuron_layer}
\end{figure*}

\cref{fig:spearman_neuron} shows the overall Spearman's rank correlations between the neuron predictivity orders of different tasks, which is averaged over the $12$ layers of \RBT. Here we further present the layer-wise correlations in \cref{fig:app_spearman_neuron_layer}, from which we can see the skill neurons are more and more task-specific from the bottom layer to the top layer, which is consistent with the probing findings~\citep{Liu2019LinguisticKA} showing that PLMs tend to learn general skills in the lower layers and learn specific skills in the higher layers. These results suggest that our neuron-finding method can find both neurons encoding general skills in the lower layers and neurons encoding specific skills in the lower layers, but the found top skill neurons are task-specific in general (\cref{fig:spearman_neuron}). In this work, we focus on the task-specific top skill neurons and leave careful study for the neurons encoding general skills in future work.

\section{More Word Selectivity Results}
\label{app:word_selectivity}
In \cref{tab:words_SST2}, we show the related words for \texttt{SST-2}. Here we further show the results for the other tasks in \cref{tab:app_word_selectivity}. We can see these related words generally do not convey clues about solving the tasks.
\begin{table*}[!t]
\small
\centering
\begin{adjustbox}{max width=1\linewidth}
{

\begin{tabular}{l|lc}
\toprule
\multirow{6}{*}{\texttt{IMDB}}    & \multicolumn{2}{c}{Cosine Similarity}                                                                                                                                                                                                         \\ \cline{2-3} 
                                  & Top    & \texttt{legged}, \texttt{turnout}, \texttt{ladder}, \texttt{heid}, \texttt{flexible}, \texttt{Quite}, \texttt{contrary}, \texttt{runs}, \texttt{Reference}, \texttt{enqu}                                                                     \\ \cline{2-3} 
                                  & Bottom & \texttt{qq}, \texttt{qa}, \texttt{Capture}, \texttt{Import}, \texttt{Tripoli}, \texttt{hereby}, \texttt{eus}, \texttt{,}, \texttt{rip}, \texttt{Lima}                                                                                    \\ \cline{2-3} 
                                  & \multicolumn{2}{c}{Average Activation}                                                                                                                                                                                                        \\ \cline{2-3} 
                                  & Top    & \texttt{success}, \texttt{Kund}, \texttt{Sanctuary}, \texttt{Lim}, \texttt{Wave}, \texttt{dele}, \texttt{Crystal}, \texttt{flung}, \texttt{Kerala}, \texttt{.............}                                                               \\ \cline{2-3} 
                                  & Bottom & \begin{tabular}[c]{@{}c@{}}\texttt{vation}, \texttt{goodbye}, \texttt{concludes}, \texttt{bye}, \texttt{Congratulations},\\ \texttt{Congratulations}, \texttt{Fare}, \texttt{farewell}, \texttt{BY}, \texttt{ceremony},\end{tabular} \\ \midrule
\multirow{6}{*}{\texttt{Tweet}}   & \multicolumn{2}{c}{Cosine Similarity}                                                                                                                                                                                                         \\ \cline{2-3} 
                                  & Top    & \texttt{atican}, \texttt{uras}, \texttt{isman}, \texttt{anan}, \texttt{Luck}, \texttt{Merit}, \texttt{Character}, \texttt{alth}, \texttt{atching}, \texttt{character},                                                                        \\ \cline{2-3} 
                                  & Bottom & \texttt{Register}, \texttt{enzymes}, \texttt{elsen}, \texttt{Registrar}, \texttt{tasting}, \texttt{regist}, \texttt{soils}, \texttt{µ}, \texttt{Chambers}, \texttt{LINE},                                                            \\ \cline{2-3} 
                                  & \multicolumn{2}{c}{Average Activation}                                                                                                                                                                                                        \\ \cline{2-3} 
                                  & Top    & \texttt{dh}, \texttt{Titan}, \texttt{utable}, \texttt{exited}, \texttt{iOS}, \texttt{chel}, \texttt{loophole}, \texttt{acious}, \texttt{520}, \texttt{Harmony},                                                                         \\ \cline{2-3} 
                                  & Bottom & \texttt{spike}, \texttt{unbelievably}, \texttt{Toxic}, \texttt{prov}, \texttt{RIS}, \texttt{resulting}, \texttt{risks}, \texttt{rising}, \texttt{ues}, \texttt{reapp},                                                                \\ \midrule
\multirow{6}{*}{\texttt{MNLI}}    & \multicolumn{2}{c}{Cosine Similarity}                                                                                                                                                                                                         \\ \cline{2-3} 
                                  & Top    & \begin{tabular}[c]{@{}c@{}}\texttt{trigger}, \texttt{Pis}, \texttt{deadlines}, \texttt{Launch}, \texttt{mares},\\ \texttt{PROGRAM}, \texttt{Congratulations}, \texttt{Success}, \texttt{Congratulations}, \texttt{Gig},\end{tabular}   \\ \cline{2-3} 
                                  & Bottom & \texttt{minim}, \texttt{xt}, \texttt{spoof}, \texttt{dism}, \texttt{avoid}, \texttt{asive}, \texttt{WN}, \texttt{offset}, \texttt{inter}, \texttt{antiqu},                                                                                \\ \cline{2-3} 
                                  & \multicolumn{2}{c}{Average Activation}                                                                                                                                                                                                        \\ \cline{2-3} 
                                  & Top    & \texttt{nickel}, \texttt{grun}, \texttt{cluded}, \texttt{91}, \texttt{handled}, \texttt{secure}, \texttt{very}, \texttt{dairy}, \texttt{gent}, \texttt{Roses},                                                                          \\ \cline{2-3} 
                                  & Bottom & \texttt{ayed}, \texttt{disl}, \texttt{ect}, \texttt{wipes}, \texttt{screwed}, \texttt{resistance}, \texttt{aw}, \texttt{ruin}, \texttt{shrinking}, \texttt{spite},                                                                    \\ \midrule
\multirow{6}{*}{\texttt{QNLI}}    & \multicolumn{2}{c}{Cosine Similarity}                                                                                                                                                                                                         \\ \cline{2-3} 
                                  & Top    & \texttt{otyp}, \texttt{disemb}, \texttt{sidel}, \texttt{melanch}, \texttt{unint}, \texttt{outwe}, \texttt{umbnails}, \texttt{precedence}, \texttt{unfl}, \texttt{Sym},                                                                \\ \cline{2-3} 
                                  & Bottom & \texttt{314}, \texttt{223}, \texttt{313}, \texttt{234}, \texttt{,}, \texttt{316}, \texttt{341}, \texttt{463}, \texttt{238}, \texttt{261},                                                                                                     \\ \cline{2-3} 
                                  & \multicolumn{2}{c}{Average Activation}                                                                                                                                                                                                        \\ \cline{2-3} 
                                  & Top    & \texttt{eds}, \texttt{adding}, \texttt{apocalypse}, \texttt{strawberry}, \texttt{apopt}, \texttt{Kid}, \texttt{leaf}, \texttt{Silent}, \texttt{technical},                                                                              \\ \cline{2-3} 
                                  & Bottom & \texttt{entrepreneurial}, \texttt{Econom}, \texttt{Columb}, \texttt{prime}, \texttt{roleum}, \texttt{Trade}, \texttt{rounded}, \texttt{isner}, \texttt{enz}, \texttt{158},                                                                \\ \midrule
\multirow{6}{*}{\texttt{AG News}} & \multicolumn{2}{c}{Cosine Similarity}                                                                                                                                                                                                         \\ \cline{2-3} 
                                  & Top    & \texttt{aukee}, \texttt{erity}, \texttt{lambda}, \texttt{ropolitan}, \texttt{roxy}, \texttt{LAN}, \texttt{ylon}, \texttt{incinn}, \texttt{oslav}, \texttt{coni},                                                                              \\ \cline{2-3} 
                                  & Bottom & \texttt{Gross}, \texttt{Villa}, \texttt{Uri}, \texttt{ende}, \texttt{Summary}, \texttt{Gallup}, \texttt{Temp}, \texttt{Rog}, \texttt{RP}, \texttt{Ram},                                                                              \\ \cline{2-3} 
                                  & \multicolumn{2}{c}{Average Activation}                                                                                                                                                                                                        \\ \cline{2-3} 
                                  & Top    & \texttt{fight}, \texttt{desert}, \texttt{Merge}, \texttt{Mail}, \texttt{Mid}, \texttt{Rankings}, \texttt{istic}, \texttt{**}, \texttt{berries}, \texttt{Pen},                                                                             \\ \cline{2-3} 
                                  & Bottom & \texttt{ETS}, \texttt{107}, \texttt{Line}, \texttt{106}, \texttt{observers}, \texttt{Ranked}, \texttt{EB}, \texttt{ido}, \texttt{Bass}, \texttt{alf},                                                                                      \\ \midrule
\multirow{6}{*}{\texttt{DBpedia}} & \multicolumn{2}{c}{Cosine Similarity}                                                                                                                                                                                                         \\ \cline{2-3} 
                                  & Top    & \texttt{ming}, \texttt{umbered}, \texttt{hind}, \texttt{utter}, \texttt{pepper}, \texttt{scr}, \texttt{increment}, \texttt{usher}, \texttt{empt}, \texttt{atmospheric},                                                               \\ \cline{2-3} 
                                  & Bottom & \texttt{Chron}, \texttt{kan}, \texttt{Div}, \texttt{Case}, \texttt{Thread}, \texttt{Role}, \texttt{Crash}, \texttt{Mode}, \texttt{Tank}, \texttt{Apps},                                                                                    \\ \cline{2-3} 
                                  & \multicolumn{2}{c}{Average Activation}                                                                                                                                                                                                        \\ \cline{2-3} 
                                  & Top    & \texttt{Bubble}, \texttt{mailed}, \texttt{Ari}, \texttt{razen}, \texttt{Perspective}, \texttt{ogical}, \texttt{Gin}, \texttt{Disney}, \texttt{icons}, \texttt{Huang},                                                                   \\ \cline{2-3} 
                                  & Bottom & \texttt{Jacob}, \texttt{Boss}, \texttt{Dad}, \texttt{trough}, \texttt{Shiny}, \texttt{carn}, \texttt{Gravity}, \texttt{toolbar}, \texttt{Sword}, \texttt{temple},                                                                      \\ \bottomrule
\end{tabular}
}
\end{adjustbox}
\caption{Related words for various tasks' top skill neurons.}
\label{tab:app_word_selectivity}
\end{table*}

\section{Discussions on Neuron-Finding Design Choices}
\label{app:design_choice}
In this section, we discuss some potential other design choices that may be used in finding important skill neurons to provide more background about why we choose the method described in \cref{sec:find_method} finally and inspire future works.

\paragraph{Perturbation-based neuron finding.}
A natural way to define the importance of a neuron (to a task) is to perturb the neurons and see how they influence the predictions. The perturbation-based method has been used in previous analysis works~\citep{michel2019sixteen}, and we also adopt them in our analytical experiments. But we and many other neuron-level analysis works~\citep{dalvi2019one,durrani2020analyzing,antverg2021pitfalls,suau2020finding,geva2021transformer,dai2021knowledge} cannot directly use this method to locate important neurons. This is because of the efficiency issue. Perturbing every individual neuron is unaffordable.

\paragraph{Is prompt tuning necessary?}
This work starts from an interesting empirical finding, i.e., the skill neuron phenomenon. This finding is based on prompt tuning. In \cref{sec:do_encode} and \cref{fig:acc_dist}, we show that previous methods without prompt tuning cannot well locate the skill neurons. Since we focus on confirming the finding and exploring the properties of skill neurons, we conduct all the experiments based on prompt tuning and do not explore whether it is necessary. Intuitively, as our experiments suggest that the emergence of skill neurons does not depend on prompt tuning but is mostly an intrinsic property for pre-trained Transformer-based language models, we believe prompt tuning may not be the only way to locate skill neurons. We will explore other methods without prompt tuning in future works, which may bring some benefits, like improving overall efficiency.

\paragraph{Other ways to define neuron's predictivity.}
In \cref{sec:binary_find}, we define the predictivity of a neuron (1) using the maximum over prompt tokens and (2) considering both the positive and negative correlations. These two choices are made with preliminary experiments. \cref{fig:design_choice} shows an example, from which we can see that when defining neuron's predictivity using the mean values over prompt tokens or only considering the positive correlations, the predictivities will be significantly under-estimated than the default definition in \cref{sec:binary_find}.

\section{Experiments following \citet{morcos2018importance}}
\label{app:single_direction}
Some previous works~\citep{Bau2017NetworkDQ,Mu2020CompositionalEO} suggest that selective neurons contribute more to model accuracies. In \cref{sec:do_encode}, we also find that perturbing selective skill neurons leads to more performance drop. However, \citet{morcos2018importance} draw opposite conclusions and find that selective and non-selective neurons are similarly important. These pose questions about why these conclusions are inconsistent.

\begin{figure*}[!t]
\small
\subcapraggedrighttrue
\subcaphangtrue
    \centering
    
    \subfigure[On \texttt{SST-2}]{
	    \includegraphics[width=0.48\textwidth]{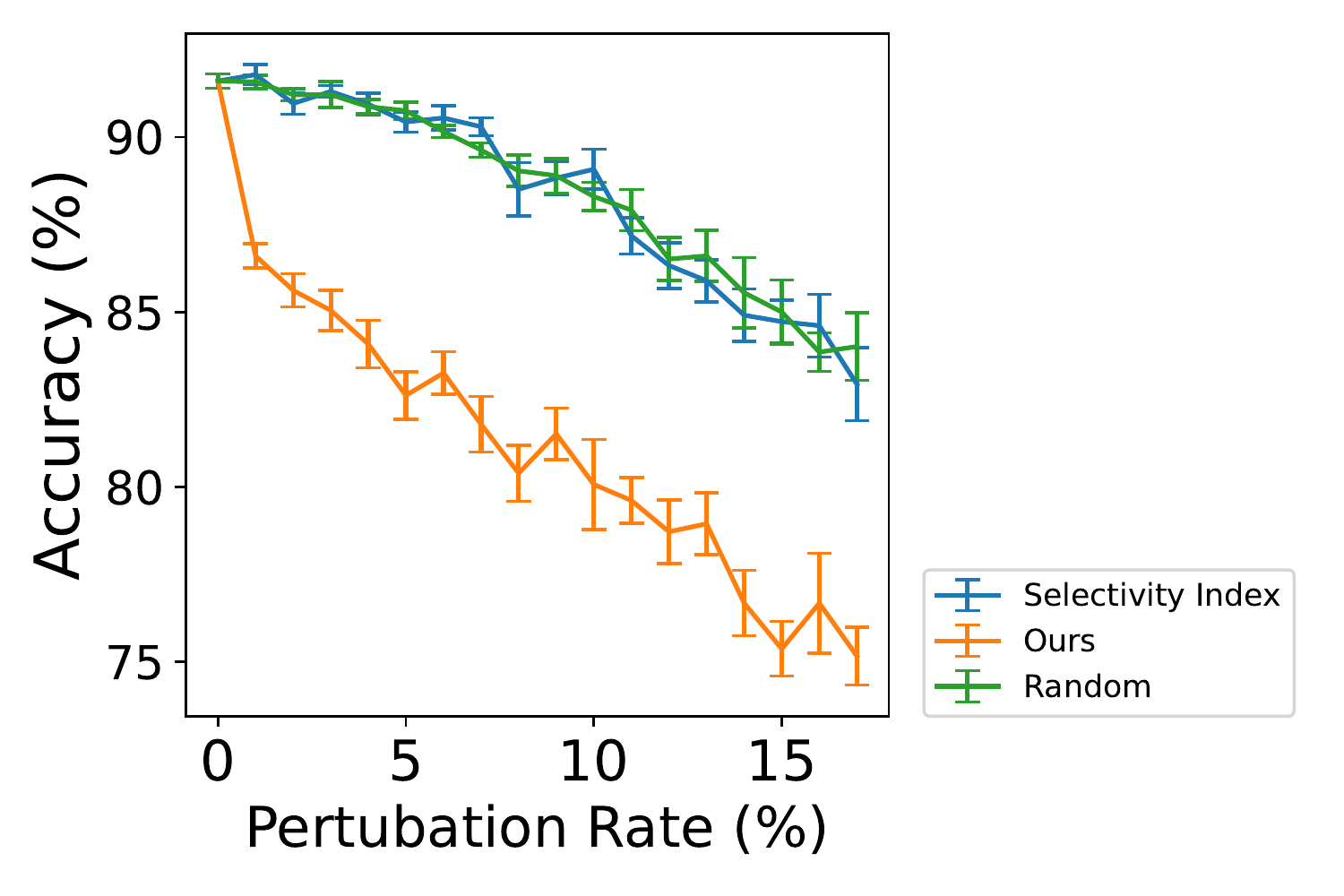}
    }
    \subfigure[On \texttt{IMDB}]{
	    \includegraphics[width=0.48\textwidth]{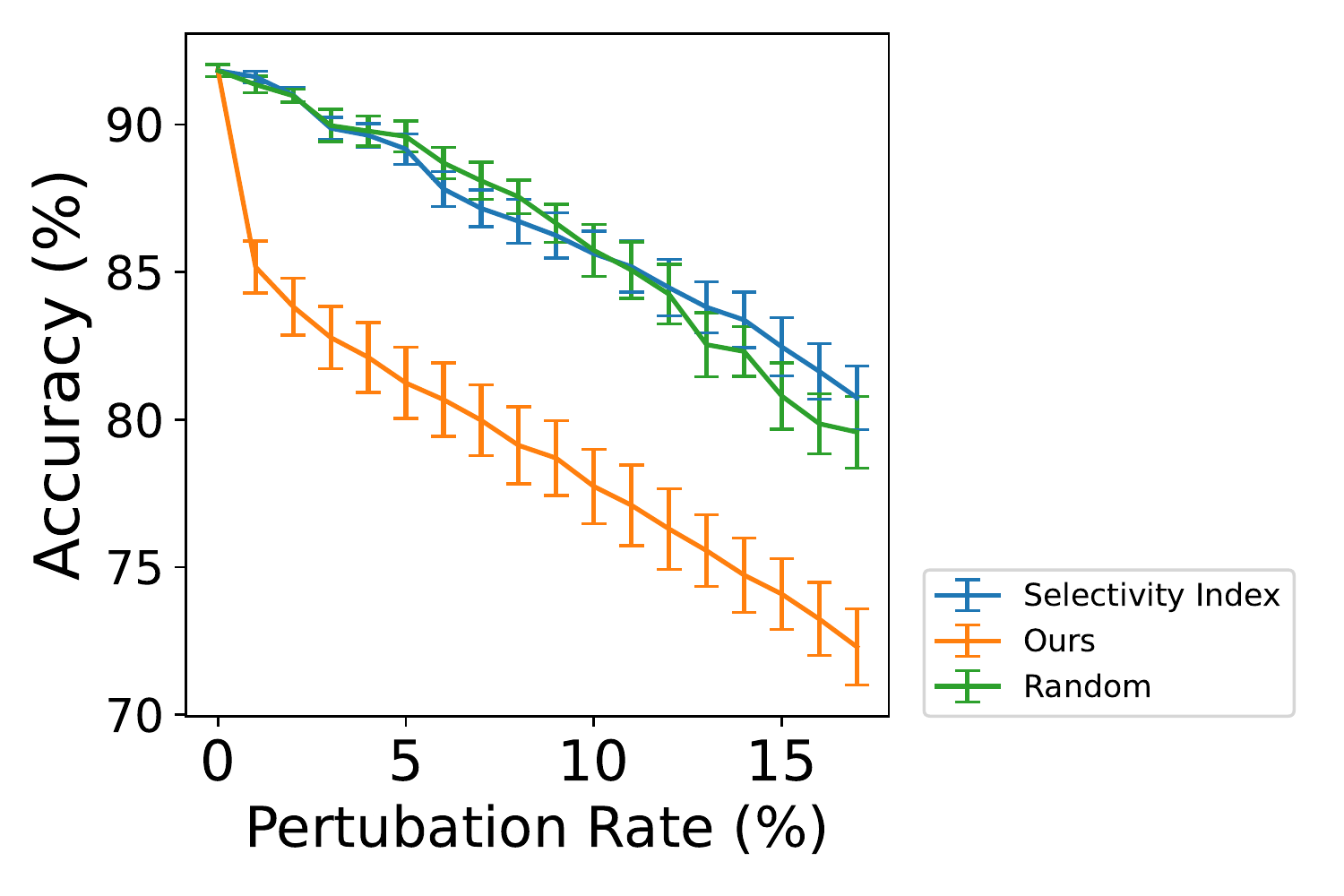}
    }
    \subfigure[On \texttt{Tweet}]{
	    \includegraphics[width=0.48\textwidth]{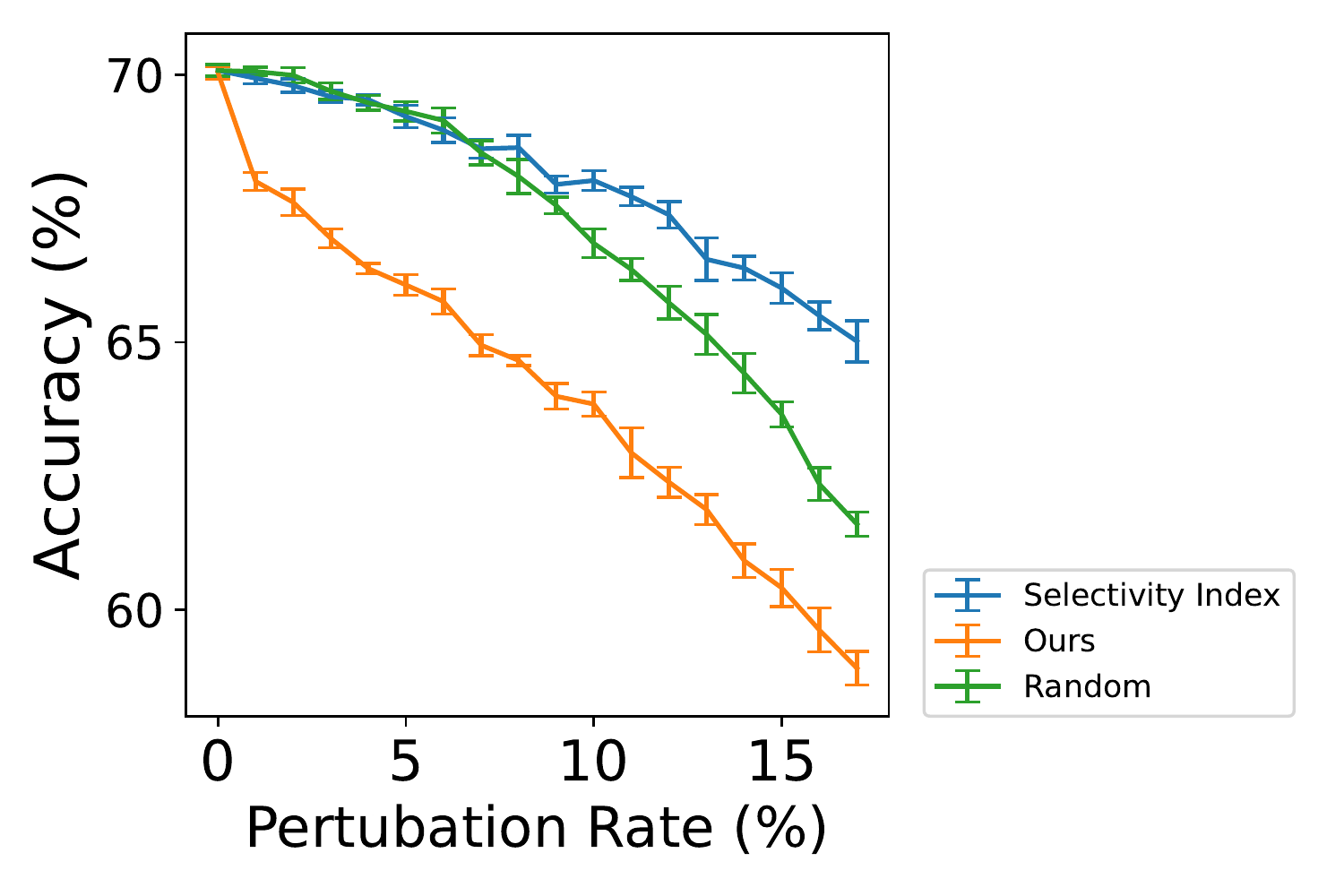}
    }
    \subfigure[On \texttt{MNLI}]{
	    \includegraphics[width=0.48\textwidth]{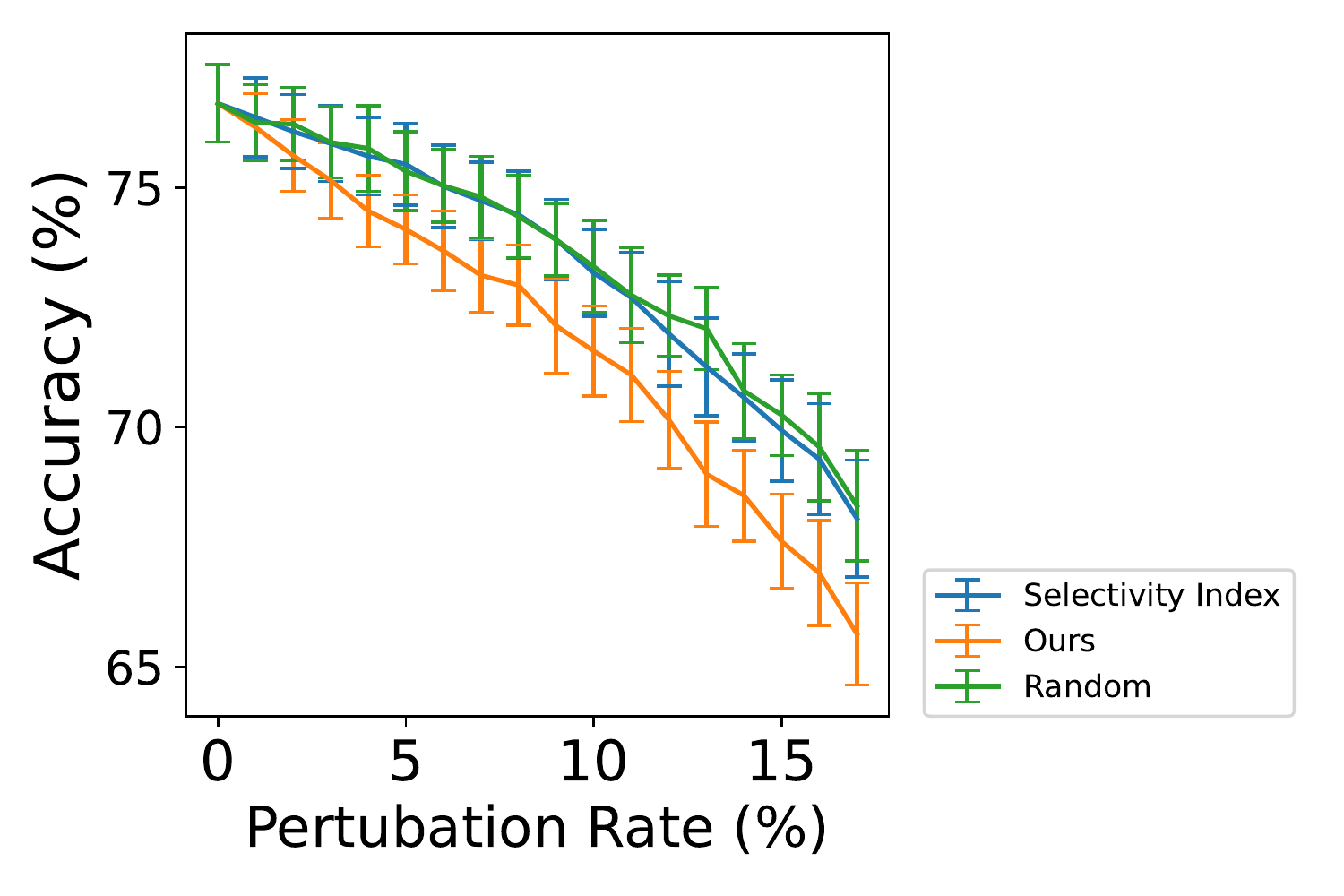}
    }
    \subfigure[On \texttt{QNLI}]{
	    \includegraphics[width=0.48\textwidth]{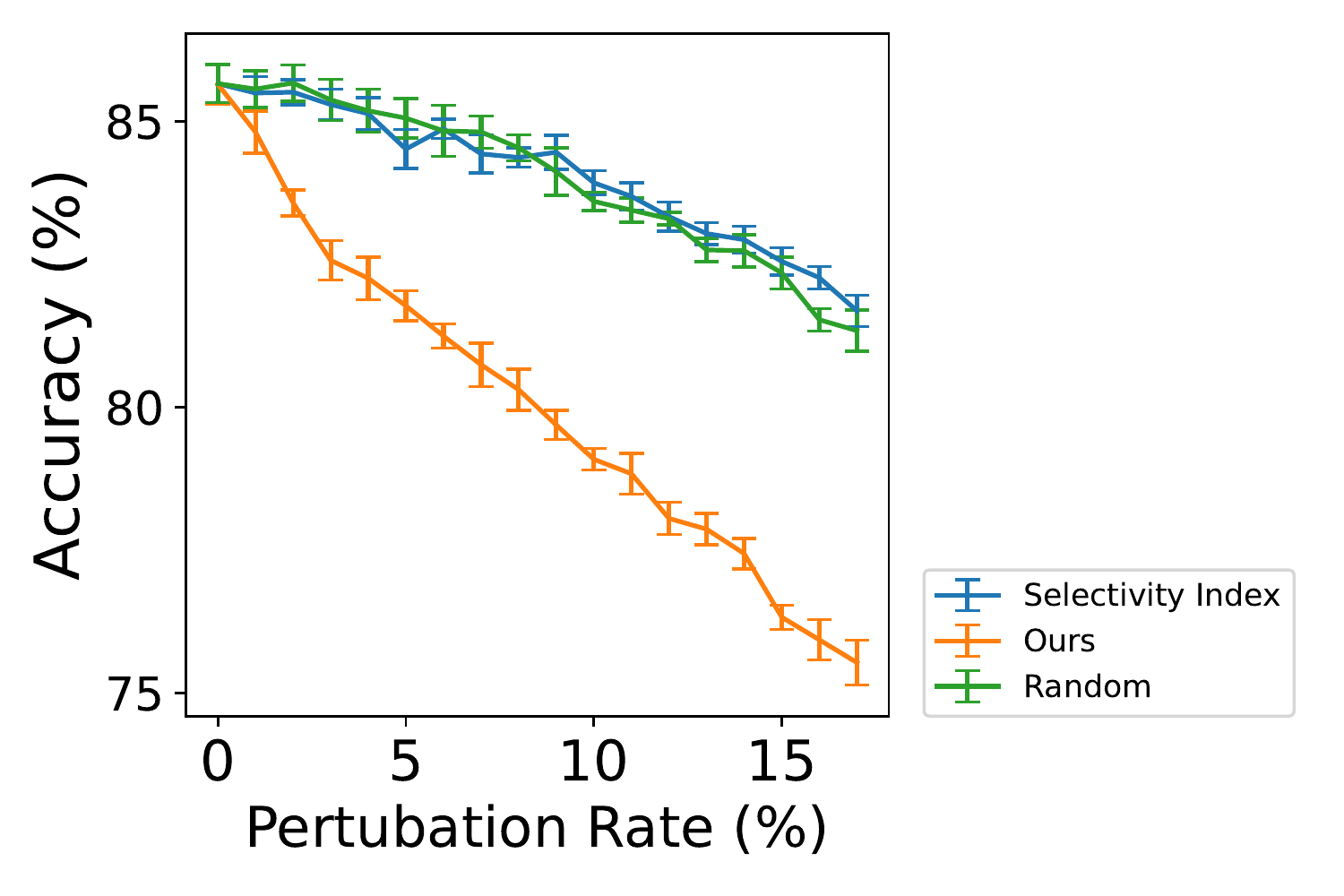}
    }
    \subfigure[On \texttt{AG News}]{
	    \includegraphics[width=0.48\textwidth]{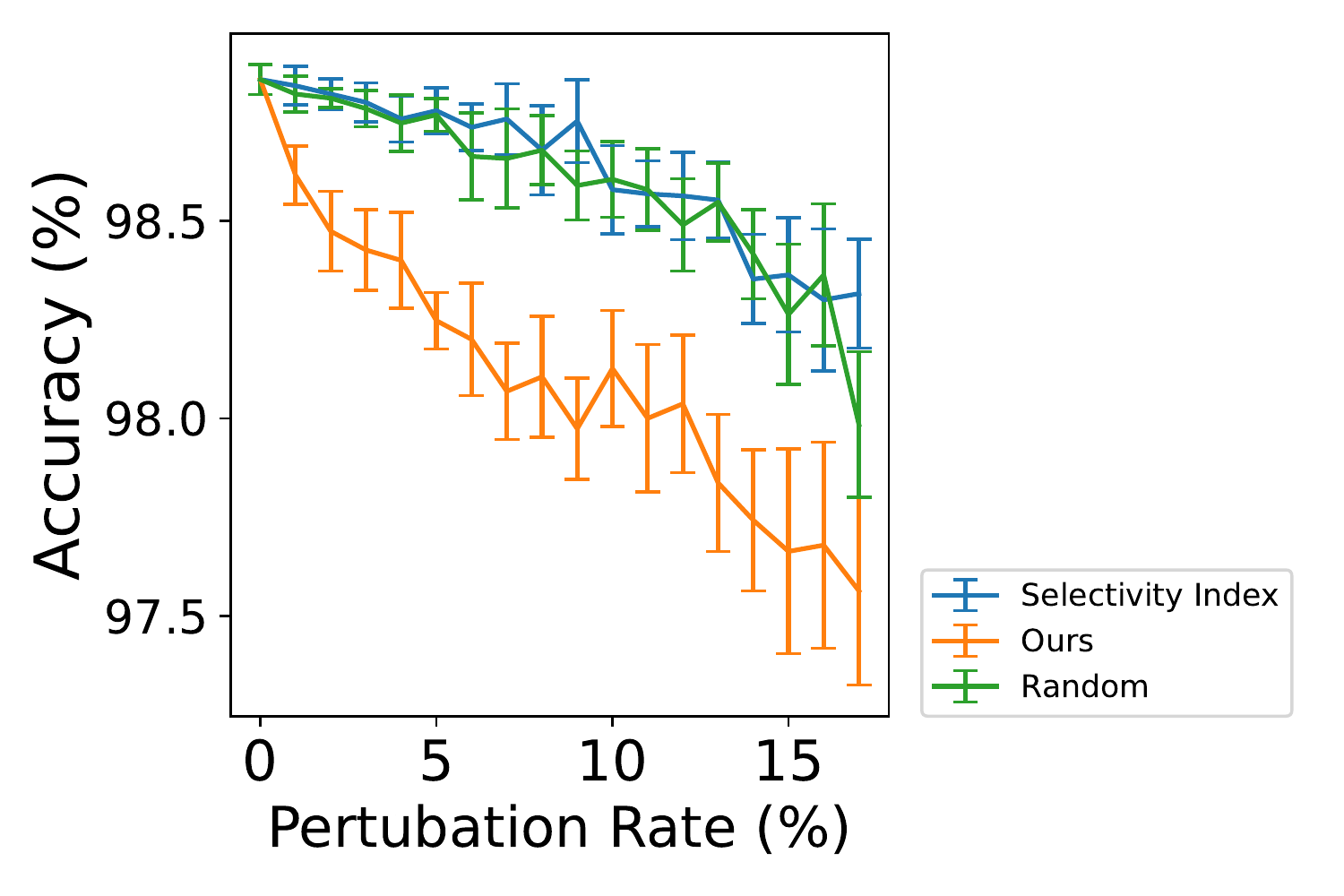}
    }
    \subfigure[On \texttt{DBpedia}]{
	    \includegraphics[width=0.48\textwidth]{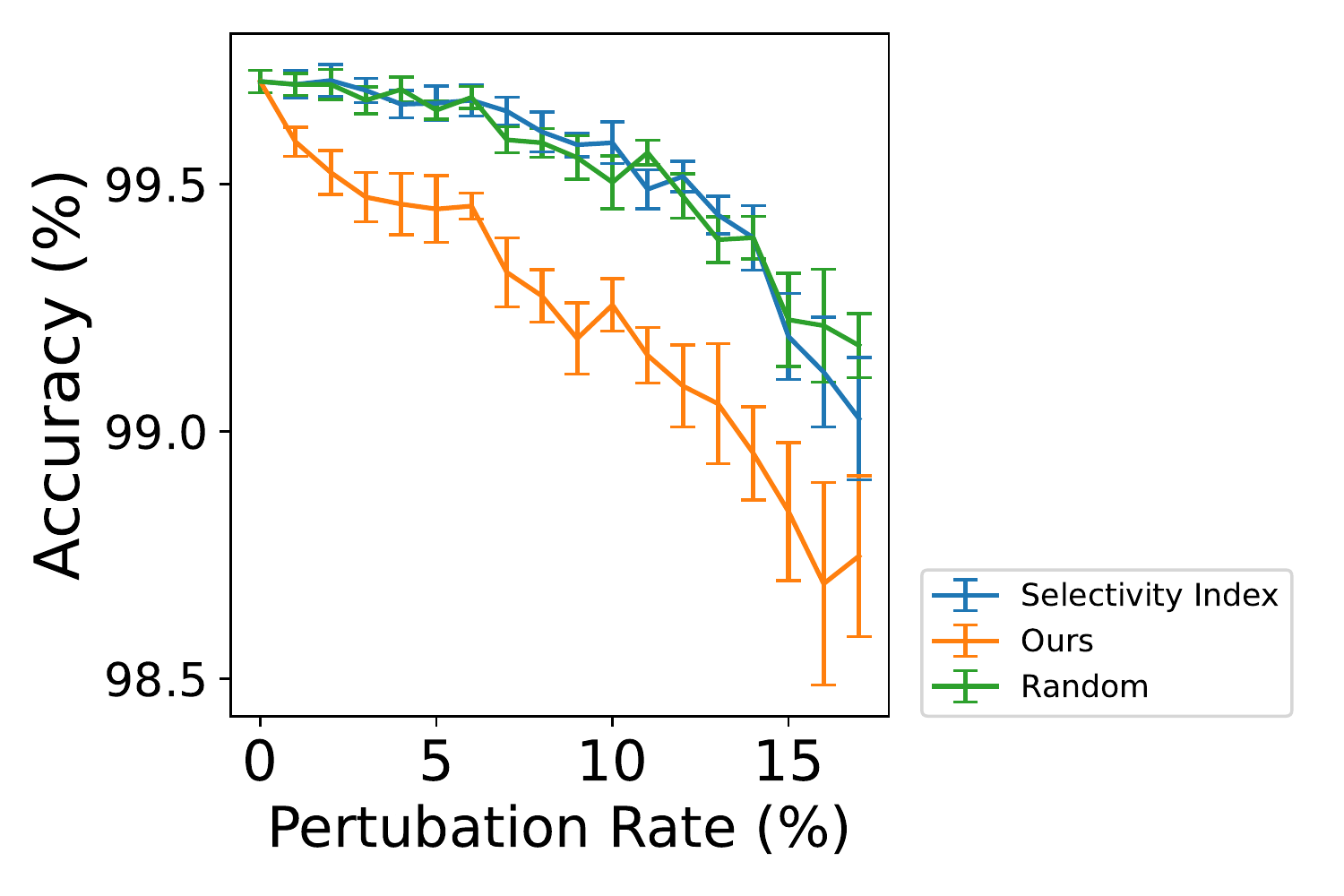}
    }
    \caption{Prompt tuning accuracies on various tasks drop along with the neuron perturbation rates. Error bars indicate $\pm 1$ s.e.m. over $5$ random trials. The perturbations are conducted in descending predictivity orders (\textit{Ours}), random orders (\textit{Random}) and descending "selectivity index"~\citep{morcos2018importance} orders (\textit{Selectivity Index}). }
    \label{fig:app_mask_trend_selectivity}
\end{figure*}

We find that except for experimental setups, the main difference between \citet{morcos2018importance} and ours lies in the definition of neuronal selectivity. \citet{morcos2018importance} define a "selectivity index" and we use the predictivity score introduced in \cref{sec:find_method}. To check whether these different definitions lead to inconsistent results, we do experiments under our setup and also try to perturb neurons in descending orders of their ``selectivity index''. The results are shown in \cref{fig:app_mask_trend_selectivity}. We can see that when using the ``selectivity index'', the found neurons are surely not more important than random neurons as reported by \citet{morcos2018importance}. But our predictivity metric can find significantly more important neurons for all the tasks.

\end{document}